\documentclass{article}
\usepackage[preprint]{neurips_2025}

\usepackage[utf8]{inputenc} % allow utf-8 input
\usepackage[T1]{fontenc}    % use 8-bit T1 fonts
\usepackage{hyperref}       % hyperlinks
\usepackage{url}            % simple URL typesetting
\usepackage{booktabs}       % professional-quality tables
\usepackage{amsfonts}
\usepackage{mathtools}
\usepackage{nicefrac}       % compact symbols for 1/2, etc.
\usepackage{microtype}      % microtypography
\usepackage{xcolor}         % colors

\usepackage{microtype}
\usepackage{graphicx}
\usepackage{subfigure}
\usepackage{booktabs} % for professional tables
\usepackage{mathrsfs}
\usepackage{url}            % simple URL typesetting
\usepackage{amsfonts}       % blackboard math symbols
\usepackage{nicefrac}       % compact symbols for 1/2, etc.
\usepackage{microtype}      % microtypography
\usepackage{multirow}
\usepackage{colortbl}
\usepackage{arydshln}

\usepackage{fancyvrb} 
\usepackage{listings}
\usepackage{placeins}
\usepackage[most]{tcolorbox}
\usepackage{bbm}

\usepackage{marginnote}

\tcbset{
  commentstyle/.style={
    colback=yellow!20, % Background color
    colframe=orange!80!black, % Border color
    boxrule=0.8pt,
    sharp corners,
    fonttitle=\bfseries,
    width=4cm, % Width of the box in the margin
    halign=flush left,
    left=1mm,
    right=1mm,
    top=1mm,
    bottom=1mm,
  }
}

\title{PairBench: Are Vision-Language Models Reliable at Comparing What They See?}

\author{Aarash Feizi\textsuperscript{1,2,3}\hspace{2ex}
Sai Rajeswar\textsuperscript{1} \hspace{2ex}
Adriana Romero-Soriano\textsuperscript{2,3}\hspace{2ex}
Reihaneh Rabbany\textsuperscript{2,3}\hspace{2ex}\\
\textbf{Valentina Zantedeschi\textsuperscript{1}\hspace{2ex}
Spandana Gella\textsuperscript{1}\hspace{2ex}
João Monteiro\textsuperscript{4}}
\\
  \textsuperscript{1}ServiceNow Research\hspace{2ex}
  \textsuperscript{2}Mila -- Québec Artifical Intelligence Institute\\
  \textsuperscript{3}McGill University\hspace{2ex}
  \textsuperscript{4}Autodesk\\
  \texttt{aarash.feizi@servicenow.com, joao.monteiro@autodesk.com}
  }
\newcommand{\wu}{WhatsUp}
\newcommand{\coco}{COCO}
\newcommand{\imagenet}{IN100}

\newcommand{\mmscorecomparesize}{\textsc{70K}}

\newcommand{\model}{VLM}

\newcommand{\modelss}{VLMs}

\newcommand{\mmscore}{\textsc{PairBench}}
\newcommand{\mmscorebold}{\textsc{\textbf{PairBench}}}

\newcommand{\mmscoreshort}{\textsc{PB}}
\newcommand{\openrouter}{\textsc{OpenRouter}}

\newcommand{\mmscoreimgimg}{\mmscoreshort$_{\text{II}}$}
\newcommand{\mmscoreimgtext}{\mmscoreshort$_{\text{IT}}$}
\newcommand{\mmscorecoco}{\mmscoreshort$_{\text{COCO}}$}
\newcommand{\mmscorein}{\mmscoreshort$_{\text{IN100}}$}
\newcommand{\mmscorewuimgimg}{\mmscoreshort$_{\text{WU-II}}$}
\newcommand{\mmscorewuimgtext}{\mmscoreshort$_{\text{WU-IT}}$}

\newcommand{\nmibold}{\textbf{MMScore}}

\newcommand{\relaxsymbold}{\textbf{$\varepsilon$-RelaxSym}}

\newcommand{\nmi}{\emph{MMScore}}

\newcommand{\relaxsymone}{1-RelaxSym}

\newcommand{\relaxsym}{$\varepsilon$-RelaxSym}

\newcommand{\smoothness}{\emph{SM}}
\newcommand{\control}{\emph{Cont}}

\newcommand{\gptFouroFive}{\texttt{GPT-4o-0513}}
\newcommand{\gptFouroEight}{\texttt{GPT-4o-0806}}
\newcommand{\gptFouroEleven}{\texttt{GPT-4o-1120}}
\newcommand{\gptFouroMini}{\texttt{GPT-4o-mini-0718}}
\newcommand{\geminiFlash}{\texttt{Gemini-1.5-Flash}}
\newcommand{\geminiPro}{\texttt{Gemini-1.5-Pro}}

\newcommand{\llavaonevision}{\texttt{{LLaVA-OneVision-7B}}}

\newcommand{\internvlTwoFiveEightB}{\texttt{{InternVL2.5-8B}}}
\newcommand{\internvlTwoFiveFourB}{\texttt{{InternVL2.5-4B}}}
\newcommand{\internvlTwoFiveTwoB}{\texttt{{InternVL2.5-2B}}}
\newcommand{\internvlTwoFiveOneB}{\texttt{{InternVL2.5-1B}}}

\newcommand{\internvlTwoEightB}{\texttt{{InternVL2-8B}}}
\newcommand{\internvlTwoFourB}{\texttt{{InternVL2-4B}}}
\newcommand{\internvlTwoTwoB}{\texttt{{InternVL2-2B}}}
\newcommand{\internvlTwoOneB}{\texttt{{InternVL2-1B}}}

\newcommand{\qwenTwoVLTwoB}{\texttt{{Qwen2-VL-2B}}}
\newcommand{\qwenTwoVLSevenB}{\texttt{{Qwen2-VL-7B}}}

\newcommand{\chameleon}{\texttt{{Chameleon-7B}}}

\newcommand{\pixtral}{\texttt{{Pixtral-12B}}}
\newcommand{\phiThreeFive}{\texttt{{Phi-3.5-vision}}}

\newcommand{\molmoDSevenB}{\texttt{{Molmo-7B-D}}}
\newcommand{\molmoOSevenB}{\texttt{{Molmo-7B-O}}}
\newcommand{\molmoEOneB}{\texttt{{MolmoE-1B}}}

\newcommand{\dinoSmall}{\texttt{{DINOv2-Small}}}
\newcommand{\dinoBase}{\texttt{{DINOv2-Base}}}
\newcommand{\dinoLarge}{\texttt{{DINOv2-Large}}}

\begin{document}

\maketitle

\vspace{1ex}
\begin{abstract}
Understanding how effectively large vision language models (\modelss{}) compare visual inputs is crucial across numerous applications, yet this fundamental capability remains insufficiently assessed. While \modelss{} are increasingly deployed for tasks requiring comparative judgment, including automated evaluation, re-ranking, and retrieval-augmented generation, no systematic framework exists to measure their performance in these scenarios. We present \mmscorebold{}, a simple framework that evaluates \modelss{} as customizable similarity tools using widely available image datasets. Our approach introduces four key metrics for reliable comparison: alignment with human annotations, consistency across pair ordering, distribution smoothness, and controllability through prompting. Our analysis reveals that no model consistently excels across all metrics, with each demonstrating distinct strengths and weaknesses. Most concerning is the widespread inability of \modelss{} to maintain symmetric similarity scores. Interestingly, we demonstrate that performance on our benchmark strongly correlates with popular benchmarks used for more complex tasks, while providing additional metrics into controllability, smoothness and ordering. This makes \mmscorebold{} a unique and comprehensive framework to evaluate the performance of \modelss{} for automatic evaluation depending on the task.
% offering an efficient predictor of model capabilities for more complex tasks. 
Our benchmark can be accessed at \verb  |https://huggingface.co/datasets/feiziaarash/pairbench|.
\end{abstract}

% Understanding how effectively large vision language models (\modelss{}) compare visual inputs is crucial across numerous applications, yet this fundamental capability remains insufficiently assessed. While \modelss{} are increasingly deployed for tasks requiring comparative judgment, including automated evaluation, re-ranking, and retrieval-augmented generation, no systematic framework exists to measure their performance in these scenarios. We present \mmscorebold{}, a simple framework that evaluates \modelss{} as customizable similarity tools using widely available image datasets. Our approach introduces four key metrics for reliable comparison: alignment with human annotations, consistency across pair ordering, distribution smoothness, and controllability through prompting. Our analysis reveals that no model consistently excels across all metrics, with each demonstrating distinct strengths and weaknesses. Most concerning is the widespread inability of \modelss{} to maintain symmetric similarity scores. Interestingly, we demonstrate that performance on our benchmark strongly correlates with popular benchmarks used for more complex tasks, while providing additional metrics into controllability, smoothness and ordering. This makes \mmscorebold{} a unique and comprehensive benchmark for evaluating the performance of VLMs as similarity kernels.  
% %offering an efficient predictor of model capabilities for more complex tasks. 
% \textcolor{red}{We share the dataset here LINK TO HUGGINGFACE}

\vspace{-2pt}

% \addcomment{

% 9. Reproduceablitiy hyper parasm

% 10. Make CHECKLIST!!!

% 11. huggingface DONE?https://huggingface.co/datasets/feiziaarash/pairbench

% 12. github https://github.com/aarashfeizi/pairbench/tree/main

% 14. make sure single lines in pages are addressed

% 15. advertise refer to results regarding model-size (scaling test) vs. performance (e.g., as models get larger, they get more symmetric and more alignment, but maybe less smooth?)

% 16. Add huggingface link to end of abstract and/or footnote in intro (which has a pointer to the github repo)

% 17. Advertise all results in the appendix

% % 18. figure 5 is too big

% 19. Add a section/explanation as to why our benchmark should be used compared to other benchmarks. Why would someone need to use our benchmark instead of something that is already out there, e.g., MMMU?

% 20. MMScore -> Given that it highly correlates with other benchmarks, it validates that MMScore is an important metric. However, we do not want to replace the other benchmarks. It's just that our tasks "correlates". But if we want to look at them as similarity kernels, our benchmark is better. 

% }
\vspace{-3ex}
\section{Introduction}
\label{sec:main-intro}
Vision language models (\modelss{}) have progressed to the point of having impressive performance on a wide array of tasks \citep{achiam2023gpt,laurenccon2024matters, reid2024gemini, abdin2024phi, wang2024enhancing,llama3}, ranging from summarization and visual question answering to image captioning and common sense reasoning \citep{kembhavi2016diagram, johnson2017clevr, zellers2019recognition, lu2023mathvista, chen2024we, liu2025mmbench}. While human evaluation remains the gold standard for assessing model outputs, it is expensive, time-consuming, and prone to inconsistency due to annotator variance \citep{liu2019user, knox2024models, feng2024sample}. Consequently, practitioners increasingly deploy more powerful \modelss{} as automated evaluators across diverse applications including model assessment, content ranking, and information retrieval systems \citep{manas2024improving, liu2024visual, liu2025mmbench}.

\begin{figure}[ht]
    \centering
    \vspace{-2ex}\includegraphics[width=0.93\linewidth]{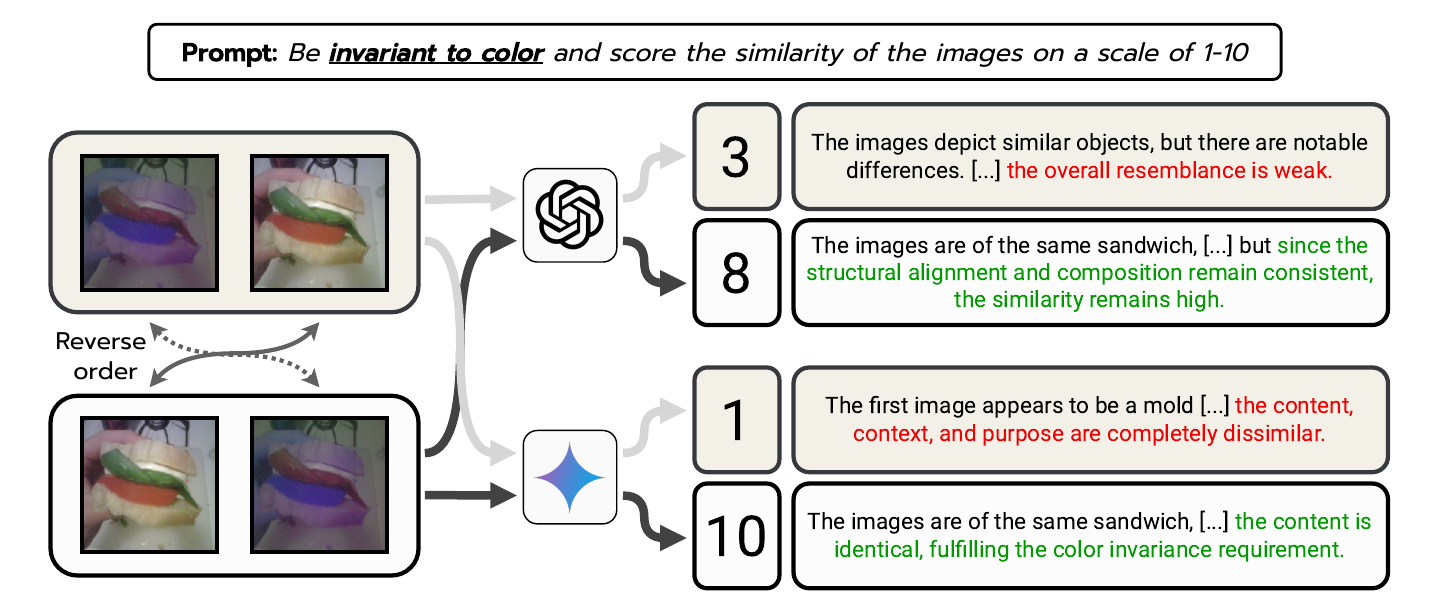}
    % \vspace{-0.5cm} % Adjust this value as necessary
    \vspace{-1ex}
    \caption{\small Image order change; prompting \gptFouroEleven{} and \geminiPro{} with identical text and image prompts, differing only in image order, leads to varying predicted scores. \emph{Auto evaluators defined by these models will yield drastically different judgments after minor changes in the prompt}. Detailed failure cases of state-of-the-art models are reported in Appendix \ref{sec:error-analysis}.\vspace{-2ex}}
    \label{fig:fig1}
\end{figure}

The efficacy of \modelss{} in these comparative tasks fundamentally depends on their ability to function as reliable similarity kernels, consistently measuring the relevance between data pairs regardless of context. However, this critical capability remains insufficiently examined. Current evaluation approaches either fail to isolate comparison abilities or require expensive expert validation, and little to no guidance exists when selecting models for comparison-dependent tasks. As illustrated in Figure \ref{fig:fig1}, even widely used and highly capable commercial models like \gptFouroEleven{} and \geminiPro{} demonstrate concerning inconsistencies when comparing visual inputs, sometimes failing to follow similarity assessment instructions or producing asymmetric scores for identical pairs presented in different orders, which exemplifies the extent to which evaluation of comparison skills are lacking.

To address this gap, we introduce \mmscore{}, a framework designed to evaluate \modelss{} as similarity estimators using readily available datasets and straightforward transformation techniques. Our approach optimizes the signal-to-evaluation cost ratio by focusing on four essential metrics: \nmibold{} (alignment with human judgment), \relaxsymbold{} (consistency across pair ordering), \textbf{Smoothness} (distribution of scores), and \textbf{Controllability} (response to prompt instructions). We instantiated \mmscore{} using easily accessible datasets. Namely, ImageNet~\citep{deng2009imagenet}, MS-COCO~\citep{lin2014microsoft}, and WhatsUp~\citep{kamath2023s} seeded the evaluation suite we built. We further conducted a human study to establish ground truth similarity scores that enable direct measurement of how well model assessments align with human perception. By applying controlled transformations to create synthetic paired images with specific feature differences, \mmscore{} enables precise examination of model biases and strengths in detecting various types of visual differences.

We carried out an extensive evaluation covering several state-of-the-art \modelss{}, both proprietary and open-source, multiple dataset configurations, and different prompt templates. Results reveal not only significant variations in comparison capabilities across different architectures and training approaches, but also show concerning asymmetries in how models process the same image pairs when presented in different orders, and highlight which models can be effectively controlled through prompt instructions. Remarkably, despite its simplicity, the performance on \mmscore{} strongly correlates with results on complex reasoning benchmarks \citep{yue2024mmmu, lu2023mathvista, chen2024we, guan2024hallusionbench, liu2024ocrbench, kembhavi2016diagram}, suggesting that many advanced tasks ultimately rely on models functioning as effective similarity kernels.

Our contributions are as follows:
% \vspace{-4mm}
\begin{itemize}
    \item We propose \mmscorebold{}, a framework for evaluating \modelss{} as similarity kernels,  which does not require additional expert annotations and is cheap to instantiate.
    \item We further create and release\footnote{https://huggingface.co/datasets/feiziaarash/pairbench} four instantiations of \mmscore{} using popular datasets - ImageNet, MS-COCO, and WhatsUp - which consist of \mmscorecomparesize{} data pairs for comparisons.
    \item We carry out a broad benchmarking of several closed- and open-source \modelss{} on the different configurations within our proposed dataset instantiations to show how models differ and give insight into to what extent they can be trusted to act as auto evaluators on image-image and image-text data pairs.
    \item Lastly, we report the correlations of the results of our framework with popular benchmarks and show the ability to compare, captured by the metrics in \mmscore{}, have predictive power of performance on several tasks and can act as a low-cost surrogate during training or validation of \modelss{}.
\end{itemize}

\begin{figure*}[ht]
    \centering
    \vspace{-2ex}    \includegraphics[width=0.95\linewidth,trim={.1cm .2cm .2cm .2cm},clip]{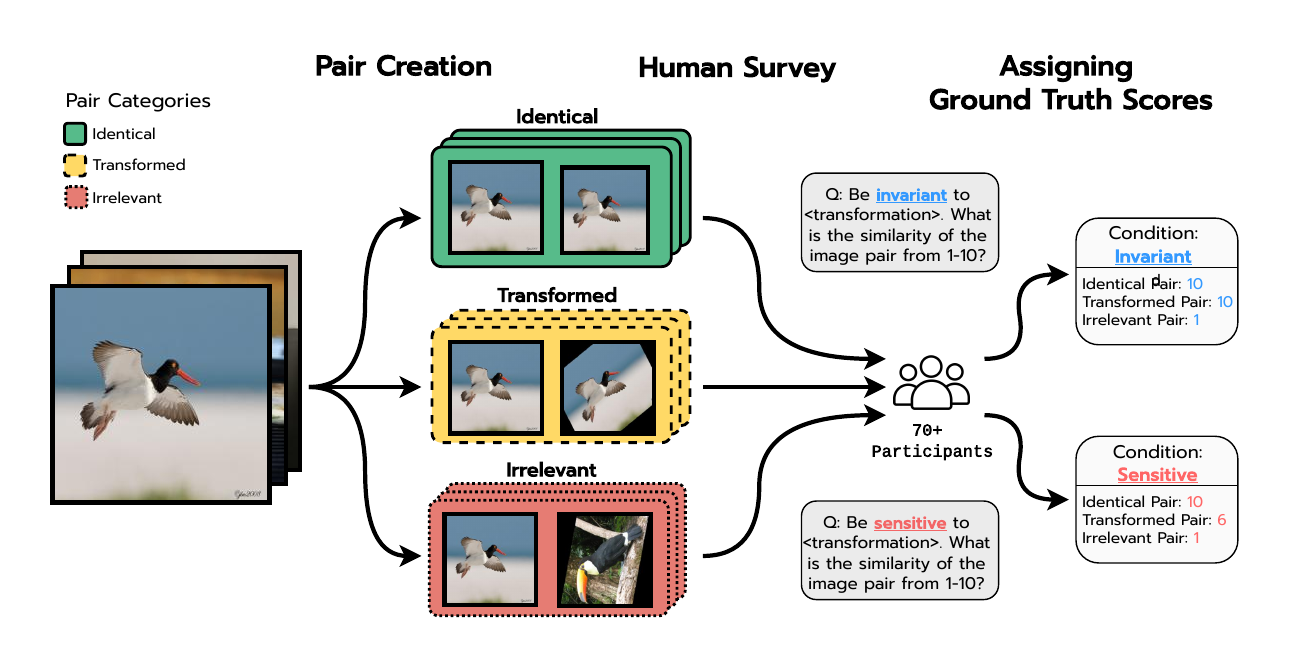}
    \caption{Dataset creation pipeline for the image-image datasets, i.e., \mmscorecoco, \mmscorein, and \mmscorewuimgimg. Each original image is used to create three pairs of image: identical, transformed, and irrelevant. Finally, based the human study, each pair is scores depending on the condition of the prompt.}
    \vspace{-2ex}
    \label{fig:dataset-creation-pipeline}
\end{figure*}
\vspace{-3mm}
\section{\mmscore{}}
\label{sec:main-pairbench}
\vspace{-1ex}
\subsection{Dataset Creation}

To evaluate how well vision-language models can assess similarity under controlled transformations, we construct a dataset using the \mmscore{} framework, illustrated in Figure~\ref{fig:dataset-creation-pipeline}. For each original image, we generate three types of pairs: (1) Identical pairs where the second image is a near-duplicate, (2) Transformed pairs where a specific transformation (e.g., color jitter or spatial shift) is applied, and (3) Irrelevant pairs with unrelated content. We then gather similarity judgments from over 70 human annotators under two distinct conditions: invariant, where models should ignore transformations and focus on semantic similarity, and sensitive, where models should penalize such transformations. These human ratings verify the assignment of ground-truth similarity scores: both identical and irrelevant pairs are assigned fixed values of 10 and 1, respectively, while transformed pairs receive a score of 10 under invariance and 6 under sensitivity.

The framework is instantiated across image-only datasets (\coco{}, \imagenet{}) and image-text datasets (\wu{}), using five standard image transformations plus a spatial position shift known to challenge \modelss{}. Full construction details, including transformation splits, prompt templates used to reduce linguistic bias, and details of the human study procedure, are provided in Appendix~\ref{sec:mmscore-info}.
\vspace{-1ex}
\subsection{Metrics}
To measure the reliability of \modelss{} in scoring data pairs, we define four metrics that we measure across datasets and models: \nmi, \relaxsym, Smoothness (\smoothness), and Controllability (\control). 

% We follow this notation to formulate the metrics: we denote the \model{} being evaluated as $\mathcal{M}$ and the prompt as $P_{C}$ where $C \in \left\{\texttt{sens}, \texttt{inv} \right\}$, as sensitive or invariant. Finally, given a dataset $\mathcal{D}_N = \{(d_1, d_2), (d_3, d_4), \dots, (d_{2N-1}, d_{2N}))\}$, we denote the similarity score of a data pair $(d_i, d_j) \in \mathcal{D}_N$ returned by an \model{} ($\mathcal{M}$) and for a given prompt ($C$) as:
% % $$s_{\mathcal{M}}^{C}(d_i, d_j) \vcentcolon= \mathcal{M}(C, d_i, d_j),$$
% $$s_{\mathcal{M}}^{C}(d_i, d_j) \vcentcolon= \mathcal{M}(C, d_i, d_j),$$

We adopt the following notation to formulate the metrics: we denote the \model{} being evaluated as $\mathcal{M}$ and the condition, which determines if the prompt instructs the model to be sensitive or invariant to a visual feature, as $C \in \left\{\texttt{sens}, \texttt{inv} \right\}$. Finally, given a dataset $\mathcal{D}_N = \{(d_1, d_2), (d_3, d_4), \dots, (d_{2N-1}, d_{2N}))\}$, we denote the similarity score of a data pair $(d_i, d_j) \in \mathcal{D}_N$ returned by an \model{} ($\mathcal{M}$) for a given condition ($C$) as:
% $$s_{\mathcal{M}}^{C}(d_i, d_j) \vcentcolon= \mathcal{M}(C, d_i, d_j),$$
$$s_{\mathcal{M}}^{C}(d_i, d_j) \vcentcolon= \mathcal{M}(C, d_i, d_j),$$
where $(d_i, d_j)$ could be an image-image or image-text pair. Note that we instruct the model to generate the output in a structured format to make sure the predicted score is parsable from the model output. If $s_{\mathcal{M}}^{C}(d_i, d_j)$ is valid, it would fall in the set
% $\mathcal{V} = [1, 2, 3, \dots, 10\}$. 
$\mathcal{V} = [1, 10]$. 
However, models often do not consistently follow the details of the prompt and may produce scores not in $\mathcal{V}$ or outputs that do not satisfy the output format, in which case we set $s_{\mathcal{M}}^{C}(d_i, d_j) = -1$. 
Finally, to evaluate a model $\mathcal{M}$ on $\mathcal{D}_N$ given condition $C$, we create and annotate the set of all its outputs as:
$$S_{\mathcal{M}}^{C}(\mathcal{D}_N) = \left\{ s_{\mathcal{M}}^{C}(d_i, d_j) \,\middle|\, (d_i, d_j) \in \mathcal{D}_N \cup \text{rev}(\mathcal{D}_N) \right\},
$$
\vspace{-1ex}
where $\text{rev}(\mathcal{D}_N) = \{(d_2, d_1), (d_4, d_3), \dots, (d_{2N}, d_{2N-1}))\}$ are the data pairs in reverse order.

\subsubsection{MMScore}
We first introduce \nmi{}, the main metric of \mmscore{}, which measures the alignment between model predictions and human assessments. To this aim, we compute it as the normalized mutual information between the predicted and the ground-truth scores. Instead of accuracy or squared error, we consider \nmi{} since we do not explicitly prompt the \model{} with examples of the correct scores and hence cannot expect it to predict them directly. \nmi{} is better suited for \mmscore{}, as it focuses on whether the \model{}'s scores are predictive of the ground-truth ones without penalizing outputs that do not exactly match them.
The better a model can reproduce the variance in the ground-truth score, the better it is able to recognize that characteristic. Hence we write,
% \vspace{-1mm}
\[
\nmi(\mathcal{M}, C, \mathcal{D}_N) = \text{NMI}(S_{\mathcal{M}}^{C}(\mathcal{D}_N), GT_C(\mathcal{D}_N)),
\]

where $\text{NMI}(.,.)$ is the normalized mutual information and $GT_C(.)$ is the ground truth of the input dataset considering the condition of $C$.

% \subsubsection{\softsymmetrytitle (\normsym)}
\vspace{-1mm}
\subsubsection{\relaxsym{}}

The second metric we introduce aims to evaluate how consistent models are with respect to input order. This metric captures a fundamental characteristic when \modelss{} are used as re-rankers or automatic evaluators. Surprisingly, however, we found that most models do not satisfy exact symmetry, i.e., the equality of $sim(a, b)$ and $sim(b, a)$. We thus introduce \relaxsym{}, which tolerates a difference of $\varepsilon$ between the scores that should be equal.
More specifically, to analyze the symmetry of \modelss{} on a dataset $\mathcal{D}_N$, we compute the \relaxsym{} of $\mathcal(M)$ on $\mathcal{D_N}$:

% $$\text{\normsym}(\mathcal{M}, \mathcal{D}_N) = 1 - \frac{1}{N}\sum_{(d_i, d_j) \in \mathcal{D}_N}\text{Diff}(\mathcal{M}, d_i, d_j),$$
$$
\text{\relaxsym}(\mathcal{M}, \mathcal{D}_N) = \frac{1}{N}\sum_{(d_i, d_j) \in \mathcal{D}_N}\text{SoftEq}_\varepsilon(\mathcal{M}, d_i, d_j),
$$

where $\text{SoftEq}_\varepsilon(\mathcal{M}, d_i, d_j)$ is defined as:
$$
% \text{Diff}(\mathcal{M}, d_i, d_j)=
\text{SoftEq}_\varepsilon(\mathcal{M}, d_i, d_j)=
\begin{cases} 
    % \frac{\lvert s_{\mathcal{M}}^{P_.}(d_i, d_j) - s_{\mathcal{M}}^{P_.}(d_j, d_i)\rvert}{9}, & s_{\mathcal{M}}^{P_.}(d_i, d_j), s_{\mathcal{M}}^{P_.}(d_j, d_i) \in \mathcal{V}, \\
    % 1, & \text{otherwise}.
    \mathbbm{1}(\lvert s_{\mathcal{M}}^{C}(d_i, d_j)-s_{\mathcal{M}}^{C}(d_j, d_i) \le \varepsilon \rvert), & s_{\mathcal{M}}^{C}(d_i, d_j), s_{\mathcal{M}}^{C}(d_j, d_i) \in \mathcal{V}, \\
    0, & \text{otherwise}.
\end{cases}
$$

In the continuation of this paper, we set $\varepsilon = 1$ and provide ablation studies in Figure \ref{fig:diff-relax-sym-eps} in the Appendix.

\vspace{-2mm}
\subsubsection{\textbf{Smoothness}}

We aim to measure how smooth the kernels induced by \modelss{} are. For instance, a non-smooth kernel would assign scores such that pairs are either exactly the same or completely different, while a smoother kernel produces more nuanced distinctions. We measure smoothness via the diversity of the predicted scores. Given $S^{C}_{\mathcal{M}}$, smoothness (\smoothness) is computed as:
% \begin{align}
$$\smoothness(\mathcal{M}, \mathcal{D}_N, C) = Ent(\left\{s \,\middle|\, s \in S_{\mathcal{M}}^{C}(\mathcal{D}_N)  \,\text{and}\, s \in \mathcal{V}\right\}),$$
% \end{align}

where $Ent(.)$ is the entropy of a set relative to its support, i.e., the set of candidate inputs.

\subsubsection{\textbf{Controllability}}

We measure how \textbf{responsive to instructions} models are. To do so, we define controllability based on the difference in \nmi{} between the sensitive and invariant settings. The more controllable a model is, the smaller the discrepancy observed between the \texttt{sens} and \texttt{invar} settings. Hence, when measuring the controllability on $\mathcal{D}_N$ for a model $\mathcal{M}$ is defined as
\vspace{-2mm}

$$ \control({\mathcal{M}, \mathcal{D}_N}) = 1 - \frac{|\nmi(\mathcal{M}, \texttt{sens}, \mathcal{D}_N) - \nmi(\mathcal{M}, \texttt{inv}, \mathcal{D}_N)|}{\sqrt{(\nmi(\mathcal{M}, \texttt{sens}, \mathcal{D}_N) \times \nmi(\mathcal{M}, \texttt{inv}, \mathcal{D}_N))}}.$$

% $$ C_{\mathcal{M}, f} = 1 - \frac{|\mathcal{M}_{sens}(f) - \mathcal{M}_{invar}(f)|}{max(\mathcal{M}_{sens}(f), \mathcal{M}_{invar}(f))}$$

\begin{figure}[ht]
    \centering
    \includegraphics[width=0.95\linewidth,trim={.1cm .2cm .2cm .2cm},clip]{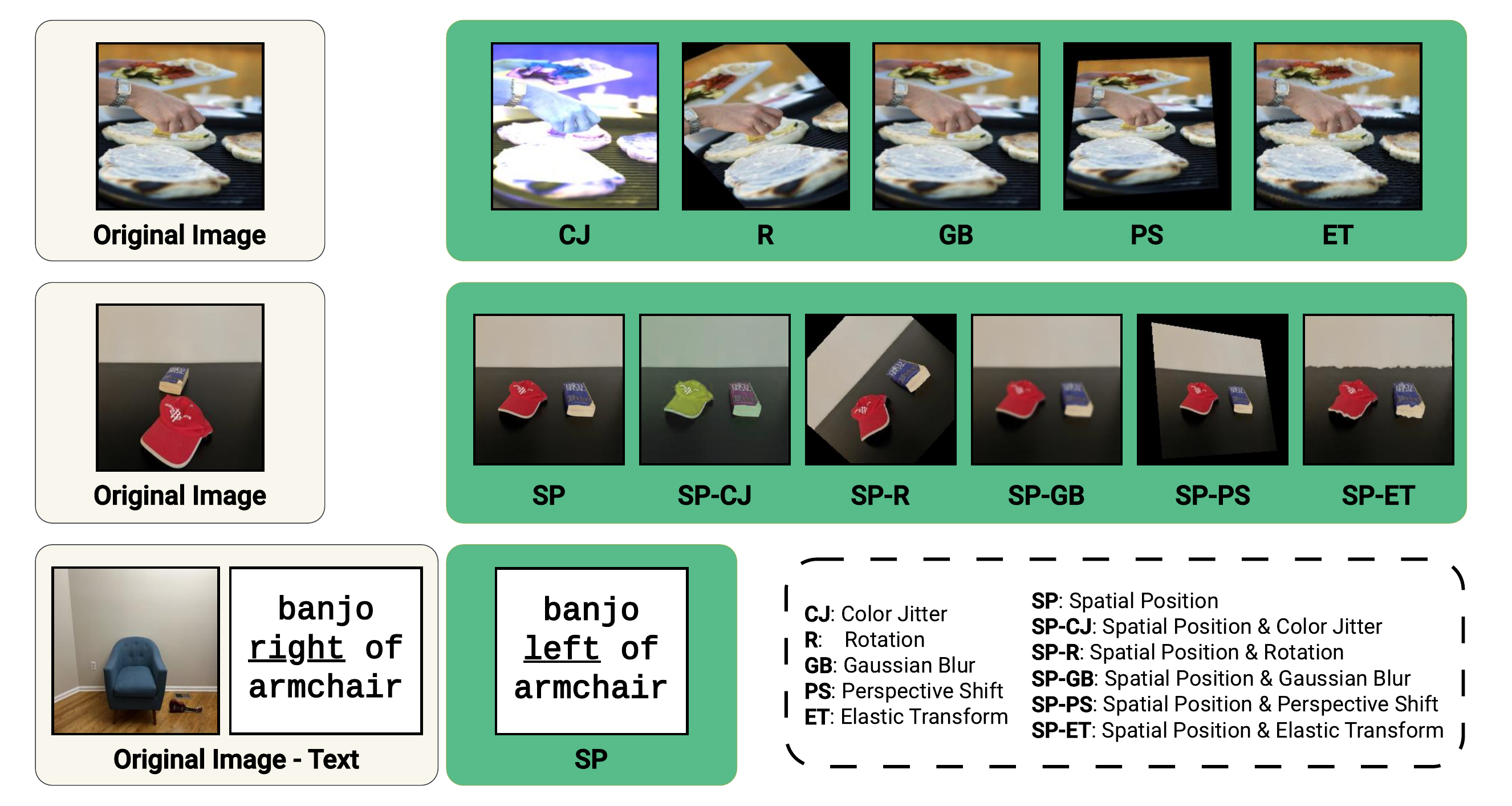}
    \caption{Examples of transformations (green boxes) applied to the original data points (gray boxes) of each subset instantiated with \mmscore. The first row shows the different splits of \mmscorecoco{} and \mmscorein, the second row for \mmscorewuimgimg, and the third for \mmscorewuimgtext.}
    \label{fig:mmscore-examples}
\end{figure}

\begin{figure*}[ht]
    \centering
\includegraphics[width=0.93\linewidth,trim={.1cm .2cm .2cm .2cm},clip]{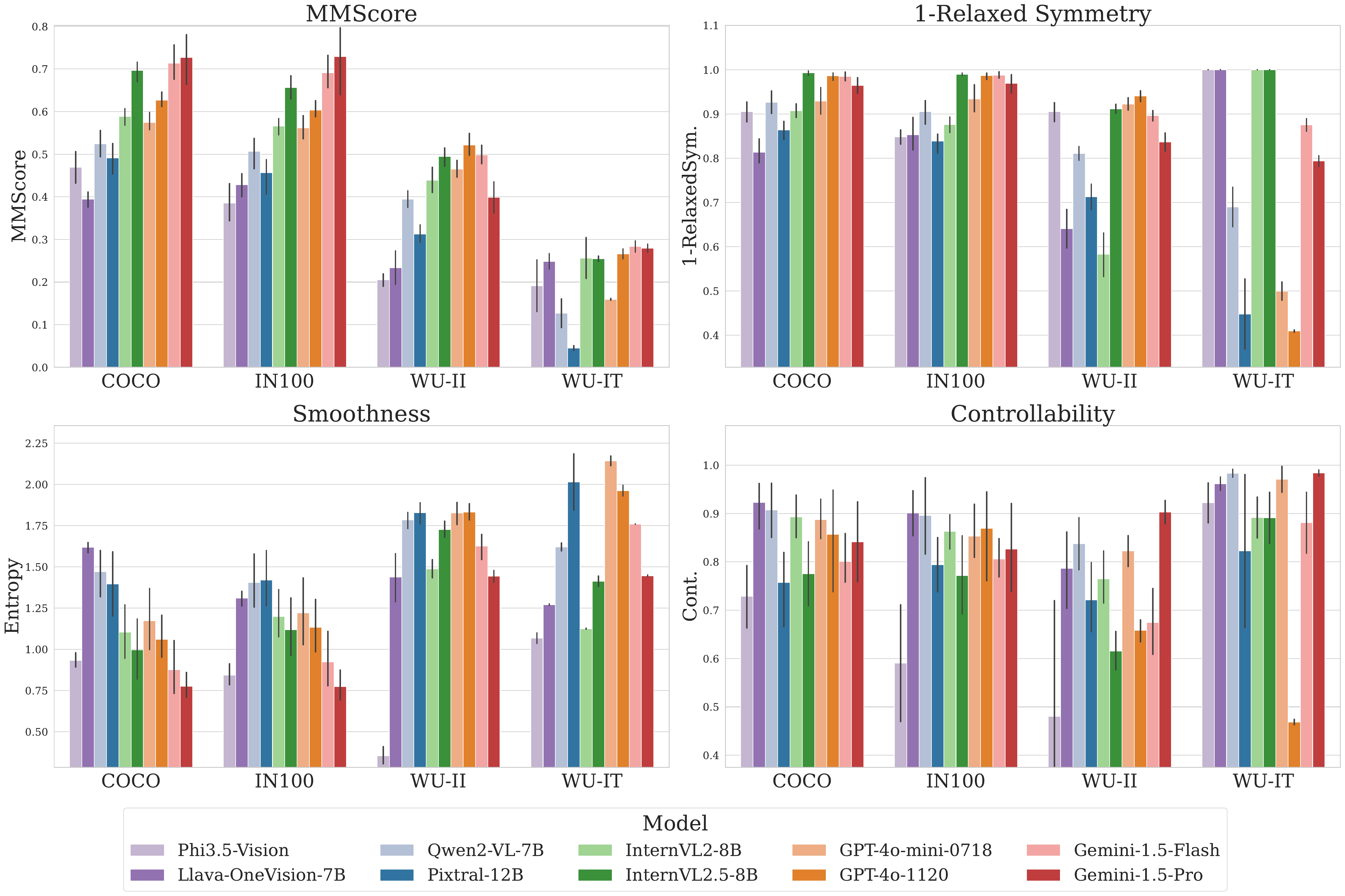}
    \caption{\small Best models performances on \mmscorecoco{}, \mmscorein{}, \mmscorewuimgimg, and \mmscorewuimgtext. No model dominates the others as a similarity kernel, hence showing the limitation of defaulting to a single model as a judge for every task and dataset. Note the full symmetry of \phiThreeFive, \llavaonevision, and InternVL models on \mmscorewuimgtext{} are due to the lack of flexibility in the prompt structure to take the image anywhere but the beginning.} 
    % \vspace{-3mm}
    % \textcolor{red}{Spandana: change the range on y-axis of each individual plot, so its clear the difference between the models. Also why is pixtral-12B so low on MM-Score?}}
    \label{fig:mi-best-models}
\end{figure*}

\section{Evaluation Results}
\subsection{Experimental Setting}
\label{sec:main-results}
% [We use a set of opensource and closed source models.]
We choose a comprehensive set of open- and closed-source vision-language models and evaluate them using the instantiations of \mmscore. From the openly available models, we evaluated \chameleon{} \citep{lu2024chameleon}, \llavaonevision{} \citep{li2024llava}, \pixtral{} \citep{agrawal2024pixtral}, \phiThreeFive{} \citep{abdin2024phi}, four model sizes (1B, 2B, 4B, and 8B) of InternVL2 \citep{wang2024enhancing},
% \internvlTwoEightB{}, \internvlTwoFourB{}, \internvlTwoTwoB{}, \internvlTwoOneB{}, 
four model sizes (1B, 2B, 4B, and 8B) of InternVL2.5 \citep{chen2024expanding},
% \internvlTwoFiveEightB{}, \internvlTwoFiveFourB{}, \internvlTwoFiveTwoB{}, \internvlTwoFiveOneB{}, 
% \vspace{-2mm}
two capacities (2B and 7B) of Qwen2-VL \citep{wang2024qwen2}, 
 % \qwenTwoVLTwoB{}, \qwenTwoVLSevenB{}, %and
% \vspace{-2mm}
and three versions (\molmoEOneB, \molmoOSevenB, and \molmoDSevenB) of Molmo \citep{deitke2024molmo}.
% \molmoEOneB{}, \molmoOSevenB{}, \molmoDSevenB{}. 

We also considered commercial grade models and benchmarked three versions of GPT-4o \citep{achiam2023gpt}(\gptFouroFive{}, \gptFouroEight{}, \gptFouroEleven{}), \gptFouroMini{}, and two versions of Gemini-1.5 \citep{reid2024gemini} (\geminiFlash{}, \geminiPro{}). 
Note that we consider multiple versions of the same architecture, as opposed to using the newest/largest version, to understand better how model capacity affects each of the metrics. We provide an extended analysis of different model versions in Appendix \ref{sec:model-versions}. 

We run all open-source models on a single NVIDIA H100 GPU using greedy sampling for inference. For closed-source models, we use API access through either \openrouter\footnote{\url{https://openrouter.ai/}} or OpenAI\footnote{\url{https://platform.openai.com/}}, applying the default inference hyperparameters provided by the respective platforms.

Also note that, since \mmscore{} aims to evaluate \modelss{} as similarity kernels on image-only or text-image pairs, we do not evaluate text-only reasoning models such as OpenAI-o1 or DeepSeek-R1 \citep{guo2025deepseek}. Further, we do not evaluate \texttt{Llama3.2-11B} \citep{llama3} as its official implementation on HuggingFace\footnote{\url{https://huggingface.co/}} does not support Flash Attention \citep{dao2022flashattention} and inference was prohibitively slow. As a result, we excluded them from our final results.

\begin{table}[ht]
    \centering
    \begin{minipage}{0.50\textwidth}
        \centering
        \tabcolsep 2pt
        \caption{Aggregated \nmi, \emph{1-RS}:\relaxsymone, \smoothness, and \control{} over all four data splits. No model performs the best across all metrics, showing the importance of \mmscore{} to rank models based on different abilities.}
        \label{tab:model_performance}
        \resizebox{\textwidth}{!}{%
            \begin{tabular}{l c c c c}
                \toprule
                Model & \nmi (\%) & \emph{1-RS}(\%) & \smoothness & \control (\%) \\
                \midrule
                \phiThreeFive          & 29.65 & 90.13 & 0.64 & 59.18 \\
                \llavaonevision        & 30.88 & 75.07 & 1.44 & 85.34 \\
                \qwenTwoVLSevenB       & 42.27 & 84.45 & 1.63 & 87.63 \\
                \internvlTwoEightB     & 48.13 & 74.63 & 1.32 & 82.27 \\
                \internvlTwoFiveEightB & 55.05 & \textbf{95.21} & 1.42 & 70.42 \\
                \pixtral               & 35.77 & 74.85 & \textbf{1.67} & 75.23 \\
                \midrule
                \gptFouroMini          & 48.28 & 89.07 & 1.59 & 85.48 \\
                \gptFouroEleven        & 53.95 & 91.53 & 1.54 & 72.77 \\
                \geminiFlash           & \textbf{56.55} & 93.19 & 1.34 & 74.54 \\
                \geminiPro             & 52.60 & 88.72 & 1.17 & \textbf{88.09} \\
                \bottomrule
            \end{tabular}
        }
    \end{minipage}
    \hfill
    \begin{minipage}{0.45\textwidth}
        \centering
        \tabcolsep 2pt
        \caption{Spearman correlation of different metrics of \mmscore{} with performance on other benchmarks for 23 models. \nmi{} has the highest correlation, making it the main metric.}
        \label{tab:benchmark_comparison}
        \resizebox{\textwidth}{!}{%
            \begin{tabular}{lcccc}
                \toprule
                Metric & \nmi & \emph{1-RS} & \smoothness & \control \\
                \midrule
                AI2D              & 76\% & 28\% & 30\% & 54\% \\
                HallusionBench    & 75\% & \textbf{43\%} & 31\% & 39\% \\
                MMBench           & 81\% & 25\% & \textbf{44\%} & 63\% \\
                MMMU              & \textbf{89\%} & 35\% & 31\% & 60\% \\
                MMStar            & 81\% & 20\% & 42\% & 58\% \\
                MMVet             & 79\% & 34\% & \textbf{44\%} & 51\% \\
                MathVista         & 73\% & 11\% & 41\% & \textbf{68\%} \\
                OCRBench          & 50\% & 10\% & 41\% & 35\% \\
                \bottomrule
            \end{tabular}
        }
    \end{minipage}
    \vspace{-3mm}
\end{table}

\begin{figure*}[ht]
    \centering
    \vspace{-2ex}
    \includegraphics[width=0.95\linewidth,trim={.1cm .2cm .2cm .2cm},clip]{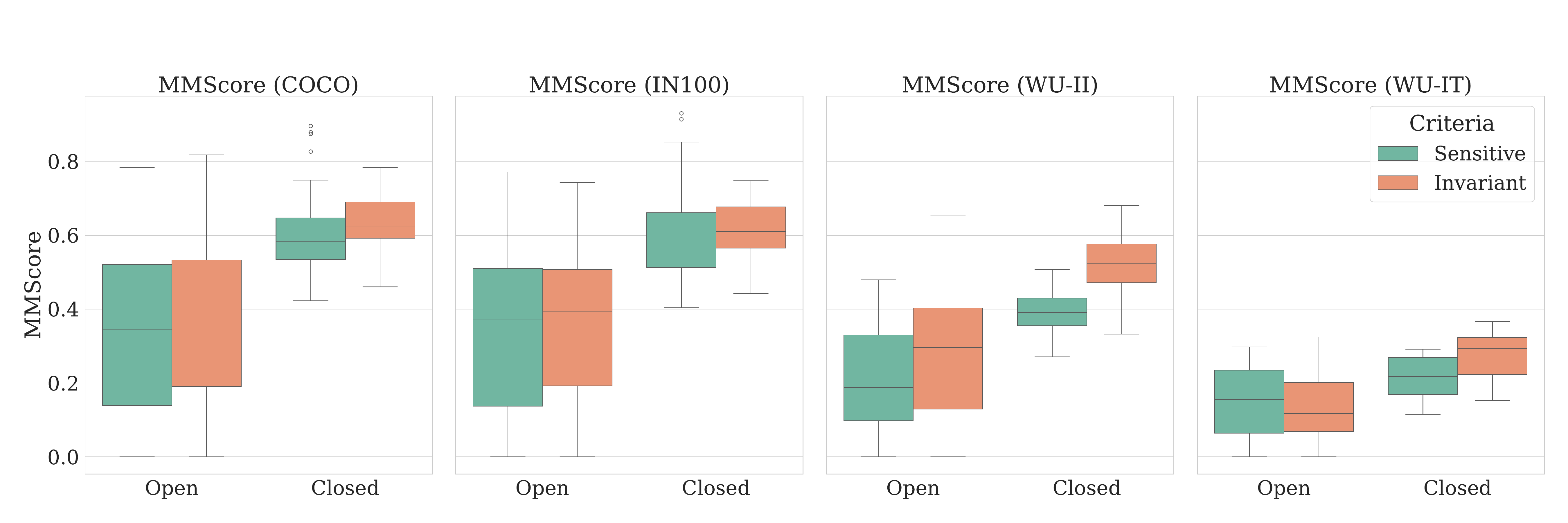}
    \caption{Closed- and open-source models perform comparable on image-text tasks. From the left to the right, the first three plots are image-image tasks, while the last is an image-text comparison task. }
    \vspace{-3ex}
    \label{fig:mi-close-vs-open}
    
    % \vspace{1ex}

    % \includegraphics[width=0.92\linewidth,trim={.1cm .2cm .2cm .2cm},clip]{imgs/Encoders-vs.-LMMs-colorjitter-elastic-gaussianblur-perspective-rotate-mi-plt.pdf}
    
    % \caption{A simple vision encoder outperforms open-sourced \modelss{} and has on par performance with closed sourced models which are much more expensive, for image-image tasks (results combine \mmscorecoco{} and \mmscorein), and similar pattern is observed across different transformations.}
    % \label{fig:mi-encoder-vs-lmms}
\end{figure*}
\subsection{Results}
\vspace{-1mm}
We analyze and plot the results of the best models in Figure \ref{fig:mi-best-models} and provide an aggregated version of the metrics over all four datasets in Table \ref{tab:model_performance}. We aggregate different splits/datasets by taking the average of them to give each sub-dataset equal importance in the final number. The full set of benchmarking results of all models for \mmscore{} on all datasets and metrics are reported in Appendix \ref{sec:full-results}.

% \vspace{-2mm}
\vspace{-1ex}
\subsubsection{General Observations}
As illustrated in Figure \ref{fig:mi-best-models} and Table \ref{tab:model_performance}, we observe no model, whether closed- or open-source, is the best performer across all four metrics. Moreover, we further observe that for each metric, no model is the `best' similarity kernel across the four different datasets either. This shows how features of the dataset and also the metrics a user might want to optimize play a crucial role in which \model{} to choose as the best similarity kernel/judge. For instance, among open-source models, although \internvlTwoFiveEightB{} outperforms the rest in \nmi, it is less controllable and smooth than \qwenTwoVLSevenB{} or \llavaonevision{}.
% in case symmetry is an important factor when comparing image-text data pairs, among closed-source models using Gemini models are preferred over GPT4o.

When considering \mmscore's main metric, \nmi, we notice that the performance of models is generally better on image-image pairs rather than image-text pairs. Furthermore, as seen in Figure \ref{fig:mi-close-vs-open}, we observe that although open-source \modelss{} are roughly comparable to closed-source ones on \mmscorewuimgtext, the gap between the two groups is larger in the image-image pairs. However, \internvlTwoFiveEightB{} is a strong competitor to closed-source models considering all four metrics and could potentially be used as a substitute to closed-source models as a similarity kernel based on the results reported in Table \ref{tab:model_performance}.

Interestingly, we further observe a pattern regarding \gptFouroEleven{}, a common default judge used in the literature, and its lower cost version, \gptFouroMini{}; they both suffer from low \relaxsymone{} when comparing image-text pairs, and the cheaper model's \control{} and \smoothness{} is higher or comparable to that of the expensive one across datasets. Another fascinating result we observed was the effect the scaling effect on different metrics of \mmscore{} for a single model family; the larger a model gets, the better it performs on \nmi{} and \relaxsymone{}. However, that does not hold for controllability and smoothness.
This emphasizes the importance of \mmscore{} in analyzing the capabilities of models, both open and closed-source, as similarity kernels to be better used as judges. We analyze and plot these results further in Appendix \ref{sec:full-results} and further have qualitative examples of the errors the best \modelss{} make in these tasks in Appendix \ref{sec:error-analysis}.~\looseness-1

\vspace{-2mm}
\subsubsection{Correlation with Benchmarks}

To showcase the effectiveness of our metrics with \mmscore{} in predicting model performance, we compute the Spearman correlation with respect to other popular benchmarks used in the literature. By showing correlations of our metrics with these benchmarks, we show that although the \mmscore{} framework introduces simple and cost-efficient methods focused on evaluating the ability to compare due to prompted \modelss{}, these metrics are predictive of an \model{}'s performance on other tasks.

We collect all the model performances from the \textsc{OpenVLM Leaderboard}\citep{duan2024vlmevalkit} and filter out the models we evaluate, resulting in all 23  (including different versions/capacities of closed- and open-source) models. By filtering out the benchmarks that have evaluation scores for all 23 models on OpenVLM, we end up with AI2D \citep{kembhavi2016diagram}, HallusionBench \citep{guan2024hallusionbench}, MMBench \citep{liu2025mmbench}, MMStar \citep{chen2024we}, MMMU \citep{yue2024mmmu}, MathVista \citep{lu2023mathvista}, MM-Vet \citep{yu2023mm}, OCRBench \citep{liu2024ocrbench}. Each metric is aggregated for each model across all the configurations created by \mmscore before computing correlations. Namely, we aggregate all features within each dataset (e.g., CJ, SP, etc.) and across all datasets (e.g., \mmscorecoco, \mmscorewuimgimg) and end up with an aggregate result per metric for each model.

\begin{figure*}[ht]
    \centering
    \includegraphics[width=0.90\linewidth,trim={.1cm .2cm .2cm .2cm},clip]{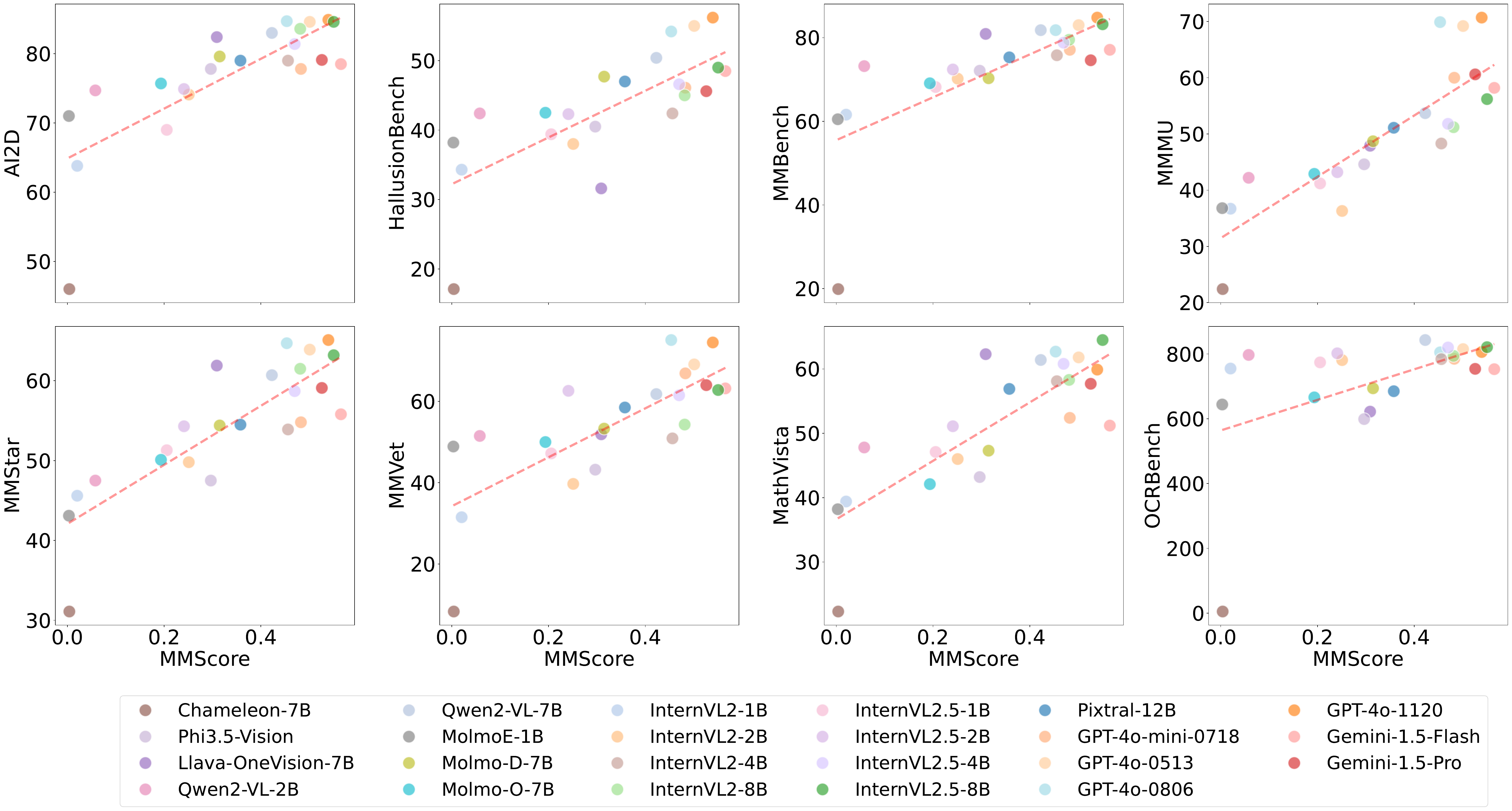}
    \caption{The main metric of \mmscore, \nmi{}, strongly correlates with previous multimodal benchmarks, showcasing its predictive power of a model's performance at a lower cost to create.}
    \vspace{-4mm}
    \label{fig:mmscore-vs-bms-nmi}
\end{figure*}

As seen in Table \ref{tab:benchmark_comparison}, all metrics in \mmscore{} have a high positive correlation with performances in benchmarks. More specifically, we observe that \nmi{} has strong correlations with all benchmarks, indicating that it aligns closely with overall model performance. Hence, we select it as the main metric of \mmscore{}. Furthermore, when analyzing the correlations of \mmscore{}'s other metrics with all benchmarks, we find that the strength of correlation reflects how much of the base skill captured by the metric is required by each benchmark. For example, HallusionBench shows the highest correlation with \relaxsymone{}, which is notable since HallusionBench focuses primarily on evaluating hallucinations in \modelss{}. This suggests a connection between lack of symmetry and hallucination. Another example is the highest and lowest correlation of \control{} with MathVista and OCRBench, respectively. Among the studied benchmarks, MathVista contains the most complex and verbose textual prompts, whereas OCRBench features simple prompts for most questions. We hypothesize that since \control{} measures how well models follow the prompt, these differences explain the highest and lowest correlations observed with MathVista and OCRBench. % Note that since \nmi{} has the highest significant correlation, we choose it to be the main metric of \mmscore. 

Note that measuring comparison skills incurs a low cost as it does not require expert-generated annotations. Our results suggest that metrics that assess these skills can serve as a low-cost surrogate of performance in various tasks: an efficient alternative to model selection. We further show scatter plots that highlight correlations in Figure~\ref{fig:mmscore-vs-bms-nmi}, and more comprehensively in Figure~\ref{fig:sym-vs-bm} in Appendix \ref{sec:full-results}.

\vspace{-1ex}
\subsubsection{Prompt Selection}
\begin{figure*}[ht]
    \centering
    \includegraphics[width=0.75\linewidth,trim={.1cm .2cm .2cm .2cm},clip]{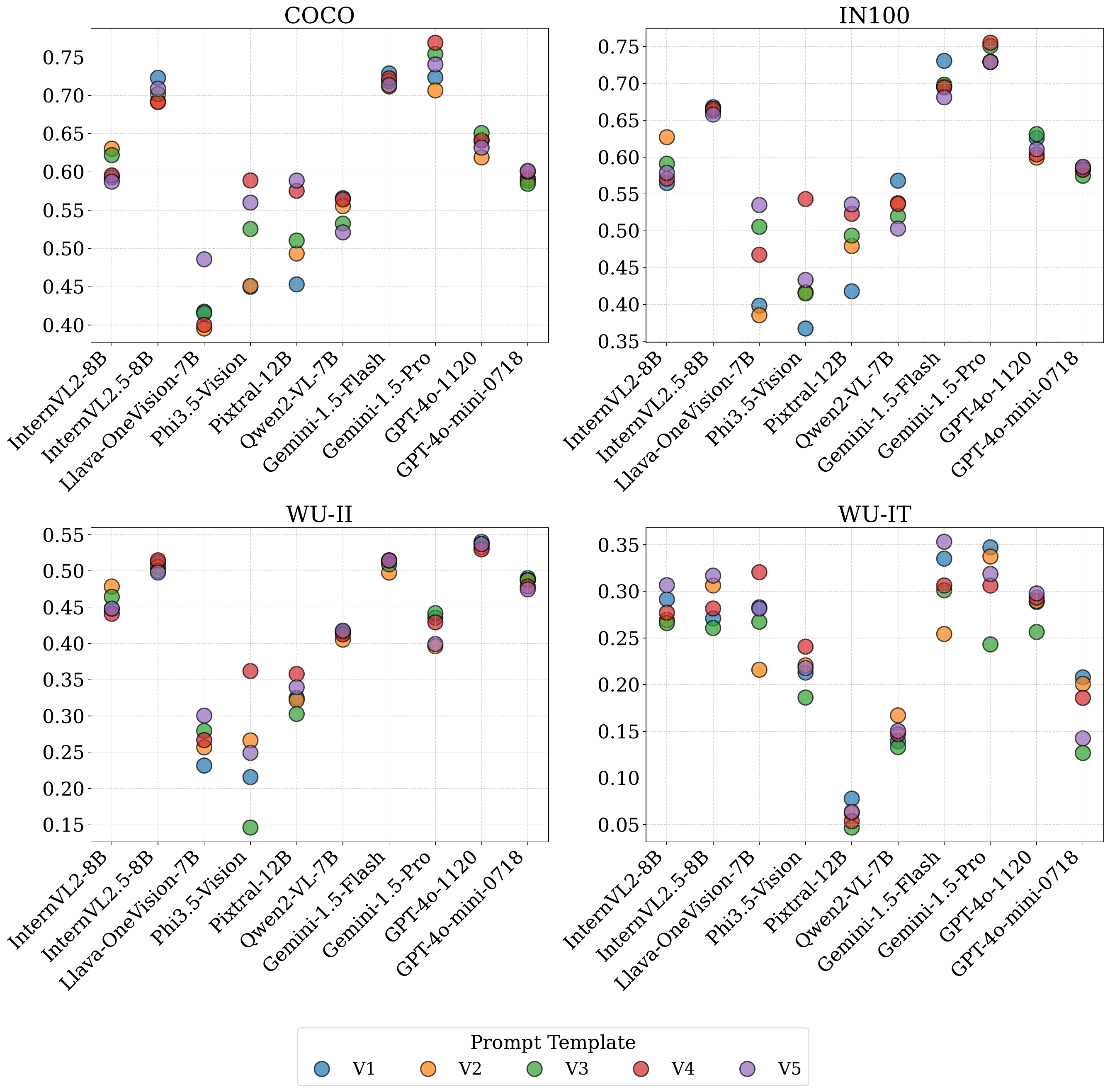}
    \caption{By using multiple prompt templates, we ensure no model is biased towards a single prompt and the mean capture the overall model performance.}
    % \vspace{-4mm}
    \label{fig:mmscore-across-prompts}
\end{figure*}

We show the \nmi{} performance of models across datasets in Figure~\ref{fig:mmscore-across-prompts} for various prompt templates. As evident from the plots, no single prompt template consistently achieves the best performance across all models. Some models perform better with certain phrasings, while others are negatively affected by the same templates. This variation highlights the significant influence of prompt wording on model behavior. Recent work~\citep{NEURIPS2024_28236482} has emphasized the importance of using diverse prompts. Similarly, our randomized approach offers empirical support for that recommendation. Evaluating models with multiple prompt templates and averaging the results eliminates prompt-induced variance and leads to more reliable and fair comparisons. We recommend this strategy as a stronger and more principled standard for future benchmarking of prompted models, whether multimodal or otherwise.
% \textcolor{red}{HAVE SOME STRONG SENTENCES MAKING CLAIMS, this is a easy way to address the issue raised in the literature \citep{NEURIPS2024_28236482} }
% We show the \nmi{} performance of models for each dataset in Figure \ref{fig:mmscore-across-prompts}. As seen, a single prompt template does not show the best performance across models and opting multiple prompt templates and averaging over them reduces prompt-related biases.

\vspace{-2ex}
\section{Related Work}
\vspace{-1ex}
Recent work has explored using language models as automated evaluators in NLP and vision-language domains, with approaches like \textsc{GPTScore} and G-eval~\citep{fu2023gptscore, liu2023geval} showing alignment with human preferences. However, concerns remain regarding their reliability, especially due to known limitations such as sensitivity to input order~\citep{fang2024rethinking} and failure to infer reversible relationships~\citep{berglund2023reversal}. In the multimodal case, work such as~\citet{zheng2023judging, thakur2024judging, murugadoss2024evaluating} evaluates \modelss{} as judges, highlighting issues of bias, prompt dependency, and limited control over evaluation criteria. Our work extends this line of research by focusing on structured pairwise comparisons, measuring not just performance alignment but also properties like symmetry, smoothness, and controllability.

While benchmarks like~\citet{chen2024mllm} and~\citet{awal2024vismin} introduce ways to test comparison abilities of VLMs, we aim to provide a more systematic and transformation-aware framework. Prior work also identifies well-known blind spots in discriminative models such as CLIP, including spatial reasoning failures~\citep{kamath2023s} and neglect of logical constructs like negation~\citep{alhamoud2025vision}. Our goal is to support the development and evaluation of models in these areas through a carefully designed testbed. A detailed review of related evaluation benchmarks, model limitations, and pair-comparison studies is included in Appendix~\ref{sec:related-work-full}.
\vspace{-2ex}
\section{Conclusion and Future Work}
\label{sec:main-conclusion}
\vspace{-2ex}

We introduced \mmscore{}, a framework that systematically evaluates the fundamental yet overlooked ability of \modelss{} to compare visual inputs, a capability critical for applications ranging from automated evaluation and re-ranking to retrieval-augmented generation. By focusing on four key metrics (alignment with human judgment, symmetry, smoothness, and controllability), \mmscore{} provides comprehensive insights into how models process comparative information while intentionally minimizing computational requirements. This cost-efficiency addresses growing concerns around the unsustainable costs of model evaluation, which increasingly constitutes a significant portion of model development budgets \citep{polo2024tinybenchmarks,pacchiardi2024100,yuan2025beyond}.

Our extensive benchmarking revealed that no model excels across all metrics, with even leading commercial systems demonstrating concerning asymmetries when comparing identical pairs in different orders. Particularly noteworthy is our finding that performance on \mmscore{} strongly correlates with results on complex reasoning benchmarks, suggesting that \emph{comparison capabilities may constitute a fundamental skill} that underlies performance across diverse tasks. This insight offers a more efficient path to model selection and validation without the computational burden of exhaustive evaluations on large-scale benchmarks. As a means to further improve evaluation efficiency while accounting for sensitivity to prompting, we applied a randomized prompting strategy, rendering comparisons across models more reliable at no additional inference cost.

Looking forward, we hypothesize that tailored post-training approaches focused on improving comparative skills and better model classes may enhance overall capabilities across diverse tasks, given the transferability our results revealed. Future research could explore architectural modifications or specialized fine-tuning techniques that optimize for these metrics, creating more reliable \modelss{} for critical evaluation tasks. By providing precise measurements of comparison capabilities, \mmscore{} enables a more principled approach to developing the next generation of multimodal systems.

\bibliographystyle{plainnat}
\bibliography{neurips_2025}

\begin{thebibliography}{54}
\providecommand{\natexlab}[1]{#1}
\providecommand{\url}[1]{\texttt{#1}}
\expandafter\ifx\csname urlstyle\endcsname\relax
  \providecommand{\doi}[1]{doi: #1}\else
  \providecommand{\doi}{doi: \begingroup \urlstyle{rm}\Url}\fi

\bibitem[Abdin et~al.(2024)Abdin, Jacobs, Awan, Aneja, Awadallah, Awadalla, Bach, Bahree, Bakhtiari, Behl, et~al.]{abdin2024phi}
Marah Abdin, Sam~Ade Jacobs, Ammar~Ahmad Awan, Jyoti Aneja, Ahmed Awadallah, Hany Awadalla, Nguyen Bach, Amit Bahree, Arash Bakhtiari, Harkirat Behl, et~al.
\newblock Phi-3 technical report: A highly capable language model locally on your phone.
\newblock \emph{arXiv preprint arXiv:2404.14219}, 2024.

\bibitem[Achiam et~al.(2023)Achiam, Adler, Agarwal, Ahmad, Akkaya, Aleman, Almeida, Altenschmidt, Altman, Anadkat, et~al.]{achiam2023gpt}
Josh Achiam, Steven Adler, Sandhini Agarwal, Lama Ahmad, Ilge Akkaya, Florencia~Leoni Aleman, Diogo Almeida, Janko Altenschmidt, Sam Altman, Shyamal Anadkat, et~al.
\newblock Gpt-4 technical report.
\newblock \emph{arXiv preprint arXiv:2303.08774}, 2023.

\bibitem[Agrawal et~al.(2024)Agrawal, Antoniak, Hanna, Bout, Chaplot, Chudnovsky, Costa, De~Monicault, Garg, Gervet, et~al.]{agrawal2024pixtral}
Pravesh Agrawal, Szymon Antoniak, Emma~Bou Hanna, Baptiste Bout, Devendra Chaplot, Jessica Chudnovsky, Diogo Costa, Baudouin De~Monicault, Saurabh Garg, Theophile Gervet, et~al.
\newblock Pixtral 12b.
\newblock \emph{arXiv preprint arXiv:2410.07073}, 2024.

\bibitem[Alhamoud et~al.(2025)Alhamoud, Alshammari, Tian, Li, Torr, Kim, and Ghassemi]{alhamoud2025vision}
Kumail Alhamoud, Shaden Alshammari, Yonglong Tian, Guohao Li, Philip Torr, Yoon Kim, and Marzyeh Ghassemi.
\newblock Vision-language models do not understand negation.
\newblock \emph{arXiv preprint arXiv:2501.09425}, 2025.

\bibitem[Awal et~al.(2024)Awal, Ahmadi, Zhang, and Agrawal]{awal2024vismin}
Rabiul Awal, Saba Ahmadi, Le~Zhang, and Aishwarya Agrawal.
\newblock Vismin: Visual minimal-change understanding.
\newblock \emph{arXiv preprint arXiv:2407.16772}, 2024.

\bibitem[Berglund et~al.(2023)Berglund, Tong, Kaufmann, Balesni, Stickland, Korbak, and Evans]{berglund2023reversal}
Lukas Berglund, Meg Tong, Max Kaufmann, Mikita Balesni, Asa~Cooper Stickland, Tomasz Korbak, and Owain Evans.
\newblock The reversal curse: Llms trained on" a is b" fail to learn" b is a".
\newblock \emph{arXiv preprint arXiv:2309.12288}, 2023.

\bibitem[Chen et~al.(2024{\natexlab{a}})Chen, Chen, Zhang, Liu, Wang, Zhou, Zhang, Zhou, Wan, and Sun]{chen2024mllm}
Dongping Chen, Ruoxi Chen, Shilin Zhang, Yinuo Liu, Yaochen Wang, Huichi Zhou, Qihui Zhang, Pan Zhou, Yao Wan, and Lichao Sun.
\newblock Mllm-as-a-judge: Assessing multimodal llm-as-a-judge with vision-language benchmark.
\newblock \emph{arXiv preprint arXiv:2402.04788}, 2024{\natexlab{a}}.

\bibitem[Chen et~al.(2024{\natexlab{b}})Chen, Li, Dong, Zhang, Zang, Chen, Duan, Wang, Qiao, Lin, et~al.]{chen2024we}
Lin Chen, Jinsong Li, Xiaoyi Dong, Pan Zhang, Yuhang Zang, Zehui Chen, Haodong Duan, Jiaqi Wang, Yu~Qiao, Dahua Lin, et~al.
\newblock Are we on the right way for evaluating large vision-language models?
\newblock \emph{arXiv preprint arXiv:2403.20330}, 2024{\natexlab{b}}.

\bibitem[Chen et~al.(2024{\natexlab{c}})Chen, Wang, Cao, Liu, Gao, Cui, Zhu, Ye, Tian, Liu, et~al.]{chen2024expanding}
Zhe Chen, Weiyun Wang, Yue Cao, Yangzhou Liu, Zhangwei Gao, Erfei Cui, Jinguo Zhu, Shenglong Ye, Hao Tian, Zhaoyang Liu, et~al.
\newblock Expanding performance boundaries of open-source multimodal models with model, data, and test-time scaling.
\newblock \emph{arXiv preprint arXiv:2412.05271}, 2024{\natexlab{c}}.

\bibitem[Chiang and Lee(2023)]{chiang2023can}
Cheng-Han Chiang and Hung-yi Lee.
\newblock Can large language models be an alternative to human evaluations?
\newblock \emph{arXiv preprint arXiv:2305.01937}, 2023.

\bibitem[Dao et~al.(2022)Dao, Fu, Ermon, Rudra, and R{\'e}]{dao2022flashattention}
Tri Dao, Daniel~Y. Fu, Stefano Ermon, Atri Rudra, and Christopher R{\'e}.
\newblock Flash{A}ttention: Fast and memory-efficient exact attention with {IO}-awareness.
\newblock In \emph{Advances in Neural Information Processing Systems (NeurIPS)}, 2022.

\bibitem[Deitke et~al.(2024)Deitke, Clark, Lee, Tripathi, Yang, Park, Salehi, Muennighoff, Lo, Soldaini, et~al.]{deitke2024molmo}
Matt Deitke, Christopher Clark, Sangho Lee, Rohun Tripathi, Yue Yang, Jae~Sung Park, Mohammadreza Salehi, Niklas Muennighoff, Kyle Lo, Luca Soldaini, et~al.
\newblock Molmo and pixmo: Open weights and open data for state-of-the-art multimodal models.
\newblock \emph{arXiv preprint arXiv:2409.17146}, 2024.

\bibitem[Deng et~al.(2009)Deng, Dong, Socher, Li, Li, and Fei-Fei]{deng2009imagenet}
Jia Deng, Wei Dong, Richard Socher, Li-Jia Li, Kai Li, and Li~Fei-Fei.
\newblock Imagenet: A large-scale hierarchical image database.
\newblock In \emph{2009 IEEE conference on computer vision and pattern recognition}, pages 248--255. Ieee, 2009.

\bibitem[Duan et~al.(2024)Duan, Yang, Qiao, Fang, Chen, Liu, Dong, Zang, Zhang, Wang, et~al.]{duan2024vlmevalkit}
Haodong Duan, Junming Yang, Yuxuan Qiao, Xinyu Fang, Lin Chen, Yuan Liu, Xiaoyi Dong, Yuhang Zang, Pan Zhang, Jiaqi Wang, et~al.
\newblock Vlmevalkit: An open-source toolkit for evaluating large multi-modality models.
\newblock In \emph{Proceedings of the 32nd ACM International Conference on Multimedia}, pages 11198--11201, 2024.

\bibitem[Fang et~al.(2024)Fang, Wang, Gatmiry, Fang, and Wang]{fang2024rethinking}
Lizhe Fang, Yifei Wang, Khashayar Gatmiry, Lei Fang, and Yisen Wang.
\newblock Rethinking invariance in in-context learning.
\newblock In \emph{ICML 2024 Workshop on Theoretical Foundations of Foundation Models}, 2024.

\bibitem[Feng et~al.(2024)Feng, Ding, Ma, Wang, Zhang, and Chen]{feng2024sample}
Kehua Feng, Keyan Ding, Kede Ma, Zhihua Wang, Qiang Zhang, and Huajun Chen.
\newblock Sample-efficient human evaluation of large language models via maximum discrepancy competition.
\newblock \emph{arXiv preprint arXiv:2404.08008}, 2024.

\bibitem[Fu et~al.(2023)Fu, Ng, Jiang, and Liu]{fu2023gptscore}
Jinlan Fu, See-Kiong Ng, Zhengbao Jiang, and Pengfei Liu.
\newblock Gptscore: Evaluate as you desire.
\newblock \emph{arXiv preprint arXiv:2302.04166}, 2023.

\bibitem[Grattafiori et~al.(2024)Grattafiori, Dubey, Jauhri, Pandey, Kadian, Al-Dahle, Letman, Mathur, Schelten, Vaughan, Yang, Fan, Goyal, Hartshorn, Yang, Mitra, Sravankumar, Korenev, Hinsvark, Rao, Zhang, Rodriguez, Gregerson, Spataru, Roziere, Biron, Tang, Chern, Caucheteux, Nayak, Bi, Marra, McConnell, Keller, Touret, Wu, Wong, Ferrer, Nikolaidis, Allonsius, Song, Pintz, Livshits, Wyatt, Esiobu, Choudhary, Mahajan, Garcia-Olano, Perino, Hupkes, Lakomkin, AlBadawy, Lobanova, Dinan, Smith, Radenovic, Guzmán, Zhang, Synnaeve, Lee, Anderson, Thattai, Nail, Mialon, Pang, Cucurell, Nguyen, Korevaar, Xu, Touvron, Zarov, Ibarra, Kloumann, Misra, Evtimov, Zhang, Copet, Lee, Geffert, Vranes, Park, Mahadeokar, Shah, van~der Linde, Billock, Hong, Lee, Fu, Chi, Huang, Liu, Wang, Yu, Bitton, Spisak, Park, Rocca, Johnstun, Saxe, Jia, Alwala, Prasad, Upasani, Plawiak, Li, Heafield, Stone, El-Arini, Iyer, Malik, Chiu, Bhalla, Lakhotia, Rantala-Yeary, van~der Maaten, Chen, Tan, Jenkins, Martin, Madaan, Malo, Blecher,
  Landzaat, de~Oliveira, Muzzi, Pasupuleti, Singh, Paluri, Kardas, Tsimpoukelli, Oldham, Rita, Pavlova, Kambadur, Lewis, Si, Singh, Hassan, Goyal, Torabi, Bashlykov, Bogoychev, Chatterji, Zhang, Duchenne, Çelebi, Alrassy, Zhang, Li, Vasic, Weng, Bhargava, Dubal, Krishnan, Koura, Xu, He, Dong, Srinivasan, Ganapathy, Calderer, Cabral, Stojnic, Raileanu, Maheswari, Girdhar, Patel, Sauvestre, Polidoro, Sumbaly, Taylor, Silva, Hou, Wang, Hosseini, Chennabasappa, Singh, Bell, Kim, Edunov, Nie, Narang, Raparthy, Shen, Wan, Bhosale, Zhang, Vandenhende, Batra, Whitman, Sootla, Collot, Gururangan, Borodinsky, Herman, Fowler, Sheasha, Georgiou, Scialom, Speckbacher, Mihaylov, Xiao, Karn, Goswami, Gupta, Ramanathan, Kerkez, Gonguet, Do, Vogeti, Albiero, Petrovic, Chu, Xiong, Fu, Meers, Martinet, Wang, Wang, Tan, Xia, Xie, Jia, Wang, Goldschlag, Gaur, Babaei, Wen, Song, Zhang, Li, Mao, Coudert, Yan, Chen, Papakipos, Singh, Srivastava, Jain, Kelsey, Shajnfeld, Gangidi, Victoria, Goldstand, Menon, Sharma, Boesenberg,
  Baevski, Feinstein, Kallet, Sangani, Teo, Yunus, Lupu, Alvarado, Caples, Gu, Ho, Poulton, Ryan, Ramchandani, Dong, Franco, Goyal, Saraf, Chowdhury, Gabriel, Bharambe, Eisenman, Yazdan, James, Maurer, Leonhardi, Huang, Loyd, Paola, Paranjape, Liu, Wu, Ni, Hancock, Wasti, Spence, Stojkovic, Gamido, Montalvo, Parker, Burton, Mejia, Liu, Wang, Kim, Zhou, Hu, Chu, Cai, Tindal, Feichtenhofer, Gao, Civin, Beaty, Kreymer, Li, Adkins, Xu, Testuggine, David, Parikh, Liskovich, Foss, Wang, Le, Holland, Dowling, Jamil, Montgomery, Presani, Hahn, Wood, Le, Brinkman, Arcaute, Dunbar, Smothers, Sun, Kreuk, Tian, Kokkinos, Ozgenel, Caggioni, Kanayet, Seide, Florez, Schwarz, Badeer, Swee, Halpern, Herman, Sizov, Guangyi, Zhang, Lakshminarayanan, Inan, Shojanazeri, Zou, Wang, Zha, Habeeb, Rudolph, Suk, Aspegren, Goldman, Zhan, Damlaj, Molybog, Tufanov, Leontiadis, Veliche, Gat, Weissman, Geboski, Kohli, Lam, Asher, Gaya, Marcus, Tang, Chan, Zhen, Reizenstein, Teboul, Zhong, Jin, Yang, Cummings, Carvill, Shepard, McPhie,
  Torres, Ginsburg, Wang, Wu, U, Saxena, Khandelwal, Zand, Matosich, Veeraraghavan, Michelena, Li, Jagadeesh, Huang, Chawla, Huang, Chen, Garg, A, Silva, Bell, Zhang, Guo, Yu, Moshkovich, Wehrstedt, Khabsa, Avalani, Bhatt, Mankus, Hasson, Lennie, Reso, Groshev, Naumov, Lathi, Keneally, Liu, Seltzer, Valko, Restrepo, Patel, Vyatskov, Samvelyan, Clark, Macey, Wang, Hermoso, Metanat, Rastegari, Bansal, Santhanam, Parks, White, Bawa, Singhal, Egebo, Usunier, Mehta, Laptev, Dong, Cheng, Chernoguz, Hart, Salpekar, Kalinli, Kent, Parekh, Saab, Balaji, Rittner, Bontrager, Roux, Dollar, Zvyagina, Ratanchandani, Yuvraj, Liang, Alao, Rodriguez, Ayub, Murthy, Nayani, Mitra, Parthasarathy, Li, Hogan, Battey, Wang, Howes, Rinott, Mehta, Siby, Bondu, Datta, Chugh, Hunt, Dhillon, Sidorov, Pan, Mahajan, Verma, Yamamoto, Ramaswamy, Lindsay, Lindsay, Feng, Lin, Zha, Patil, Shankar, Zhang, Zhang, Wang, Agarwal, Sajuyigbe, Chintala, Max, Chen, Kehoe, Satterfield, Govindaprasad, Gupta, Deng, Cho, Virk, Subramanian, Choudhury,
  Goldman, Remez, Glaser, Best, Koehler, Robinson, Li, Zhang, Matthews, Chou, Shaked, Vontimitta, Ajayi, Montanez, Mohan, Kumar, Mangla, Ionescu, Poenaru, Mihailescu, Ivanov, Li, Wang, Jiang, Bouaziz, Constable, Tang, Wu, Wang, Wu, Gao, Kleinman, Chen, Hu, Jia, Qi, Li, Zhang, Zhang, Adi, Nam, Yu, Wang, Zhao, Hao, Qian, Li, He, Rait, DeVito, Rosnbrick, Wen, Yang, Zhao, and Ma]{llama3}
Aaron Grattafiori, Abhimanyu Dubey, Abhinav Jauhri, Abhinav Pandey, Abhishek Kadian, Ahmad Al-Dahle, Aiesha Letman, Akhil Mathur, Alan Schelten, Alex Vaughan, Amy Yang, Angela Fan, Anirudh Goyal, Anthony Hartshorn, Aobo Yang, Archi Mitra, Archie Sravankumar, Artem Korenev, Arthur Hinsvark, Arun Rao, Aston Zhang, Aurelien Rodriguez, Austen Gregerson, Ava Spataru, Baptiste Roziere, Bethany Biron, Binh Tang, Bobbie Chern, Charlotte Caucheteux, Chaya Nayak, Chloe Bi, Chris Marra, Chris McConnell, Christian Keller, Christophe Touret, Chunyang Wu, Corinne Wong, Cristian~Canton Ferrer, Cyrus Nikolaidis, Damien Allonsius, Daniel Song, Danielle Pintz, Danny Livshits, Danny Wyatt, David Esiobu, Dhruv Choudhary, Dhruv Mahajan, Diego Garcia-Olano, Diego Perino, Dieuwke Hupkes, Egor Lakomkin, Ehab AlBadawy, Elina Lobanova, Emily Dinan, Eric~Michael Smith, Filip Radenovic, Francisco Guzmán, Frank Zhang, Gabriel Synnaeve, Gabrielle Lee, Georgia~Lewis Anderson, Govind Thattai, Graeme Nail, Gregoire Mialon, Guan Pang,
  Guillem Cucurell, Hailey Nguyen, Hannah Korevaar, Hu~Xu, Hugo Touvron, Iliyan Zarov, Imanol~Arrieta Ibarra, Isabel Kloumann, Ishan Misra, Ivan Evtimov, Jack Zhang, Jade Copet, Jaewon Lee, Jan Geffert, Jana Vranes, Jason Park, Jay Mahadeokar, Jeet Shah, Jelmer van~der Linde, Jennifer Billock, Jenny Hong, Jenya Lee, Jeremy Fu, Jianfeng Chi, Jianyu Huang, Jiawen Liu, Jie Wang, Jiecao Yu, Joanna Bitton, Joe Spisak, Jongsoo Park, Joseph Rocca, Joshua Johnstun, Joshua Saxe, Junteng Jia, Kalyan~Vasuden Alwala, Karthik Prasad, Kartikeya Upasani, Kate Plawiak, Ke~Li, Kenneth Heafield, Kevin Stone, Khalid El-Arini, Krithika Iyer, Kshitiz Malik, Kuenley Chiu, Kunal Bhalla, Kushal Lakhotia, Lauren Rantala-Yeary, Laurens van~der Maaten, Lawrence Chen, Liang Tan, Liz Jenkins, Louis Martin, Lovish Madaan, Lubo Malo, Lukas Blecher, Lukas Landzaat, Luke de~Oliveira, Madeline Muzzi, Mahesh Pasupuleti, Mannat Singh, Manohar Paluri, Marcin Kardas, Maria Tsimpoukelli, Mathew Oldham, Mathieu Rita, Maya Pavlova, Melanie Kambadur,
  Mike Lewis, Min Si, Mitesh~Kumar Singh, Mona Hassan, Naman Goyal, Narjes Torabi, Nikolay Bashlykov, Nikolay Bogoychev, Niladri Chatterji, Ning Zhang, Olivier Duchenne, Onur Çelebi, Patrick Alrassy, Pengchuan Zhang, Pengwei Li, Petar Vasic, Peter Weng, Prajjwal Bhargava, Pratik Dubal, Praveen Krishnan, Punit~Singh Koura, Puxin Xu, Qing He, Qingxiao Dong, Ragavan Srinivasan, Raj Ganapathy, Ramon Calderer, Ricardo~Silveira Cabral, Robert Stojnic, Roberta Raileanu, Rohan Maheswari, Rohit Girdhar, Rohit Patel, Romain Sauvestre, Ronnie Polidoro, Roshan Sumbaly, Ross Taylor, Ruan Silva, Rui Hou, Rui Wang, Saghar Hosseini, Sahana Chennabasappa, Sanjay Singh, Sean Bell, Seohyun~Sonia Kim, Sergey Edunov, Shaoliang Nie, Sharan Narang, Sharath Raparthy, Sheng Shen, Shengye Wan, Shruti Bhosale, Shun Zhang, Simon Vandenhende, Soumya Batra, Spencer Whitman, Sten Sootla, Stephane Collot, Suchin Gururangan, Sydney Borodinsky, Tamar Herman, Tara Fowler, Tarek Sheasha, Thomas Georgiou, Thomas Scialom, Tobias Speckbacher,
  Todor Mihaylov, Tong Xiao, Ujjwal Karn, Vedanuj Goswami, Vibhor Gupta, Vignesh Ramanathan, Viktor Kerkez, Vincent Gonguet, Virginie Do, Vish Vogeti, Vítor Albiero, Vladan Petrovic, Weiwei Chu, Wenhan Xiong, Wenyin Fu, Whitney Meers, Xavier Martinet, Xiaodong Wang, Xiaofang Wang, Xiaoqing~Ellen Tan, Xide Xia, Xinfeng Xie, Xuchao Jia, Xuewei Wang, Yaelle Goldschlag, Yashesh Gaur, Yasmine Babaei, Yi~Wen, Yiwen Song, Yuchen Zhang, Yue Li, Yuning Mao, Zacharie~Delpierre Coudert, Zheng Yan, Zhengxing Chen, Zoe Papakipos, Aaditya Singh, Aayushi Srivastava, Abha Jain, Adam Kelsey, Adam Shajnfeld, Adithya Gangidi, Adolfo Victoria, Ahuva Goldstand, Ajay Menon, Ajay Sharma, Alex Boesenberg, Alexei Baevski, Allie Feinstein, Amanda Kallet, Amit Sangani, Amos Teo, Anam Yunus, Andrei Lupu, Andres Alvarado, Andrew Caples, Andrew Gu, Andrew Ho, Andrew Poulton, Andrew Ryan, Ankit Ramchandani, Annie Dong, Annie Franco, Anuj Goyal, Aparajita Saraf, Arkabandhu Chowdhury, Ashley Gabriel, Ashwin Bharambe, Assaf Eisenman, Azadeh
  Yazdan, Beau James, Ben Maurer, Benjamin Leonhardi, Bernie Huang, Beth Loyd, Beto~De Paola, Bhargavi Paranjape, Bing Liu, Bo~Wu, Boyu Ni, Braden Hancock, Bram Wasti, Brandon Spence, Brani Stojkovic, Brian Gamido, Britt Montalvo, Carl Parker, Carly Burton, Catalina Mejia, Ce~Liu, Changhan Wang, Changkyu Kim, Chao Zhou, Chester Hu, Ching-Hsiang Chu, Chris Cai, Chris Tindal, Christoph Feichtenhofer, Cynthia Gao, Damon Civin, Dana Beaty, Daniel Kreymer, Daniel Li, David Adkins, David Xu, Davide Testuggine, Delia David, Devi Parikh, Diana Liskovich, Didem Foss, Dingkang Wang, Duc Le, Dustin Holland, Edward Dowling, Eissa Jamil, Elaine Montgomery, Eleonora Presani, Emily Hahn, Emily Wood, Eric-Tuan Le, Erik Brinkman, Esteban Arcaute, Evan Dunbar, Evan Smothers, Fei Sun, Felix Kreuk, Feng Tian, Filippos Kokkinos, Firat Ozgenel, Francesco Caggioni, Frank Kanayet, Frank Seide, Gabriela~Medina Florez, Gabriella Schwarz, Gada Badeer, Georgia Swee, Gil Halpern, Grant Herman, Grigory Sizov, Guangyi, Zhang, Guna
  Lakshminarayanan, Hakan Inan, Hamid Shojanazeri, Han Zou, Hannah Wang, Hanwen Zha, Haroun Habeeb, Harrison Rudolph, Helen Suk, Henry Aspegren, Hunter Goldman, Hongyuan Zhan, Ibrahim Damlaj, Igor Molybog, Igor Tufanov, Ilias Leontiadis, Irina-Elena Veliche, Itai Gat, Jake Weissman, James Geboski, James Kohli, Janice Lam, Japhet Asher, Jean-Baptiste Gaya, Jeff Marcus, Jeff Tang, Jennifer Chan, Jenny Zhen, Jeremy Reizenstein, Jeremy Teboul, Jessica Zhong, Jian Jin, Jingyi Yang, Joe Cummings, Jon Carvill, Jon Shepard, Jonathan McPhie, Jonathan Torres, Josh Ginsburg, Junjie Wang, Kai Wu, Kam~Hou U, Karan Saxena, Kartikay Khandelwal, Katayoun Zand, Kathy Matosich, Kaushik Veeraraghavan, Kelly Michelena, Keqian Li, Kiran Jagadeesh, Kun Huang, Kunal Chawla, Kyle Huang, Lailin Chen, Lakshya Garg, Lavender A, Leandro Silva, Lee Bell, Lei Zhang, Liangpeng Guo, Licheng Yu, Liron Moshkovich, Luca Wehrstedt, Madian Khabsa, Manav Avalani, Manish Bhatt, Martynas Mankus, Matan Hasson, Matthew Lennie, Matthias Reso, Maxim
  Groshev, Maxim Naumov, Maya Lathi, Meghan Keneally, Miao Liu, Michael~L. Seltzer, Michal Valko, Michelle Restrepo, Mihir Patel, Mik Vyatskov, Mikayel Samvelyan, Mike Clark, Mike Macey, Mike Wang, Miquel~Jubert Hermoso, Mo~Metanat, Mohammad Rastegari, Munish Bansal, Nandhini Santhanam, Natascha Parks, Natasha White, Navyata Bawa, Nayan Singhal, Nick Egebo, Nicolas Usunier, Nikhil Mehta, Nikolay~Pavlovich Laptev, Ning Dong, Norman Cheng, Oleg Chernoguz, Olivia Hart, Omkar Salpekar, Ozlem Kalinli, Parkin Kent, Parth Parekh, Paul Saab, Pavan Balaji, Pedro Rittner, Philip Bontrager, Pierre Roux, Piotr Dollar, Polina Zvyagina, Prashant Ratanchandani, Pritish Yuvraj, Qian Liang, Rachad Alao, Rachel Rodriguez, Rafi Ayub, Raghotham Murthy, Raghu Nayani, Rahul Mitra, Rangaprabhu Parthasarathy, Raymond Li, Rebekkah Hogan, Robin Battey, Rocky Wang, Russ Howes, Ruty Rinott, Sachin Mehta, Sachin Siby, Sai~Jayesh Bondu, Samyak Datta, Sara Chugh, Sara Hunt, Sargun Dhillon, Sasha Sidorov, Satadru Pan, Saurabh Mahajan,
  Saurabh Verma, Seiji Yamamoto, Sharadh Ramaswamy, Shaun Lindsay, Shaun Lindsay, Sheng Feng, Shenghao Lin, Shengxin~Cindy Zha, Shishir Patil, Shiva Shankar, Shuqiang Zhang, Shuqiang Zhang, Sinong Wang, Sneha Agarwal, Soji Sajuyigbe, Soumith Chintala, Stephanie Max, Stephen Chen, Steve Kehoe, Steve Satterfield, Sudarshan Govindaprasad, Sumit Gupta, Summer Deng, Sungmin Cho, Sunny Virk, Suraj Subramanian, Sy~Choudhury, Sydney Goldman, Tal Remez, Tamar Glaser, Tamara Best, Thilo Koehler, Thomas Robinson, Tianhe Li, Tianjun Zhang, Tim Matthews, Timothy Chou, Tzook Shaked, Varun Vontimitta, Victoria Ajayi, Victoria Montanez, Vijai Mohan, Vinay~Satish Kumar, Vishal Mangla, Vlad Ionescu, Vlad Poenaru, Vlad~Tiberiu Mihailescu, Vladimir Ivanov, Wei Li, Wenchen Wang, Wenwen Jiang, Wes Bouaziz, Will Constable, Xiaocheng Tang, Xiaojian Wu, Xiaolan Wang, Xilun Wu, Xinbo Gao, Yaniv Kleinman, Yanjun Chen, Ye~Hu, Ye~Jia, Ye~Qi, Yenda Li, Yilin Zhang, Ying Zhang, Yossi Adi, Youngjin Nam, Yu, Wang, Yu~Zhao, Yuchen Hao, Yundi
  Qian, Yunlu Li, Yuzi He, Zach Rait, Zachary DeVito, Zef Rosnbrick, Zhaoduo Wen, Zhenyu Yang, Zhiwei Zhao, and Zhiyu Ma.
\newblock The llama 3 herd of models, 2024.
\newblock URL \url{https://arxiv.org/abs/2407.21783}.

\bibitem[Guan et~al.(2024)Guan, Liu, Wu, Xian, Li, Liu, Wang, Chen, Huang, Yacoob, et~al.]{guan2024hallusionbench}
Tianrui Guan, Fuxiao Liu, Xiyang Wu, Ruiqi Xian, Zongxia Li, Xiaoyu Liu, Xijun Wang, Lichang Chen, Furong Huang, Yaser Yacoob, et~al.
\newblock Hallusionbench: an advanced diagnostic suite for entangled language hallucination and visual illusion in large vision-language models.
\newblock In \emph{Proceedings of the IEEE/CVF Conference on Computer Vision and Pattern Recognition}, pages 14375--14385, 2024.

\bibitem[Guo et~al.(2025)Guo, Yang, Zhang, Song, Zhang, Xu, Zhu, Ma, Wang, Bi, et~al.]{guo2025deepseek}
Daya Guo, Dejian Yang, Haowei Zhang, Junxiao Song, Ruoyu Zhang, Runxin Xu, Qihao Zhu, Shirong Ma, Peiyi Wang, Xiao Bi, et~al.
\newblock Deepseek-r1: Incentivizing reasoning capability in llms via reinforcement learning.
\newblock \emph{arXiv preprint arXiv:2501.12948}, 2025.

\bibitem[Johnson et~al.(2017)Johnson, Hariharan, Van Der~Maaten, Fei-Fei, Lawrence~Zitnick, and Girshick]{johnson2017clevr}
Justin Johnson, Bharath Hariharan, Laurens Van Der~Maaten, Li~Fei-Fei, C~Lawrence~Zitnick, and Ross Girshick.
\newblock Clevr: A diagnostic dataset for compositional language and elementary visual reasoning.
\newblock In \emph{Proceedings of the IEEE conference on computer vision and pattern recognition}, pages 2901--2910, 2017.

\bibitem[Kamath et~al.(2023)Kamath, Hessel, and Chang]{kamath2023s}
Amita Kamath, Jack Hessel, and Kai-Wei Chang.
\newblock What's" up" with vision-language models? investigating their struggle with spatial reasoning.
\newblock \emph{arXiv preprint arXiv:2310.19785}, 2023.

\bibitem[Kembhavi et~al.(2016)Kembhavi, Salvato, Kolve, Seo, Hajishirzi, and Farhadi]{kembhavi2016diagram}
Aniruddha Kembhavi, Mike Salvato, Eric Kolve, Minjoon Seo, Hannaneh Hajishirzi, and Ali Farhadi.
\newblock A diagram is worth a dozen images.
\newblock In \emph{Computer Vision--ECCV 2016: 14th European Conference, Amsterdam, The Netherlands, October 11--14, 2016, Proceedings, Part IV 14}, pages 235--251. Springer, 2016.

\bibitem[Knox et~al.(2024)Knox, Hatgis-Kessell, Booth, Niekum, Stone, and Allievi]{knox2024models}
W.~Bradley Knox, Stephane Hatgis-Kessell, Serena Booth, Scott Niekum, Peter Stone, and Alessandro~G Allievi.
\newblock Models of human preference for learning reward functions.
\newblock \emph{Transactions on Machine Learning Research}, 2024.
\newblock ISSN 2835-8856.
\newblock URL \url{https://openreview.net/forum?id=hpKJkVoThY}.

\bibitem[Lauren{\c{c}}on et~al.(2024)Lauren{\c{c}}on, Tronchon, Cord, and Sanh]{laurenccon2024matters}
Hugo Lauren{\c{c}}on, L{\'e}o Tronchon, Matthieu Cord, and Victor Sanh.
\newblock What matters when building vision-language models?
\newblock \emph{arXiv preprint arXiv:2405.02246}, 2024.

\bibitem[Li et~al.(2024)Li, Zhang, Guo, Zhang, Li, Zhang, Zhang, Zhang, Li, Liu, et~al.]{li2024llava}
Bo~Li, Yuanhan Zhang, Dong Guo, Renrui Zhang, Feng Li, Hao Zhang, Kaichen Zhang, Peiyuan Zhang, Yanwei Li, Ziwei Liu, et~al.
\newblock Llava-onevision: Easy visual task transfer.
\newblock \emph{arXiv preprint arXiv:2408.03326}, 2024.

\bibitem[Lin et~al.(2014)Lin, Maire, Belongie, Hays, Perona, Ramanan, Doll{\'a}r, and Zitnick]{lin2014microsoft}
Tsung-Yi Lin, Michael Maire, Serge Belongie, James Hays, Pietro Perona, Deva Ramanan, Piotr Doll{\'a}r, and C~Lawrence Zitnick.
\newblock Microsoft coco: Common objects in context.
\newblock In \emph{Computer Vision--ECCV 2014: 13th European Conference, Zurich, Switzerland, September 6-12, 2014, Proceedings, Part V 13}, pages 740--755. Springer, 2014.

\bibitem[Liu et~al.(2019)Liu, Cheng, Sun, Wang, Nie, and Kankanhalli]{liu2019user}
Fan Liu, Zhiyong Cheng, Changchang Sun, Yinglong Wang, Liqiang Nie, and Mohan Kankanhalli.
\newblock User diverse preference modeling by multimodal attentive metric learning.
\newblock In \emph{Proceedings of the 27th ACM international conference on multimedia}, pages 1526--1534, 2019.

\bibitem[Liu et~al.(2024{\natexlab{a}})Liu, Li, Wu, and Lee]{liu2024visual}
Haotian Liu, Chunyuan Li, Qingyang Wu, and Yong~Jae Lee.
\newblock Visual instruction tuning.
\newblock \emph{Advances in neural information processing systems}, 36, 2024{\natexlab{a}}.

\bibitem[Liu et~al.(2023)Liu, Iter, Xu, Wang, Xu, and Zhu]{liu2023geval}
Yang Liu, Dan Iter, Yichong Xu, Shuohang Wang, Ruochen Xu, and Chenguang Zhu.
\newblock G-eval: Nlg evaluation using gpt-4 with better human alignment.
\newblock In \emph{Proceedings of the 2023 Conference on Empirical Methods in Natural Language Processing}, pages 2511--2522, 2023.

\bibitem[Liu et~al.(2025)Liu, Duan, Zhang, Li, Zhang, Zhao, Yuan, Wang, He, Liu, et~al.]{liu2025mmbench}
Yuan Liu, Haodong Duan, Yuanhan Zhang, Bo~Li, Songyang Zhang, Wangbo Zhao, Yike Yuan, Jiaqi Wang, Conghui He, Ziwei Liu, et~al.
\newblock Mmbench: Is your multi-modal model an all-around player?
\newblock In \emph{European conference on computer vision}, pages 216--233. Springer, 2025.

\bibitem[Liu et~al.(2024{\natexlab{b}})Liu, Li, Huang, Yang, Yu, Li, Yin, Liu, Jin, and Bai]{liu2024ocrbench}
Yuliang Liu, Zhang Li, Mingxin Huang, Biao Yang, Wenwen Yu, Chunyuan Li, Xu-Cheng Yin, Cheng-Lin Liu, Lianwen Jin, and Xiang Bai.
\newblock Ocrbench: on the hidden mystery of ocr in large multimodal models.
\newblock \emph{Science China Information Sciences}, 67\penalty0 (12):\penalty0 220102, 2024{\natexlab{b}}.

\bibitem[Lu et~al.(2023)Lu, Bansal, Xia, Liu, Li, Hajishirzi, Cheng, Chang, Galley, and Gao]{lu2023mathvista}
Pan Lu, Hritik Bansal, Tony Xia, Jiacheng Liu, Chunyuan Li, Hannaneh Hajishirzi, Hao Cheng, Kai-Wei Chang, Michel Galley, and Jianfeng Gao.
\newblock Mathvista: Evaluating mathematical reasoning of foundation models in visual contexts.
\newblock \emph{arXiv preprint arXiv:2310.02255}, 2023.

\bibitem[Lu et~al.(2024)Lu, Peng, Cheng, Galley, Chang, Wu, Zhu, and Gao]{lu2024chameleon}
Pan Lu, Baolin Peng, Hao Cheng, Michel Galley, Kai-Wei Chang, Ying~Nian Wu, Song-Chun Zhu, and Jianfeng Gao.
\newblock Chameleon: Plug-and-play compositional reasoning with large language models.
\newblock \emph{Advances in Neural Information Processing Systems}, 36, 2024.

\bibitem[Ma{\~n}as et~al.(2024)Ma{\~n}as, Krojer, and Agrawal]{manas2024improving}
Oscar Ma{\~n}as, Benno Krojer, and Aishwarya Agrawal.
\newblock Improving automatic vqa evaluation using large language models.
\newblock In \emph{Proceedings of the AAAI Conference on Artificial Intelligence}, volume~38, pages 4171--4179, 2024.

\bibitem[Murugadoss et~al.(2024)Murugadoss, Poelitz, Drosos, Le, McKenna, Negreanu, Parnin, and Sarkar]{murugadoss2024evaluating}
Bhuvanashree Murugadoss, Christian Poelitz, Ian Drosos, Vu~Le, Nick McKenna, Carina~Suzana Negreanu, Chris Parnin, and Advait Sarkar.
\newblock Evaluating the evaluator: Measuring llms' adherence to task evaluation instructions.
\newblock \emph{arXiv preprint arXiv:2408.08781}, 2024.

\bibitem[Ouali et~al.(2024)Ouali, Bulat, Xenos, Zaganidis, Metaxas, Tzimiropoulos, and Martinez]{ouali2024discriminative}
Yassine Ouali, Adrian Bulat, Alexandros Xenos, Anestis Zaganidis, Ioannis~Maniadis Metaxas, Georgios Tzimiropoulos, and Brais Martinez.
\newblock Discriminative fine-tuning of lvlms.
\newblock \emph{arXiv preprint arXiv:2412.04378}, 2024.

\bibitem[Pacchiardi et~al.(2024)Pacchiardi, Cheke, and Hern{\'a}ndez-Orallo]{pacchiardi2024100}
Lorenzo Pacchiardi, Lucy~G Cheke, and Jos{\'e} Hern{\'a}ndez-Orallo.
\newblock 100 instances is all you need: predicting the success of a new llm on unseen data by testing on a few instances.
\newblock \emph{arXiv preprint arXiv:2409.03563}, 2024.

\bibitem[Polo et~al.(2024{\natexlab{a}})Polo, Weber, Choshen, Sun, Xu, and Yurochkin]{polo2024tinybenchmarks}
Felipe~Maia Polo, Lucas Weber, Leshem Choshen, Yuekai Sun, Gongjun Xu, and Mikhail Yurochkin.
\newblock tinybenchmarks: evaluating {LLM}s with fewer examples.
\newblock In \emph{Forty-first International Conference on Machine Learning}, 2024{\natexlab{a}}.
\newblock URL \url{https://openreview.net/forum?id=qAml3FpfhG}.

\bibitem[Polo et~al.(2024{\natexlab{b}})Polo, Xu, Weber, Silva, Bhardwaj, Choshen, de~Oliveira, Sun, and Yurochkin]{NEURIPS2024_28236482}
Felipe~Maia Polo, Ronald Xu, Lucas Weber, M\'{\i}rian Silva, Onkar Bhardwaj, Leshem Choshen, Allysson Flavio~Melo de~Oliveira, Yuekai Sun, and Mikhail Yurochkin.
\newblock Efficient multi-prompt evaluation of llms.
\newblock In A.~Globerson, L.~Mackey, D.~Belgrave, A.~Fan, U.~Paquet, J.~Tomczak, and C.~Zhang, editors, \emph{Advances in Neural Information Processing Systems}, volume~37, pages 22483--22512. Curran Associates, Inc., 2024{\natexlab{b}}.
\newblock URL \url{https://proceedings.neurips.cc/paper_files/paper/2024/file/28236482f64a72eec43706b6f3a6c511-Paper-Conference.pdf}.

\bibitem[Radford et~al.(2021)Radford, Kim, Hallacy, Ramesh, Goh, Agarwal, Sastry, Askell, Mishkin, Clark, et~al.]{radford2021learning}
Alec Radford, Jong~Wook Kim, Chris Hallacy, Aditya Ramesh, Gabriel Goh, Sandhini Agarwal, Girish Sastry, Amanda Askell, Pamela Mishkin, Jack Clark, et~al.
\newblock Learning transferable visual models from natural language supervision.
\newblock In \emph{International conference on machine learning}, pages 8748--8763. PMLR, 2021.

\bibitem[Reid et~al.(2024)Reid, Savinov, Teplyashin, Lepikhin, Lillicrap, Alayrac, Soricut, Lazaridou, Firat, Schrittwieser, et~al.]{reid2024gemini}
Machel Reid, Nikolay Savinov, Denis Teplyashin, Dmitry Lepikhin, Timothy Lillicrap, Jean-baptiste Alayrac, Radu Soricut, Angeliki Lazaridou, Orhan Firat, Julian Schrittwieser, et~al.
\newblock Gemini 1.5: Unlocking multimodal understanding across millions of tokens of context.
\newblock \emph{arXiv preprint arXiv:2403.05530}, 2024.

\bibitem[Sam et~al.(2024)Sam, Willmott, Semedo, and Kolter]{sam2024finetuning}
Dylan Sam, Devin Willmott, Joao~D Semedo, and J~Zico Kolter.
\newblock Finetuning clip to reason about pairwise differences.
\newblock \emph{arXiv preprint arXiv:2409.09721}, 2024.

\bibitem[Shankar et~al.(2024)Shankar, Zamfirescu-Pereira, Hartmann, Parameswaran, and Arawjo]{shankar2024validates}
Shreya Shankar, JD~Zamfirescu-Pereira, Bj{\"o}rn Hartmann, Aditya~G Parameswaran, and Ian Arawjo.
\newblock Who validates the validators? aligning llm-assisted evaluation of llm outputs with human preferences.
\newblock \emph{arXiv preprint arXiv:2404.12272}, 2024.

\bibitem[Thakur et~al.(2024)Thakur, Choudhary, Ramayapally, Vaidyanathan, and Hupkes]{thakur2024judging}
Aman~Singh Thakur, Kartik Choudhary, Venkat~Srinik Ramayapally, Sankaran Vaidyanathan, and Dieuwke Hupkes.
\newblock Judging the judges: Evaluating alignment and vulnerabilities in llms-as-judges.
\newblock \emph{arXiv preprint arXiv:2406.12624}, 2024.

\bibitem[Wang et~al.(2024{\natexlab{a}})Wang, Ming, Shi, Vineet, Wang, Li, and Joshi]{wang2024picture}
Jiayu Wang, Yifei Ming, Zhenmei Shi, Vibhav Vineet, Xin Wang, Yixuan Li, and Neel Joshi.
\newblock Is a picture worth a thousand words? delving into spatial reasoning for vision language models.
\newblock \emph{arXiv preprint arXiv:2406.14852}, 2024{\natexlab{a}}.

\bibitem[Wang et~al.(2024{\natexlab{b}})Wang, Bai, Tan, Wang, Fan, Bai, Chen, Liu, Wang, Ge, et~al.]{wang2024qwen2}
Peng Wang, Shuai Bai, Sinan Tan, Shijie Wang, Zhihao Fan, Jinze Bai, Keqin Chen, Xuejing Liu, Jialin Wang, Wenbin Ge, et~al.
\newblock Qwen2-vl: Enhancing vision-language model's perception of the world at any resolution.
\newblock \emph{arXiv preprint arXiv:2409.12191}, 2024{\natexlab{b}}.

\bibitem[Wang et~al.(2024{\natexlab{c}})Wang, Chen, Wang, Cao, Liu, Gao, Zhu, Zhu, Lu, Qiao, et~al.]{wang2024enhancing}
Weiyun Wang, Zhe Chen, Wenhai Wang, Yue Cao, Yangzhou Liu, Zhangwei Gao, Jinguo Zhu, Xizhou Zhu, Lewei Lu, Yu~Qiao, et~al.
\newblock Enhancing the reasoning ability of multimodal large language models via mixed preference optimization.
\newblock \emph{arXiv preprint arXiv:2411.10442}, 2024{\natexlab{c}}.

\bibitem[Yu et~al.(2023)Yu, Yang, Li, Wang, Lin, Liu, Wang, and Wang]{yu2023mm}
Weihao Yu, Zhengyuan Yang, Linjie Li, Jianfeng Wang, Kevin Lin, Zicheng Liu, Xinchao Wang, and Lijuan Wang.
\newblock Mm-vet: Evaluating large multimodal models for integrated capabilities.
\newblock \emph{arXiv preprint arXiv:2308.02490}, 2023.

\bibitem[Yuan et~al.(2025)Yuan, Zhang, Feng, Li, Wang, Shi, Tan, Pan, Hu, and Li]{yuan2025beyond}
Peiwen Yuan, Yueqi Zhang, Shaoxiong Feng, Yiwei Li, Xinglin Wang, Jiayi Shi, Chuyi Tan, Boyuan Pan, Yao Hu, and Kan Li.
\newblock Beyond one-size-fits-all: Tailored benchmarks for efficient evaluation.
\newblock \emph{arXiv preprint arXiv:2502.13576}, 2025.

\bibitem[Yue et~al.(2024)Yue, Ni, Zhang, Zheng, Liu, Zhang, Stevens, Jiang, Ren, Sun, et~al.]{yue2024mmmu}
Xiang Yue, Yuansheng Ni, Kai Zhang, Tianyu Zheng, Ruoqi Liu, Ge~Zhang, Samuel Stevens, Dongfu Jiang, Weiming Ren, Yuxuan Sun, et~al.
\newblock Mmmu: A massive multi-discipline multimodal understanding and reasoning benchmark for expert agi.
\newblock In \emph{Proceedings of the IEEE/CVF Conference on Computer Vision and Pattern Recognition}, pages 9556--9567, 2024.

\bibitem[Yuksekgonul et~al.(2023)Yuksekgonul, Bianchi, Kalluri, Jurafsky, and Zou]{yuksekgonul2023when}
Mert Yuksekgonul, Federico Bianchi, Pratyusha Kalluri, Dan Jurafsky, and James Zou.
\newblock When and why vision-language models behave like bags-of-words, and what to do about it?
\newblock In \emph{The Eleventh International Conference on Learning Representations}, 2023.
\newblock URL \url{https://openreview.net/forum?id=KRLUvxh8uaX}.

\bibitem[Zellers et~al.(2019)Zellers, Bisk, Farhadi, and Choi]{zellers2019recognition}
Rowan Zellers, Yonatan Bisk, Ali Farhadi, and Yejin Choi.
\newblock From recognition to cognition: Visual commonsense reasoning.
\newblock In \emph{Proceedings of the IEEE/CVF conference on computer vision and pattern recognition}, pages 6720--6731, 2019.

\bibitem[Zheng et~al.(2023)Zheng, Chiang, Sheng, Zhuang, Wu, Zhuang, Lin, Li, Li, Xing, et~al.]{zheng2023judging}
Lianmin Zheng, Wei-Lin Chiang, Ying Sheng, Siyuan Zhuang, Zhanghao Wu, Yonghao Zhuang, Zi~Lin, Zhuohan Li, Dacheng Li, Eric Xing, et~al.
\newblock Judging llm-as-a-judge with mt-bench and chatbot arena.
\newblock \emph{Advances in Neural Information Processing Systems}, 36:\penalty0 46595--46623, 2023.

\end{thebibliography}

\appendix
\clearpage
\section{Error Analysis}
\label{sec:error-analysis}
In this section, we look into the outputs of each model and their errors given different data pairs. More specifically, we look into errors made by \geminiPro{}, \gptFouroEleven{}, \internvlTwoFiveEightB{}, \pixtral{}, \qwenTwoVLSevenB{}, and \phiThreeFive{}.
\subsection{Gemini Model}
Below is an error example of \geminiPro{}{} on a data-pair from \mmscorecoco with color jittering (CJ).

\begin{tcolorbox}[enhanced,attach boxed title to top center={yshift=-3mm,yshifttext=-1mm},
  colback=red!5!white,colframe=red!20!gray,colbacktitle=red!20!gray,
  title=\geminiPro{} Evaluation Instructions,fonttitle=\bfseries,
  boxed title style={size=small,colframe=red!20!gray} ]

\textbf{User prompt}: Evaluate the similarity of the images based on the following conditions:
\emph{- Be invariant to color jittering while evaluating images. Even if one image has been modified with random color changes (e.g., brightness or contrast adjustments), the similarity score should remain high. If the images are different, the similarity score should decrease regardless of the jittering.}

The score should take into account how well the images align with these conditions, as well as their overall resemblance. Even if the images are identical but fail to meet the conditions, they should still receive a higher score than completely different images. Provide a score from 1 to 10, where 1 indicates no similarity and 10 indicates identical images that fully satisfy the conditions. Respond only in this format:

\begin{verbatim}
Score: <1-10>
Reason: <reason for score>
\end{verbatim}
Nothing else should be included. What score would you give?  ``\texttt{data1}'' ``\texttt{data2}'' 
\end{tcolorbox}

\noindent
\begin{minipage}{0.48\textwidth}
\begin{tcolorbox}[enhanced,attach boxed title to top center={yshift=-3mm,yshifttext=-1mm},
  colback=red!5!white,colframe=red!20!gray,colbacktitle=red!20!gray,
  title=\geminiPro{} Response - Original Order,fonttitle=\bfseries,
  boxed title style={size=small,colframe=red!20!gray},
  equal height group=examplegroup]

\begin{center}
    \includegraphics[width=0.9\linewidth]{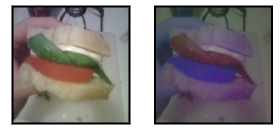}
\end{center}

\textbf{Model response:}
\begin{tcolorbox}[colback=gray!10!white, colframe=gray!50, sharp corners=south, boxrule=0.8pt]
\textit{Score: 10\\Reason: The images are of the same sandwich, one in normal color and the other with a strong purple color filter applied.  
Despite the heavy color distortion, the content is identical, fulfilling the color invariance requirement.}
\end{tcolorbox}

\end{tcolorbox}
\end{minipage}
\hfill
\begin{minipage}{0.48\textwidth}
\begin{tcolorbox}[enhanced,attach boxed title to top center={yshift=-3mm,yshifttext=-1mm},
  colback=red!5!white,colframe=red!20!gray,colbacktitle=red!20!gray,
  title=\geminiPro{} Response - Reversed Order,fonttitle=\bfseries,
  boxed title style={size=small,colframe=red!20!gray},
  equal height group=examplegroup]

\begin{center}
    \includegraphics[width=0.9\linewidth]{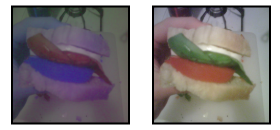}
\end{center}

\textbf{Model response:}
\begin{tcolorbox}[colback=gray!10!white, colframe=gray!50, sharp corners=south, boxrule=0.8pt]
\textit{Score: 1\\Reason: The first image appears to be a mold of some sort, possibly for dental work or other medical purposes.  
The second image is a Caprese sandwich. They are entirely different objects. While the layered structure might be vaguely similar,  
the content, context, and purpose are completely dissimilar.}
\end{tcolorbox}

\end{tcolorbox}
\end{minipage}

\subsection{GPT4o Model}
Below is an error example of \gptFouroEleven{} on a data-pair from \mmscorecoco with color jittering (CJ).
\begin{tcolorbox}[enhanced,attach boxed title to top center={yshift=-3mm,yshifttext=-1mm},
  colback=red!5!white,colframe=red!20!gray,colbacktitle=red!20!gray,
  title=\gptFouroEleven{} Evaluation Instructions,fonttitle=\bfseries,
  boxed title style={size=small,colframe=red!20!gray} ]

\textbf{User prompt}: Evaluate the similarity of the images based on the following conditions:

\emph{- Be invariant to color jittering while evaluating images. Even if one image has been modified with random color changes (e.g., brightness or contrast adjustments), the similarity score should remain high. If the images are different, the similarity score should decrease regardless of the jittering.}

The score should take into account how well the images align with these conditions, as well as their overall resemblance. Even if the images are identical but fail to meet the conditions, they should still receive a higher score than completely different images. 

Provide a score from 1 to 10, where 1 indicates no similarity and 10 indicates identical images that fully satisfy the conditions.

Respond only in this format:
\begin{verbatim}
Score: <1-10>
Reason: <reason for score>
\end{verbatim}

Nothing else should be included. What score would you give? \texttt{"data1"} \texttt{"data2"} 

\end{tcolorbox}

\noindent
\begin{minipage}{0.48\textwidth}
\begin{tcolorbox}[enhanced,attach boxed title to top center={yshift=-3mm,yshifttext=-1mm},
  colback=red!5!white,colframe=red!20!gray,colbacktitle=red!20!gray,
  title=\gptFouroEleven{} Response - Original Order,fonttitle=\bfseries,
  boxed title style={size=small,colframe=red!20!gray},
  equal height group=examplegroup]

\begin{center}
\includegraphics[width=0.9\linewidth]{imgs/pair-burger-color1.png}
\end{center}

\textbf{Model response:}
\begin{tcolorbox}[colback=gray!10!white, colframe=gray!50, sharp corners=south, boxrule=0.8pt]
\textit{Score: 8\\Reason: The images are of the same sandwich, with similar shapes and arrangements of the bread, tomato, mozzarella, and basil.  
The second image has substantial color jittering, but since the structural alignment and composition remain consistent, the similarity remains high.}
\end{tcolorbox}

\end{tcolorbox}
\end{minipage}
\hfill
\begin{minipage}{0.48\textwidth}
\begin{tcolorbox}[enhanced,attach boxed title to top center={yshift=-3mm,yshifttext=-1mm},
  colback=red!5!white,colframe=red!20!gray,colbacktitle=red!20!gray,
  title=\gptFouroEleven{} Response - Reversed Order,fonttitle=\bfseries,
  boxed title style={size=small,colframe=red!20!gray},
  equal height group=examplegroup]

\begin{center}
    \includegraphics[width=0.9\linewidth]{imgs/pair-burger-color2.png}
\end{center}

\textbf{Model response:}
\begin{tcolorbox}[colback=gray!10!white, colframe=gray!50, sharp corners=south, boxrule=0.8pt]
\textit{Score: 3\\Reason: The images depict similar objects, presumably a sandwich with layers, but there are notable differences.  
The first image appears distorted with altered coloration, and while it shares structural similarities with the second image,  
the overall resemblance is weak.}
\end{tcolorbox}

\end{tcolorbox}
\end{minipage}

\subsection{InternVL2.5 Model}
Below is an error example of \internvlTwoFiveEightB{} on a data-pair from \mmscorecoco with that are irrelevant to each other.

\begin{tcolorbox}[enhanced,attach boxed title to top center={yshift=-3mm,yshifttext=-1mm},
  colback=red!5!white,colframe=red!20!gray,colbacktitle=red!20!gray,
  title=\internvlTwoFiveEightB{} Evaluation Instructions,fonttitle=\bfseries,
  boxed title style={size=small,colframe=red!20!gray} ]

\textbf{User prompt}: Evaluate the similarity of the images based on the following conditions: 
\emph{- Be invariant to Gaussian blurs while evaluating images. For example, even if both images are the exact same, even if one is more blurred compared to the other, the similarity score should remain high. If the images are different, the similarity score should decrease regardless of how blurred they are.}  
The score should take into account how well the images align with these conditions, as well as their overall resemblance.  
Even if the images are identical but fail to meet the conditions, they should still receive a higher score than completely different images.  

Provide a score from 1 to 10, where 1 indicates no similarity and 10 indicates identical images that fully satisfy the conditions.  
Respond only in this format:

\begin{verbatim}
Score: <1-10>
Reason: <reason for score>
\end{verbatim}

Nothing else should be included. What score would you give? ``\texttt{data1}'' ``\texttt{data2}'' 
\end{tcolorbox}

\noindent
\begin{minipage}{0.48\textwidth}
\begin{tcolorbox}[enhanced,attach boxed title to top center={yshift=-3mm,yshifttext=-1mm},
  colback=red!5!white,colframe=red!20!gray,colbacktitle=red!20!gray,
  title=\internvlTwoFiveEightB{} - Original Order,fonttitle=\bfseries,
  boxed title style={size=small,colframe=red!20!gray},
  equal height group=examplegroup]

\begin{center}
    \includegraphics[width=0.9\linewidth]{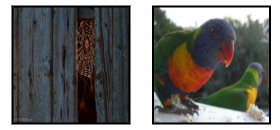}
\end{center}

\textbf{Model response:}
\begin{tcolorbox}[colback=gray!10!white, colframe=gray!50, sharp corners=south, boxrule=0.8pt]
\textit{Score: 7\\Reason: The images are similar in content, featuring parrots, but they are not identical.  
The first image is a close-up of a spider web on a wooden surface, while the second image shows two parrots on a ledge.  
The parrots are the main subjects in the second image, and the first image has a different focus and subject matter.  
The images do not meet the condition of being identical, but they share some visual elements like color and subject type.}
\end{tcolorbox}

\end{tcolorbox}
\end{minipage}
\hfill
\begin{minipage}{0.48\textwidth}
\begin{tcolorbox}[enhanced,attach boxed title to top center={yshift=-3mm,yshifttext=-1mm},
  colback=red!5!white,colframe=red!20!gray,colbacktitle=red!20!gray,
  title=\internvlTwoFiveEightB{} - Reverse Order,fonttitle=\bfseries,
  boxed title style={size=small,colframe=red!20!gray},
  equal height group=examplegroup]

\begin{center}
    \includegraphics[width=0.9\linewidth]{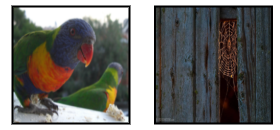}
\end{center}

\textbf{Model response:}
\begin{tcolorbox}[colback=gray!10!white, colframe=gray!50, sharp corners=south, boxrule=0.8pt]
\textit{Score: 1\\Reason: The images are completely different in terms of content, subject matter, and context.  
The first image features colorful parrots, while the second image shows a wooden door with a spider web.  
There is no similarity in terms of subject, color, or composition.}
\end{tcolorbox}

\end{tcolorbox}
\end{minipage}

\subsection{Pixtral Model}

Below is an error example of \pixtral{} on a data-pair from \mmscorecoco with rotation (R).

\begin{tcolorbox}[enhanced,attach boxed title to top center={yshift=-3mm,yshifttext=-1mm},
  colback=red!5!white,colframe=red!20!gray,colbacktitle=red!20!gray,
  title=\pixtral{} Evaluation Instructions,fonttitle=\bfseries,
  boxed title style={size=small,colframe=red!20!gray} ]

\textbf{User prompt}: Rate the similarity of these images on a scale of 1 to 10 while considering the following condition:  

\emph{- Be invariant to rotations while evaluating images. For example, if one image is a rotated version of the other, the similarity score should remain high. If the images are different, the similarity score should decrease regardless of their orientation.}  

The score should account for both how well the images satisfy the conditions and their overall resemblance. If the images are identical but do not meet the conditions, they should still score higher than unrelated images.  

Your response must follow this exact format:
\begin{verbatim}
Score: <1-10>
Reason: <reason for score>
\end{verbatim}

Provide nothing else. What’s your score? ``\texttt{data1}'' ``\texttt{data2}'' 
\end{tcolorbox}

\noindent
\begin{minipage}{0.48\textwidth}
\begin{tcolorbox}[enhanced,attach boxed title to top center={yshift=-3mm,yshifttext=-1mm},
  colback=red!5!white,colframe=red!20!gray,colbacktitle=red!20!gray,
  title=\pixtral{} Response - Original Order, fonttitle=\bfseries,
  boxed title style={size=small,colframe=red!20!gray},
  equal height group=examplegroup]

\begin{center}
    \includegraphics[width=0.9\linewidth]{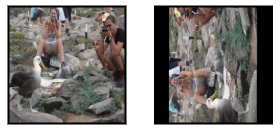}
\end{center}

\textbf{Model response:}
\begin{tcolorbox}[colback=gray!10!white, colframe=gray!50, sharp corners=south, boxrule=0.8pt]
\textit{Score: 8\\Reason: The images depict similar scenes of people interacting with birds in natural settings.  
The first image features a group of people sitting on rocks with birds nearby, while the second image has a person extending their hand  
towards a bird in a mountainous area. Despite slight differences in the environment and specific actions,  
the overall theme and presence of birds and human interaction score high.}
\end{tcolorbox}

\end{tcolorbox}
\end{minipage}
\hfill
\begin{minipage}{0.48\textwidth}
\begin{tcolorbox}[enhanced,attach boxed title to top center={yshift=-3mm,yshifttext=-1mm},
  colback=red!5!white,colframe=red!20!gray,colbacktitle=red!20!gray,
  title=\pixtral{} Response - Reversed Order,fonttitle=\bfseries,
  boxed title style={size=small,colframe=red!20!gray},
  equal height group=examplegroup]

\begin{center}
    \includegraphics[width=0.9\linewidth]{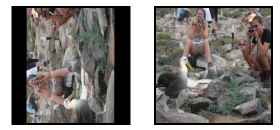}
\end{center}

\textbf{Model response:}
\begin{tcolorbox}[colback=gray!10!white, colframe=gray!50, sharp corners=south, boxrule=0.8pt]
\textit{Score: 2\\Reason: The images depict completely different scenes and subjects, with no recognizable similarity between them.  
One image shows a bird and people in a rocky area, while the other does not bear any resemblance to it.}
\end{tcolorbox}

\end{tcolorbox}
\end{minipage}

\subsection{Qwen2-VL Model}
Below is an error example of \qwenTwoVLSevenB{} on a data-pair from \mmscorecoco with gaussian blur (GB).

\begin{tcolorbox}[enhanced,attach boxed title to top center={yshift=-3mm,yshifttext=-1mm},
  colback=red!5!white,colframe=red!20!gray,colbacktitle=red!20!gray,
  title=\qwenTwoVLSevenB{} Evaluation Instructions,fonttitle=\bfseries,
  boxed title style={size=small,colframe=red!20!gray} ]

\textbf{User prompt}: As a similarity evaluator, your responsibility is to assess the similarity of the given images while considering these conditions:  
\emph{- Be sensitive to Gaussian blurs while evaluating images. For example, if both images are the same except one is a blurred version of the other, the similarity score should decrease. If the images are the same and have the same exact sharpness, the similarity score should remain high.}  

The similarity score should reflect both how well the images align with the conditions and their overall resemblance.  
Images that are identical but do not meet the conditions should receive a moderate score, while completely unrelated images should receive the lowest score.  

Provide your score on a scale of 1 to 10, with 10 being identical images that fully meet the conditions.  
Ensure your response is in the following format:
\begin{verbatim}
Score: <1-10>
Reason: <reason for score>
\end{verbatim}

Provide nothing else. What is your score? ``\texttt{data1}'' ``\texttt{data2}'' 
\end{tcolorbox}

\noindent
\begin{minipage}{0.48\textwidth}
\begin{tcolorbox}[enhanced,attach boxed title to top center={yshift=-3mm,yshifttext=-1mm},
  colback=red!5!white,colframe=red!20!gray,colbacktitle=red!20!gray,
  title=\qwenTwoVLSevenB{} - Original Order,fonttitle=\bfseries,
  boxed title style={size=small,colframe=red!20!gray},
  equal height group=examplegroup]

\begin{center}
    \includegraphics[width=0.9\linewidth]{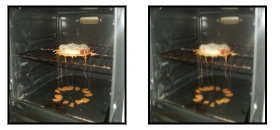}
\end{center}

\textbf{Model response:}
\begin{tcolorbox}[colback=gray!10!white, colframe=gray!50, sharp corners=south, boxrule=0.8pt]
\textit{Score: 6\\Reason: The images are very similar, with the second image being a slightly blurred version of the first.  
However, they do not fully meet the conditions as the images are not identical and the blurring is not perfect.}
\end{tcolorbox}

\end{tcolorbox}
\end{minipage}
\hfill
\begin{minipage}{0.48\textwidth}
\begin{tcolorbox}[enhanced,attach boxed title to top center={yshift=-3mm,yshifttext=-1mm},
  colback=red!5!white,colframe=red!20!gray,colbacktitle=red!20!gray,
  title=\qwenTwoVLSevenB{} - Reverse Order,fonttitle=\bfseries,
  boxed title style={size=small,colframe=red!20!gray},
  equal height group=examplegroup]

\begin{center}
    \includegraphics[width=0.9\linewidth]{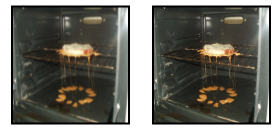}
\end{center}

\textbf{Model response:}
\begin{tcolorbox}[colback=gray!10!white, colframe=gray!50, sharp corners=south, boxrule=0.8pt]
\textit{Score: 9\\Reason: The images are nearly identical, with only a slight difference in the sharpness of the objects.  
The overall resemblance is high, and the conditions are met.}
\end{tcolorbox}

\end{tcolorbox}
\end{minipage}

\subsection{Phi3.5 Model}
Below is an error example of \phiThreeFive{} on a data-pair from \mmscorecoco with elastic transform (ET).

\begin{tcolorbox}[enhanced,attach boxed title to top center={yshift=-3mm,yshifttext=-1mm},
  colback=red!5!white,colframe=red!20!gray,colbacktitle=red!20!gray,
  title=\phiThreeFive{} Evaluation Instructions,fonttitle=\bfseries,
  boxed title style={size=small,colframe=red!20!gray} ]

\textbf{User prompt}: Judge the similarity of these images based on:  
\emph{- Be sensitive to elastic transformations while evaluating images. For example, if the images are the same but have differing distortions due to elastic transformations, the similarity score should decrease. If the images are identical and have matching elastic transformations, the similarity score should remain high.}  

The similarity score should reflect both the overall resemblance of the images and how well they satisfy the conditions.  
Identical images that do not meet the conditions should still score higher than completely unrelated images.  

Provide a score on a scale of 1 to 10, with 1 being no similarity and 10 being identical images that satisfy all conditions.  
Respond strictly in this format:
\begin{verbatim}
Score: <1-10>
Reason: <reason for score>
\end{verbatim}

Do not include additional text. What’s your rating? ``\texttt{data1}'' ``\texttt{data2}'' 
\end{tcolorbox}

\noindent
\begin{minipage}{0.48\textwidth}
\begin{tcolorbox}[enhanced,attach boxed title to top center={yshift=-3mm,yshifttext=-1mm},
  colback=red!5!white,colframe=red!20!gray,colbacktitle=red!20!gray,
  title=\phiThreeFive{} - Original Order,fonttitle=\bfseries,
  boxed title style={size=small,colframe=red!20!gray},
  equal height group=examplegroup]

\begin{center}
    \includegraphics[width=0.9\linewidth]{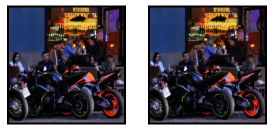}
\end{center}

\textbf{Model response:}
\begin{tcolorbox}[colback=gray!10!white, colframe=gray!50, sharp corners=south, boxrule=0.8pt]
\textit{Score: 1\\Reason: The images are highly distorted with elastic transformations, making it impossible to accurately compare the content.}
\end{tcolorbox}

\end{tcolorbox}
\end{minipage}
\hfill
\begin{minipage}{0.48\textwidth}
\begin{tcolorbox}[enhanced,attach boxed title to top center={yshift=-3mm,yshifttext=-1mm},
  colback=red!5!white,colframe=red!20!gray,colbacktitle=red!20!gray,
  title=\phiThreeFive{} - Reverse Order,fonttitle=\bfseries,
  boxed title style={size=small,colframe=red!20!gray},
  equal height group=examplegroup]

\begin{center}
    \includegraphics[width=0.9\linewidth]{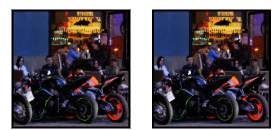}
\end{center}

\textbf{Model response:}
\begin{tcolorbox}[colback=gray!10!white, colframe=gray!50, sharp corners=south, boxrule=0.8pt]
\textit{Score: 10\\Reason: The images are identical with no differences in content, color, or composition.  
They both depict the same scene with motorcycles and people in a bar setting, and the elastic transformations do not alter the image in any way.}
\end{tcolorbox}

\end{tcolorbox}
\end{minipage}

\section{Related Work}
\label{sec:related-work-full}
% [what are some different benchmarks]

% [how are people evaluating vlms]

% [encoders are also being used as judges (text-image and image-image)]

% [what benchmarks are using vlms as judges]

% [literature on looking into invariance in LMMs]

% [downsides of using clip and other models (do not get negation), look at it as bag of words, spatially flawed (whatsup), looking into propperties that are not captures in benchmarks]
Using language models as automatic evaluators has become a somewhat common practice with popular approaches such as \textsc{GPTScore} and G-eval~\citep{fu2023gptscore,liu2023geval} being used to rank responses in the NLP domain. 
Due to that, there has been a significant amount of recent work that has investigated the capabilities and limitations of using LLMs as judges~\citep{thakur2024judging, chiang2023can, murugadoss2024evaluating, shankar2024validates}. 
\citet{chiang2023can} have shown that LLM evaluations are consistent and reproducible, making them suitable alternatives for human evaluation, they argue that these models inherent biases should prevent them using independently rather than \textit{alongside} human experts. 
Furthermore, \citet{zheng2023judging} reveal that large \modelss{{}, e.g., GPT-4 Turbo, align well with human judgments and \citet{thakur2024judging} further states that simpler models may still outperform GPT-4 Turbo in ranking tasks due to superior alignment metrics. Also, recent work assessed how humans can help LLMs evaluate better by testing different instruction types or designing tools that result in more balanced evaluations~\citep{murugadoss2024evaluating, shankar2024validates}.

It is worth noting that known limitations of LLMs such as their lack of invariance to the order of examples given in a prompt, which is a well studied issue of natural language models~\citep{fang2024rethinking}, 
% and also observed in the multimodal case by the lack of symmetry we reported,
may render auto evaluation unreliable. Similarly, \citet{berglund2023reversal} show failure cases where models trained on unidirectional relationships do not infer the reverse, indicating systemic limitations even in state-of-the-art LLMs such as \textsc{GPT-4} (as seen in Figure~\ref{fig:fig1} and in Appendix \ref{sec:error-analysis} for \modelss{}). Our main goal is
to investigate the reliability of automated evaluation in the multimodal context, by probing the models to compare data pairs. 
%thus to assess to what extent auto-evaluation can be done reliably, focusing in the multimodal case, by probing models on their ability to compare.

Namely, the evaluations we carry out focus on testing in multiple different ways how good \modelss{} are when it comes to comparing data instances, such as whether \modelss{} prompted to compare are symmetric or smooth for instance, and to what extent they can be controlled, i.e., instructed to pay attention to or ignore certain features of the inputs. While the literature is more sparse regarding testing \modelss{} in this setting, recent work has tested for something along those lines. \citet{chen2024mllm} for instance propose a benchmark for evaluating \modelss{} in multiple different scenarios, including checking whether pairwise comparisons of responses to a query correlated with human judgments. They concluded that although correlations are relatively high on comparison tasks, biases and inconsistencies affect performance on pair scoring and batch ranking. Similarly, \citet{awal2024vismin} introduced a synthetic dataset containing paired images that differ only along one feature (e.g., the color of an object). We seek to add to this branch of the literature by introducing a framework where controlled experiments can be carried out to anticipate the performance of models when being used as judges, and various different characteristics of automatic judges can be identified (e.g., how smooth they are).

Unlike the case of generative \modelss{} discussed above, discriminative visual language models such as CLIP~\citep{radford2021learning} are covered by a significant amount of recent work, and several failure modes are well reported, mostly deriving from the fact this class of \modelss{} tends to behave as bag-of-words models, focusing on nouns and ignoring relationships and semantics in their input data~\citep{yuksekgonul2023when}. For instance, CLIP was observed to struggle with spatial reasoning~\citep{kamath2023s} and ignore negation~\citep{alhamoud2025vision}. On the other hand, fine-tuning CLIP to reason about pairwise differences \citet{sam2024finetuning} showed that discriminative \modelss{} can improve on how well they manage to reason about pairwise differences if training is tailored for enabling so, highlighting the benefits that being able to measure these skills may inform training and improve models as a consequence. \citet{ouali2024discriminative} showed that fine-tuning generative \modelss{} to turn them into discriminative models results in improved image-retrieval from text, which aligns with results we reported in Section~\ref{sec:ecoders_vs_decoders} showing a gap between open-sources \modelss{} and CLIP-style encoders.

\FloatBarrier
\section{Full Results}
\label{sec:full-results}

In this section, we provide the \nmi{} of all models on all the different splits of \mmscorecoco, \mmscorein, \mmscorewuimgimg, and \mmscorewuimgtext{} in Tables \ref{tab:mi-coco-in100-var}, \ref{tab:mi-coco-in100-invar}, \ref{tab:mi-coco-in100-var}, \ref{tab:mi-coco-in100-invar}, \ref{tab:mi-wu-imgimg-var}, \ref{tab:mi-wu-imgimg-invar}, and \ref{tab:mi-wu-imgtext}. We further report the coverage, the number of times the \modelss{} give a valid output, of each model on our different proposed datasets.

\begin{table*}[ht]
\centering
\caption{Comparison of the \nmi{} metric ($\times 100$) of \modelss{} on \mmscorecoco{} and \mmscorein{} benchmarks in the \textit{sensitive} setting. Models are evaluated across multiple criteria:  color jitter (CJ), elastic transform (ET), gaussian blur (GB), perspective shift (PS), and rotation (R). Higher scores indicate better performance.}
\resizebox{0.95\textwidth}{!}{%
\begin{tabular}{l*{5}{c}*{5}{c}}
\toprule
\multirow{2}{*}{\textbf{Model}} & \multicolumn{5}{c}{\textbf{\mmscorecoco}} & \multicolumn{5}{c}{\textbf{\mmscorein}} \\
        \cmidrule(lr){2-6} \cmidrule(lr){7-11} & \textbf{CJ} & \textbf{ET} & \textbf{GB} & \textbf{PS} & \textbf{R} &  \textbf{CJ} & \textbf{ET} & \textbf{GB} & \textbf{PS} & \textbf{R} \\
\midrule
% model & coco &  &  &  &  & in100 &  &  &  &  \\
%  & CJ & ET & GB & PS & R & CJ & ET & GB & PS & R \\
\chameleon & 00.37 & 00.34 & 00.19 & 00.31 & 00.60 & 00.38 & 00.26 & 00.31 & 00.50 & 00.52 \\
\llavaonevision & 36.51 & 44.05 & 38.57 & 43.80 & 41.41 & 37.05 & 49.89 & 40.00 & 46.01 & 49.30 \\
\phiThreeFive & 38.21 & 51.61 & 61.94 & 47.33 & 34.56 & 25.74 & 43.03 & 51.40 & 32.51 & 23.61 \\
\pixtral & 37.67 & 56.25 & 54.32 & 49.53 & 36.80 & 30.75 & 52.30 & 51.94 & 46.04 & 40.76 \\
% \rowcolor{blue!15}
\internvlTwoOneB & 03.23 & 03.47 & 03.27 & 03.63 & 03.51 & 02.59 & 02.38 & 01.70 & 02.02 & 02.23 \\
% \rowcolor{blue!15}
\internvlTwoTwoB & 23.89 & 32.76 & 34.32 & 31.53 & 24.76 & 18.32 & 34.02 & 33.35 & 28.17 & 23.35 \\
% \rowcolor{blue!15}
\internvlTwoFourB & 52.13 & 69.43 & 62.46 & 63.77 & 52.68 & 45.25 & 65.90 & 59.90 & 60.28 & 51.04 \\
% \rowcolor{blue!15}
\internvlTwoEightB & 51.58 & 62.80 & 62.35 & 60.27 & 54.80 & 47.94 & 60.18 & 58.60 & 56.66 & 53.00 \\
% \rowcolor{purple!15}
\internvlTwoFiveOneB & 16.74 & 25.38 & 27.67 & 24.83 & 16.54 & 15.63 & 33.67 & 39.23 & 37.97 & 22.53 \\
% \rowcolor{purple!15}
\internvlTwoFiveTwoB & 12.48 & 19.58 & 25.26 & 18.33 & 13.84 & 17.27 & 38.28 & 39.21 & 31.23 & 21.45 \\
% \rowcolor{purple!15}
\internvlTwoFiveFourB & 42.61 & 59.78 & 54.33 & 55.34 & 49.47 & 41.35 & 62.35 & 54.21 & 56.18 & 49.90 \\
% \rowcolor{purple!15}
\internvlTwoFiveEightB & 54.51 & 73.37 & 78.31 & 63.17 & 60.71 & 51.76 & 77.10 & 76.40 & 60.40 & 55.30 \\
% \rowcolor{orange!15}
\molmoEOneB & 00.40 & 00.09 & 01.20 & 00.03 & 00.05 & 00.41 & 00.01 & 00.45 & 00.01 & 00.01 \\
% \rowcolor{orange!15}
\molmoOSevenB & 14.32 & 16.02 & 48.93 & 16.12 & 15.40 & 12.91 & 14.20 & 48.43 & 13.83 & 12.16 \\
% \rowcolor{orange!15}
\molmoDSevenB & 27.06 & 45.28 & 34.46 & 49.60 & 30.39 & 22.88 & 41.06 & 35.83 & 44.49 & 32.22 \\
% \rowcolor{yellow!15}
\qwenTwoVLTwoB & 09.91 & 11.82 & 09.01 & 13.13 & 11.95 & 10.63 & 13.69 & 10.41 & 13.21 & 12.23 \\
% \rowcolor{yellow!15}
\qwenTwoVLSevenB & 42.58 & 61.90 & 50.22 & 55.81 & 51.10 & 38.24 & 61.73 & 50.23 & 53.07 & 52.29 \\
\midrule
% \rowcolor{green!15}
\gptFouroMini & 49.98 & 65.97 & 58.29 & 53.23 & 53.60 & 47.06 & 67.06 & 56.43 & 49.97 & 52.59 \\
% \rowcolor{green!15}
\gptFouroFive & 50.96 & 65.54 & 61.67 & 56.69 & 56.71 & 48.55 & 65.68 & 57.48 & 54.11 & 55.00 \\
% \rowcolor{green!15}
\gptFouroEight & 42.26 & 60.58 & 56.62 & 50.13 & 53.63 & 40.35 & 60.66 & 52.65 & 49.62 & 49.77 \\
% \rowcolor{green!15}
\gptFouroEleven & 51.31 & 63.50 & 61.35 & 57.84 & 57.16 & 50.88 & 66.55 & 58.14 & 56.25 & 55.52 \\
% \rowcolor{green!30}
\geminiFlash & \textbf{58.26} & 82.64 & 87.41 & 65.92 & 61.08 & \textbf{56.25} & 79.69 & 85.21 & 62.07 & 61.15 \\
% \rowcolor{green!30}
\geminiPro & 53.33 & \textbf{87.86} & \textbf{89.56} & \textbf{74.92} & \textbf{71.04} & 51.19 & \textbf{91.36} & \textbf{92.98} & \textbf{71.56} & \textbf{74.22} \\

\bottomrule
\end{tabular}
}
\label{tab:mi-coco-in100-var}
\end{table*}

\begin{table*}[ht]
\centering
\caption{Comparison of the \nmi{} metric ($\times 100$) of \modelss{} on \mmscorecoco{} and \mmscorein{} benchmarks in the \textit{invariant} setting. Models are evaluated across multiple criteria: color jitter (CJ), elastic transform (ET), gaussian blur (GB), perspective shift (PS), and rotation (R). Higher scores indicate better performance.}
\resizebox{0.95\textwidth}{!}{%
\begin{tabular}{l*{5}{c}*{5}{c}}
\toprule
\multirow{2}{*}{\textbf{Model}} & \multicolumn{5}{c}{\textbf{\mmscorecoco}} & \multicolumn{5}{c}{\textbf{\mmscorein}} \\
        \cmidrule(lr){2-6} \cmidrule(lr){7-11} & \textbf{CJ} & \textbf{ET} & \textbf{GB} & \textbf{PS} & \textbf{R} &  \textbf{CJ} & \textbf{ET} & \textbf{GB} & \textbf{PS} & \textbf{R} \\
\midrule
% model & coco &  &  &  &  & in100 &  &  &  &  \\
%  & CJ & ET & GB & PS & R & CJ & ET & GB & PS & R \\
\chameleon & 00.89 & 00.34 & 00.44 & 00.51 & 00.38 & 00.57 & 00.35 & 00.53 & 00.58 & 00.45 \\
\llavaonevision & 35.13 & 37.26 & 39.22 & 40.29 & 38.29 & 38.09 & 43.04 & 41.83 & 40.86 & 42.24 \\
\phiThreeFive & 49.41 & 40.19 & 42.93 & 55.03 & 47.90 & 45.88 & 33.79 & 39.72 & 50.41 & 39.46 \\
\pixtral  & 48.26 & 47.34 & 45.35 & 60.20 & 55.65 & 41.53 & 45.30 & 42.84 & 52.63 & 52.65 \\
% \rowcolor{blue!15}
\internvlTwoOneB & 02.69 & 01.76 & 02.71 & 02.00 & 02.69 & 01.39 & 00.82 & 01.22 & 00.90 & 01.40 \\
% \rowcolor{blue!15}
\internvlTwoTwoB & 36.38 & 31.55 & 31.99 & 39.18 & 37.28 & 32.68 & 31.40 & 30.13 & 35.98 & 34.70 \\
% \rowcolor{blue!15}
\internvlTwoFourB & 59.44 & 55.47 & 51.35 & 59.61 & 59.02 & 51.74 & 52.77 & 49.60 & 54.63 & 53.11 \\
% \rowcolor{blue!15}
\internvlTwoEightB & 58.69 & 58.56 & 53.60 & 61.91 & 64.22 & 58.44 & 54.48 & 51.78 & 61.97 & 62.90 \\
% \rowcolor{purple!15}
\internvlTwoFiveOneB  & 21.39 & 18.59 & 21.65 & 23.19 & 22.86 & 22.52 & 14.63 & 24.34 & 22.76 & 19.24 \\
% \rowcolor{purple!15}
\internvlTwoFiveTwoB & 22.85 & 19.05 & 21.46 & 27.62 & 25.99 & 32.09 & 33.03 & 37.34 & 34.65 & 34.75 \\
% \rowcolor{purple!15}
\internvlTwoFiveFourB & 56.24 & 47.41 & 43.93 & 53.71 & 55.28 & 61.80 & 50.50 & 47.33 & 51.58 & 58.56 \\
% \rowcolor{purple!15}
\internvlTwoFiveEightB & \textbf{75.11} & 65.18 & 66.32 & \textbf{78.56} & \textbf{81.77} & \textbf{72.53} & 61.61 & 62.23 & 65.18 & 74.27 \\
% \rowcolor{orange!15}
\molmoEOneB & 00.10 & 00.11 & 00.06 & 00.02 & 00.00 & 00.02 & 00.11 & 00.10 & 00.07 & 00.25 \\
% \rowcolor{orange!15}
\molmoOSevenB & 26.86 & 34.58 & 33.46 & 34.70 & 24.55 & 25.04 & 30.81 & 38.52 & 32.79 & 27.65 \\
% \rowcolor{orange!15}
\molmoDSevenB & 47.20 & 45.02 & 43.02 & 50.54 & 48.64 & 45.01 & 45.83 & 45.47 & 49.25 & 40.87 \\
% \rowcolor{yellow!15}
\qwenTwoVLTwoB & 09.55 & 09.10 & 10.21 & 12.65 & 08.83 & 09.02 & 09.61 & 10.01 & 14.97 & 09.33 \\
% \rowcolor{yellow!15}
\qwenTwoVLSevenB & 50.52 & 51.80 & 52.70 & 54.50 & 53.29 & 47.86 & 49.73 & 51.18 & 51.55 & 50.67 \\
\midrule
% \rowcolor{green!15}
\gptFouroMini & 59.76 & 57.94 & 56.55 & 61.31 & 58.17 & 56.33 & 55.56 & 55.35 & 60.99 & 60.83 \\
% \rowcolor{green!15}
\gptFouroFive & 70.83 & 61.70 & 59.40 & 61.13 & 62.10 & 68.82 & 56.16 & 56.70 & 57.79 & 59.80 \\
% \rowcolor{green!15}
\gptFouroEight & 55.14 & 50.31 & 46.00 & 52.15 & 52.45 & 54.13 & 45.43 & 44.25 & 48.26 & 52.18 \\
% \rowcolor{green!15}
\gptFouroEleven & 73.48 & 69.06 & 61.51 & 67.60 & 63.99 & 70.16 & 61.33 & 58.89 & 65.06 & 60.84 \\
% \rowcolor{green!30}
\geminiFlash & 72.11 & 67.81 & 68.17 & 71.88 & 78.31 & 70.32 & 65.94 & 66.58 & 69.10 & 74.77 \\
% \rowcolor{green!30}
\geminiPro & 68.93 & \textbf{69.64} & \textbf{71.50} & 72.06 & 68.42 & 66.31 & \textbf{70.03} & \textbf{72.17} & \textbf{70.13} & \textbf{69.32} \\

\bottomrule
\end{tabular}
}
\label{tab:mi-coco-in100-invar}
\end{table*}

\begin{table*}[ht]
\centering
\caption{Comparison of the \nmi{} metric ($\times 100$) of \modelss{} on \mmscorewuimgimg{} (subset A and B) benchmark in the \textit{sensitive} setting. Models are evaluated across multiple criteria: spatial position (SP), spatial position and color jitter (SP-CJ), spatial position and elastic transform (SP-ET), spatial position and gaussian blur (SP-GB), spatial position and perspective shift (SP-PS), and spatial position and rotation (SP-R). Higher scores indicate better performance.}
\resizebox{0.95\textwidth}{!}{%
\begin{tabular}{l*{6}{c}*{6}{c}}
\toprule
\multirow{2}{*}{\textbf{Model}} & \multicolumn{6}{c}{\textbf{\mmscore$_{WU_a}$}} & \multicolumn{6}{c}{\textbf{\mmscore$_{WU_b}$}} \\
        \cmidrule(lr){2-7} \cmidrule(lr){8-13} & \textbf{SP} & \textbf{SP-CJ} & \textbf{SP-ET} & \textbf{SP-GB} & \textbf{SP-PS} & \textbf{SP-R} & \textbf{SP} &  \textbf{SP-CJ} & \textbf{SP-ET} & \textbf{SP-GB} & \textbf{SP-PS} & \textbf{SP-R}\\
\midrule
% model & coco &  &  &  &  & in100 &  &  &  &  \\
%  & CJ & ET & GB & PS & R & CJ & ET & GB & PS & R \\
\chameleon & 00.28 & 00.47 & 00.23 & 00.52 & 0.21 & 00.20 & 00.34 & 00.38 & 00.35 & 00.26 & 00.31 & 00.33 \\
\llavaonevision & 38.95 & 18.83 & 24.03 & 26.78 & 29.46 & 24.63 & 19.70 & 14.03 & 16.51 & 16.78 & 17.76 & 17.02 \\
\phiThreeFive & 23.44 & 08.46 & 15.70 & 19.41 & 13.34 & 10.83 & 15.38 & 12.98 & 18.91 & 20.19 & 11.69 & 17.06 \\
\pixtral & 37.91 & 26.09 & 32.05 & 33.52 & 32.47 & 25.00 & 28.02 & 19.58 & 22.32 & 22.31 & 23.46 & 24.50 \\
% \rowcolor{blue!15}
\internvlTwoOneB & 00.44 & 00.98 & 00.79 & 00.65 & 00.30 & 00.28 & 00.20 & - & - & 00.41 & 01.18 & 00.90 \\
% \rowcolor{blue!15}
\internvlTwoTwoB & 22.85 & 12.03 & 14.37 & 17.84 & 18.66 & 15.50 & 20.72 & 10.89 & 11.22 & 15.74 & 17.74 & 13.58 \\
% \rowcolor{blue!15}
\internvlTwoFourB & 46.89 & 27.91 & 36.67 & 43.03 & 44.27 & 27.76 & 44.89 & 27.77 & 33.35 & 38.12 & 42.23 & 36.16 \\
% \rowcolor{blue!15}
\internvlTwoEightB & 41.99 & 32.06 & 35.71 & 41.02 & 40.12 & 29.11 & 46.36 & 32.17 & 39.24 & 41.90 & 45.59 & 40.30 \\
% \rowcolor{purple!15}
\internvlTwoFiveOneB & 25.50 & 14.16 & 21.32 & 15.69 & 21.49 & 16.30 & 24.77 & 16.16 & 21.10 & 19.95 & 27.89 & 21.47 \\
% \rowcolor{purple!15}
\internvlTwoFiveTwoB & 20.63 & 11.76 & 16.75 & 15.21 & 18.03 & 13.79 & 23.44 & 09.33 & 15.90 & 17.64 & 18.17 & 17.56 \\
% \rowcolor{purple!15}
\internvlTwoFiveFourB & 46.15 & 32.74 & 39.05 & 39.24 & 42.28 & 32.94 & 47.93 & 33.75 & 40.23 & 39.82 & 44.07 & 42.57 \\
% \rowcolor{purple!15}
\internvlTwoFiveEightB & 44.27 & 36.99 & 41.49 & 42.60 & 43.65 & 33.24 & 41.32 & 31.69 & 40.10 & 39.73 & 44.03 & 42.99 \\
% \rowcolor{orange!15}
\molmoEOneB & 00.47 & 01.03 & 00.00 & 00.03 & 00.14 & 00.01 & 00.32 & 00.36 & 00.01 & 00.04 & 00.04 & 00.09 \\
% \rowcolor{orange!15}
\molmoOSevenB & 15.94 & 09.90 & 11.32 & 15.38 & 12.92 & 12.01 & 15.15 & 08.40 & 11.39 & 11.33 & 13.60 & 12.50 \\
% \rowcolor{orange!15}
\molmoDSevenB & 23.82 & 17.75 & 20.41 & 18.40 & 22.21 & 17.81 & 26.74 & 18.37 & 19.55 & 18.77 & 18.19 & 22.21 \\
% \rowcolor{yellow!15}
\qwenTwoVLTwoB & 02.26 & 01.76 & 02.58 & 02.15 & 03.17 & 01.68 & 00.88 & 00.44 & 00.73 & 00.37 & 00.72 & 00.82 \\
% \rowcolor{yellow!15}
\qwenTwoVLSevenB & 41.95 & 29.47 & 36.32 & 39.93 & 40.33 & 34.11 & 42.80 & 28.75 & 31.42 & 37.27 & 39.76 & 36.25 \\
\midrule
% \rowcolor{green!15}
\gptFouroMini & 42.55 & 37.21 & 39.50 & 40.44 & 38.83 & 41.05 & 48.86 & 38.38 & 43.82 & 45.42 & 46.32 & 46.66 \\
% \rowcolor{green!15}
\gptFouroFive & 40.27 & 37.83 & 36.79 & 38.52 & 38.84 & 38.07 & 44.13 & 39.46 & 39.46 & 43.58 & 43.49 & 46.25 \\
% \rowcolor{green!15}
\gptFouroEight & 37.58 & 33.72 & 34.24 & 33.36 & 34.80 & 33.17 & 40.11 & 33.36 & 32.36 & 34.32 & 39.91 & 38.67 \\
% \rowcolor{green!15}
\gptFouroEleven & 40.68 & 39.06 & 40.10 & 40.35 & 40.96 & 40.40 & 47.34 & 40.91 & 43.07 & 47.18 & 50.22 & 50.68 \\
% \rowcolor{green!30}
\geminiFlash & 44.63 & 38.85 & 37.19 & 39.11 & 35.76 & 34.57 & 49.91 & 40.29 & 42.92 & 46.34 & 47.01 & 46.40 \\
% \rowcolor{green!30}
\geminiPro & 40.38 & 36.07 & 31.52 & 37.85 & 29.92 & 30.37 & 49.20 & 38.26 & 39.16 & 44.98 & 41.70 & 40.72 \\
\bottomrule
\end{tabular}
}
\label{tab:mi-wu-imgimg-var}
\end{table*}

\begin{table*}[ht]
\centering
\caption{Comparison of the \nmi{} metric ($\times 100$) of \modelss{} on \mmscorewuimgimg{} (subset A and B) benchmark in the \textit{invariant} setting. Models are evaluated across multiple criteria:spatial position (SP), spatial position and color jitter (SP-CJ), spatial position and elastic transform (SP-ET), spatial position and gaussian blur (SP-GB), spatial position and perspective shift (SP-PS), and spatial position and rotation (SP-R). Higher scores indicate better performance.}
\resizebox{0.95\textwidth}{!}{%
\begin{tabular}{l*{6}{c}*{6}{c}}
\toprule
\multirow{2}{*}{\textbf{Model}} & \multicolumn{6}{c}{\textbf{\mmscore$_{WU_a}$}} & \multicolumn{6}{c}{\textbf{\mmscore$_{WU_b}$}} \\
        \cmidrule(lr){2-7} \cmidrule(lr){8-13} & \textbf{SP} & \textbf{SP-CJ} & \textbf{SP-ET} & \textbf{SP-GB} & \textbf{SP-PS} & \textbf{SP-R} & \textbf{SP} &  \textbf{SP-CJ} & \textbf{SP-ET} & \textbf{SP-GB} & \textbf{SP-PS} & \textbf{SP-R} \\
\midrule
% model & coco &  &  &  &  & in100 &  &  &  &  \\
%  & CJ & ET & GB & PS & R & CJ & ET & GB & PS & R \\
\chameleon & 00.34 & 00.39 & 00.76 & 00.47 & 00.43 & 00.41 & 00.47 & 00.34 & 00.56 & 00.24 & 00.62 & 00.34 \\
\llavaonevision & 34.79 & 31.56 & 30.23 & 34.14 & 32.61 & 28.69 & 13.12 & 18.41 & 16.21 & 22.69 & 15.34 & 17.91 \\
\phiThreeFive & 23.66 & 32.84 & 18.90 & 21.36 & 30.14 & 19.10 & 19.88 & 36.74 & 22.40 & 23.47 & 30.04 & 26.06 \\
\pixtral & 36.93 & 37.32 & 41.17 & 35.31 & 38.52 & 36.05 & 36.03 & 30.44 & 33.32 & 29.84 & 35.48 & 33.32 \\
% \rowcolor{blue!15}
\internvlTwoOneB & 00.57 & 01.08 & 02.02 & 01.02 & 00.89 & 00.37 & 00.65 & 00.81 & 00.96 & 00.50 & 00.56 & 00.54 \\
% \rowcolor{blue!15}
\internvlTwoTwoB & 26.25 & 25.53 & 25.76 & 21.12 & 26.57 & 26.98 & 26.03 & 24.52 & 26.49 & 25.81 & 31.01 & 29.33 \\
% \rowcolor{blue!15}
\internvlTwoFourB & 39.33 & 40.23 & 37.80 & 42.25 & 43.10 & 34.57 & 51.43 & 41.55 & 45.96 & 50.20 & 54.94 & 50.34 \\
% \rowcolor{blue!15}
\internvlTwoEightB & 43.80 & 44.31 & 44.53 & 43.99 & 46.02 & 40.43 & 60.92 & 46.63 & 54.53 & 51.31 & 56.94 & 53.88 \\
% \rowcolor{purple!15}
\internvlTwoFiveOneB & 12.82 & 13.84 & 09.34 & 07.24 & 12.91 & 16.93 & 19.87 & 24.92 & 19.36 & 17.94 & 22.66 & 30.60 \\
% \rowcolor{purple!15}
\internvlTwoFiveTwoB & 31.38 & 29.79 & 30.53 & 23.16 & 31.75 & 24.69 & 36.01 & 30.13 & 35.52 & 27.07 & 37.01 & 31.18 \\
% \rowcolor{purple!15}
\internvlTwoFiveFourB & 48.79 & 53.58 & 54.52 & 48.09 & 52.78 & 46.46 & 50.51 & 48.71 & 53.45 & 52.03 & 53.77 & 50.12 \\
% \rowcolor{purple!15}
\internvlTwoFiveEightB & 59.03 & 55.57 & 59.70 & 57.16 & 58.01 & 50.84 & 65.21 & 51.31 & 61.10 & 63.54 & 62.38 & 60.83 \\
% \rowcolor{orange!15}
\molmoEOneB & 03.83 & 00.09 & 00.02 & 00.02 & 00.10 & 00.17 & 04.22 & 00.07 & 00.02 & 00.07 & 00.12 & 00.00 \\
% \rowcolor{orange!15}
\molmoOSevenB & 18.63 & 17.50 & 19.68 & 16.42 & 19.58 & 14.99 & 15.94 & 19.46 & 20.93 & 17.98 & 24.21 & 21.68 \\
% \rowcolor{orange!15}
\molmoDSevenB & 28.21 & 36.47 & 31.95 & 26.89 & 35.57 & 33.58 & 37.50 & 35.90 & 34.70 & 33.51 & 33.04 & 34.35 \\
% \rowcolor{yellow!15}
\qwenTwoVLTwoB & 02.63 & 02.88 & 03.58 & 03.53 & 03.34 & 02.97 & 00.79 & 00.73 & 00.99 & 00.88 & 00.71 & 00.82 \\
% \rowcolor{yellow!15}
\qwenTwoVLSevenB & 40.21 & 38.96 & 39.94 & 46.88 & 40.11 & 39.55 & 47.65 & 39.51 & 40.94 & 48.63 & 44.68 & 41.88 \\
\midrule
% \rowcolor{green!15}
\gptFouroMini & 47.60 & 48.33 & 51.04 & 46.15 & 48.86 & 43.75 & 57.50 & 49.19 & 51.38 & 53.76 & 55.82 & 54.07 \\
% \rowcolor{green!15}
\gptFouroFive & 52.39 & 51.58 & 48.78 & 47.11 & 47.50 & 52.68 & 61.59 & 59.77 & 58.08 & 60.95 & 61.53 & 63.74 \\
% \rowcolor{green!15}
\gptFouroEight & 50.94 & 47.21 & 46.52 & 42.90 & 45.84 & 52.50 & 62.75 & 54.23 & 53.20 & 51.19 & 58.50 & 57.21 \\
% \rowcolor{green!15}
\gptFouroEleven & 57.47 & 56.25 & 54.40 & 56.11 & 54.40 & 57.93 & 65.91 & 62.22 & 63.93 & 67.96 & 66.86 & 68.10 \\
% \rowcolor{green!30}
\geminiFlash & 46.62 & 55.28 & 54.31 & 57.98 & 57.01 & 58.74 & 62.04 & 54.43 & 56.89 & 62.24 & 66.88 & 60.72 \\
% \rowcolor{green!30}
\geminiPro & 38.07 & 35.08 & 35.05 & 36.11 & 33.21 & 33.23 & 56.43 & 42.24 & 43.74 & 48.41 & 50.40 & 45.83 \\
\bottomrule
\end{tabular}
}
\label{tab:mi-wu-imgimg-invar}
\end{table*}

\begin{table*}[ht]
\centering
\caption{Comparison of the \nmi{} metric ($\times 100$) of \modelss{} on the \mmscorewuimgtext{} (Subset A and B) benchmark in the \textit{sensitive} and \textit{invariant} settings. Models are evaluated across the spatial position (SP) criterion. Higher scores indicate better performance.}
\begin{tabular}{l*{2}{c}*{2}{c}}
\toprule
\multirow{2}{*}{\textbf{Model}} & \multicolumn{2}{c}{\textbf{\mmscore$_{WU_a}$}} & \multicolumn{2}{c}{\textbf{\mmscore$_{WU_b}$}} \\
        \cmidrule(lr){2-3} \cmidrule(lr){4-5} & \textbf{Sens.} & \textbf{Invar.} & \textbf{Sens.} & \textbf{Invar.} \\
\midrule
% model & coco &  &  &  &  & in100 &  &  &  &  \\
%  & CJ & ET & GB & PS & R & CJ & ET & GB & PS & R \\
\chameleon & 00.25 & 00.34 & 00.23 & 00.47 \\
\llavaonevision & 23.35 & 22.78 & 27.38 & 25.98  \\
\phiThreeFive & 13.86 & 12.30 & 25.67 & 24.74  \\
\pixtral & 05.14 & 05.04 & 03.27 & 04.58  \\
% \rowcolor{blue!15}
\internvlTwoOneB & 06.29 & 03.75 & 15.90 & 08.31  \\
% \rowcolor{blue!15}
\internvlTwoTwoB & 17.07 & 14.26 & 24.46 & 16.49  \\
% \rowcolor{blue!15}
\internvlTwoFourB & 15.69 & 15.69 & 24.27 & 22.96  \\
% \rowcolor{blue!15}
\internvlTwoEightB & 22.40 & 19.27 & 29.45 & 31.46  \\
% \rowcolor{purple!15}
\internvlTwoFiveOneB & 20.80 & 09.49 & 16.86 & 13.23  \\
% \rowcolor{purple!15}
\internvlTwoFiveTwoB & 15.36 & 11.15 & 19.69 & 18.42  \\
% \rowcolor{purple!15}
\internvlTwoFiveFourB & 23.90 & 23.85 & 29.75 & 32.45  \\
% \rowcolor{purple!15}
\internvlTwoFiveEightB & 24.16 & 25.55 & 24.00 & 28.22  \\
% \rowcolor{orange!15}
\molmoEOneB & 00.12 & 00.04 & 00.02 & 00.21  \\
% \rowcolor{orange!15}
\molmoOSevenB & 07.53 & 07.45 & 07.18 & 08.29  \\
% \rowcolor{orange!15}
\molmoDSevenB & 09.45 & 12.26 & 08.34 & 11.26  \\
% \rowcolor{yellow!15}
\qwenTwoVLTwoB & 02.65 & 03.09 & 05.09 & 05.86  \\
% \rowcolor{yellow!15}
\qwenTwoVLSevenB & 09.43 & 09.19 & 15.99 & 16.13  \\
\midrule 
% \rowcolor{green!15}
\gptFouroMini & 16.18 & 16.14 & 16.18 & 15.30  \\
% \rowcolor{green!15}
\gptFouroFive & 11.49 & 20.48 & 12.63 & 20.98  \\
% \rowcolor{green!15}
\gptFouroEight & 20.27 & 31.80 & 22.97 & 36.56  \\
% \rowcolor{green!15}
\gptFouroEleven & 18.97 & 31.91 & 20.57 & 34.99  \\
% \rowcolor{green!30}
\geminiFlash & 27.46 & 26.54 & 26.53 & 32.07  \\
% \rowcolor{green!30}
\geminiPro & 26.89 & 27.16 & 28.57 & 29.23  \\

\bottomrule
\end{tabular}
\label{tab:mi-wu-imgtext}
\end{table*}
\begin{figure*}[ht]
    \centering
    \includegraphics[width=0.95\linewidth,trim={.1cm .2cm .2cm .2cm},clip]{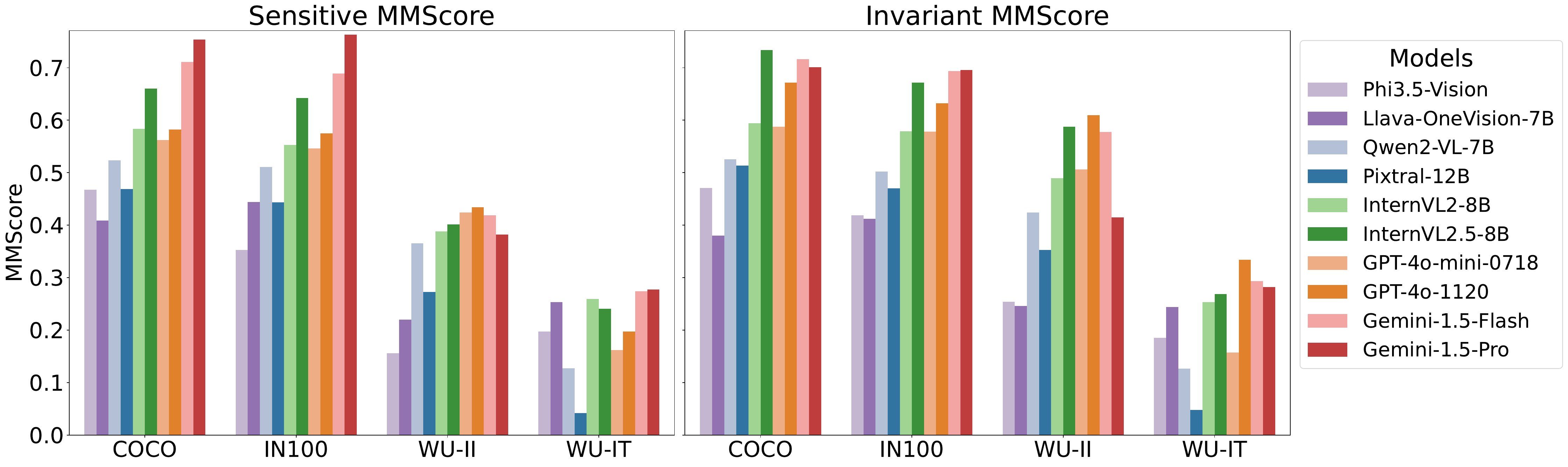}
    \includegraphics[width=0.95\linewidth,trim={.1cm .2cm .2cm .2cm},clip]{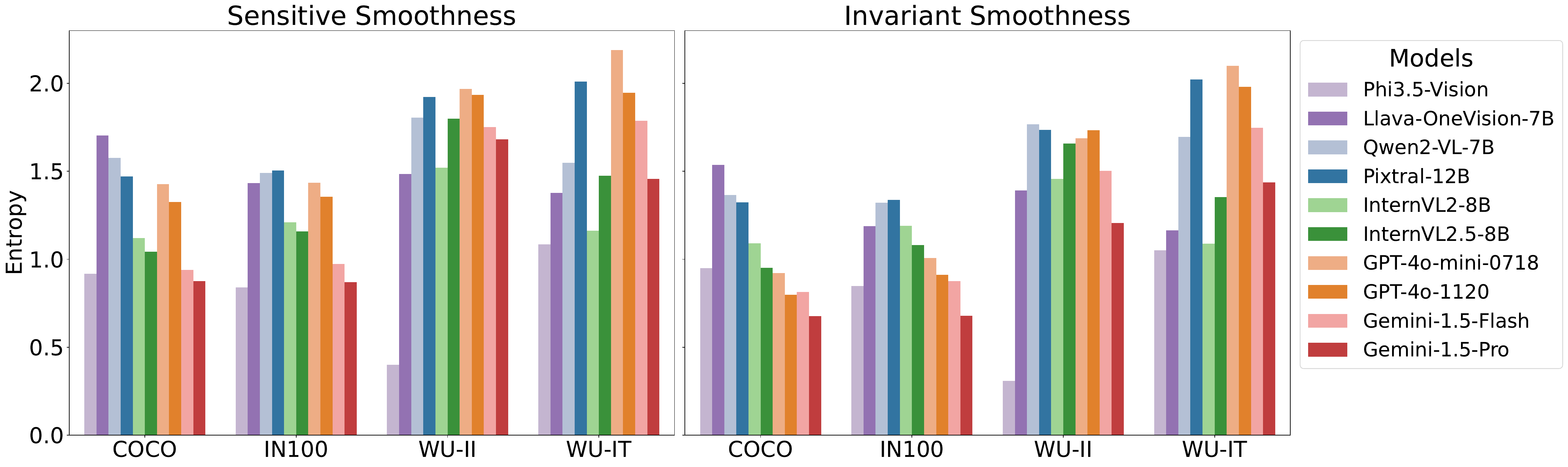}
    \includegraphics[width=0.95\linewidth,trim={.1cm .2cm .2cm .2cm},clip]{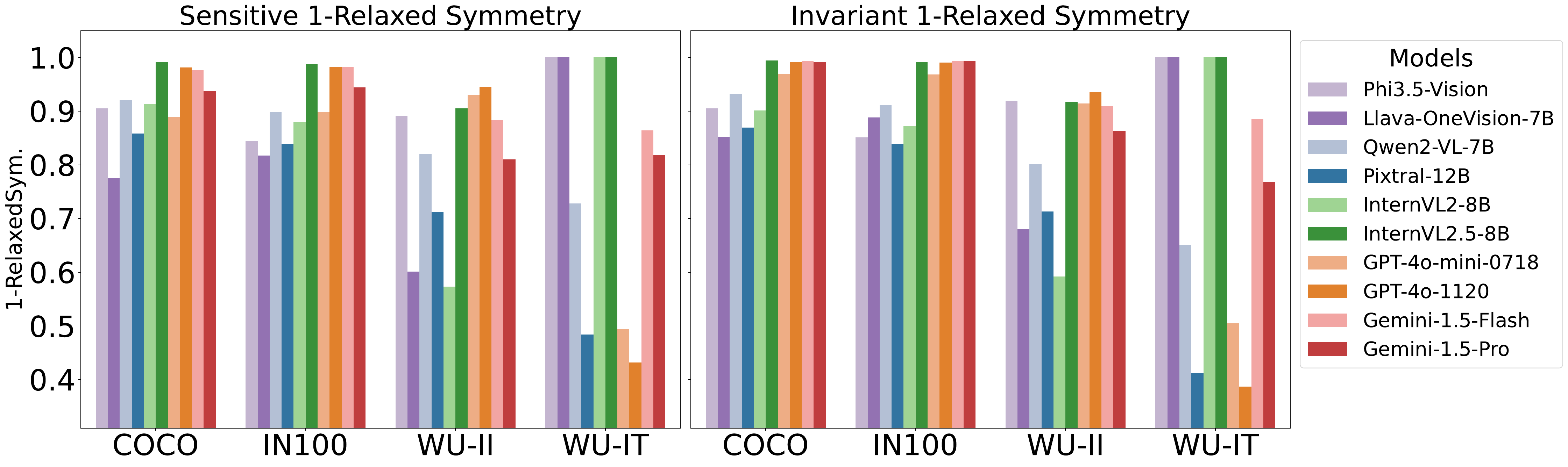}
    \caption{\nmi, Smoothness, and Controllability for the best performing models in both \texttt{sens} and \texttt{invar} settings.}
    \label{fig:best-models-mmscore-smoothness-sym}
\end{figure*}

\begin{figure}
    \centering
    \includegraphics[width=0.95\linewidth]{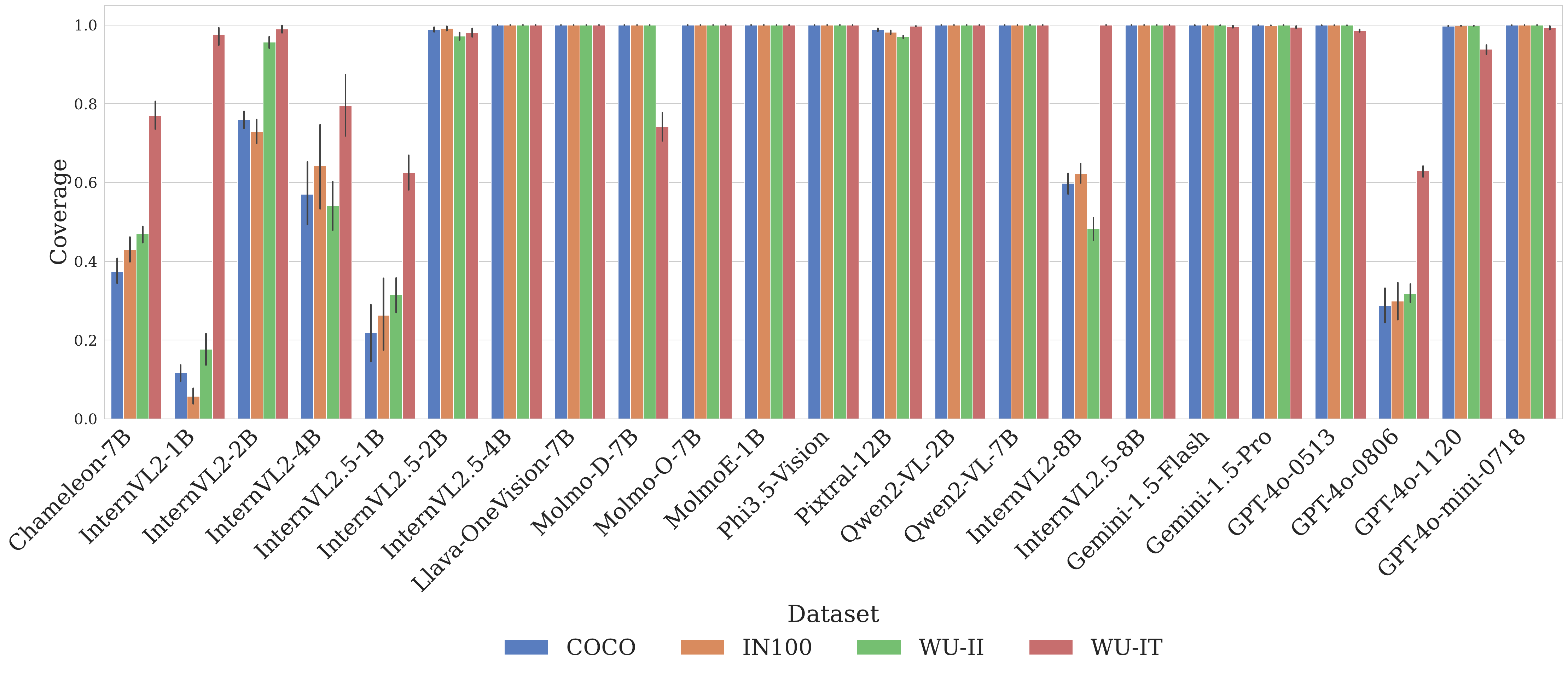}
    \caption{Coverage of each model.}
    \label{fig:coverage-bars}
\end{figure}

\subsection{All \relaxsym for different $\epsilon$s}
To show the \relaxsym{} for different values of $\varepsilon$, we plot Figure \ref{fig:diff-relax-sym-eps} and show as $\varepsilon$ gets higher, the values go higher. However, some models such as the GPT4o models struggle with symmetry. Please note that if $\varepsilon = 0$, it is the same as not having a threshold and hence calculating exact symmetry rather than a relaxed version.

\begin{figure*}[ht]
    \centering
    \includegraphics[width=0.95\linewidth,trim={.1cm .2cm .2cm .2cm},clip]{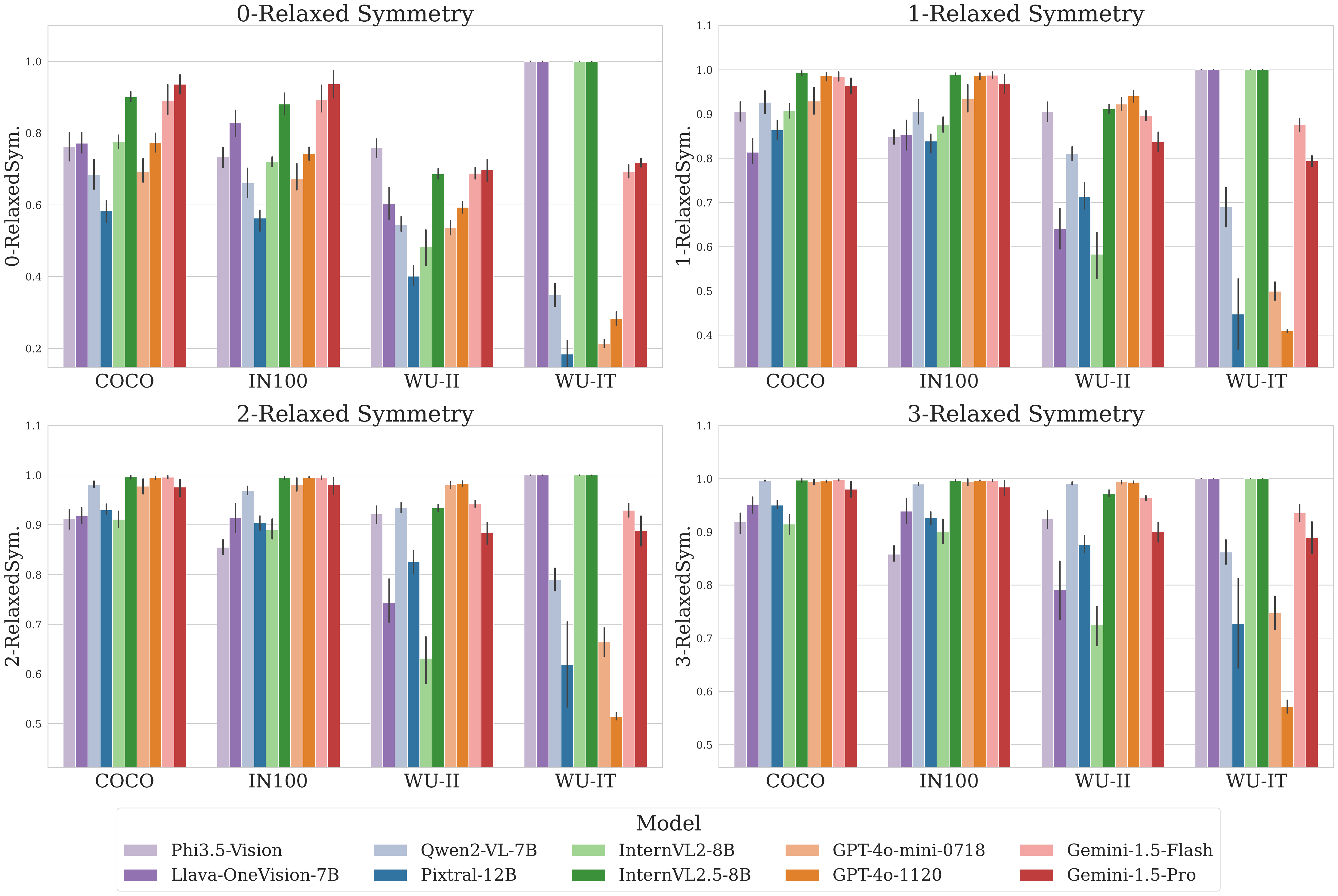}
    \caption{\relaxsym{} for different $\varepsilon$s.}
    \label{fig:diff-relax-sym-eps}
\end{figure*}

\subsection{Different versions of same model}
\label{sec:model-versions}
% We further look into the affect of capacity on the different metrics of \mmscore. As seen in Figure \ref{fig:mi-model-versions} and \ref{fig:all-model-versions}, the larger capacity models tend to better across \nmi, \relaxsym, and \control. However, we observe there are exceptions, e.g., \internvlTwoFourB{} being more controllable in rotation (R) and perspective shift (PS), compared to \internvlTwoEightB. Also, we see \smoothness{} is not monotonically increasing as the model capacity increases. This shows that the stronger models may tend to be more certain about their responses, hence not generating similarity scores as diverse as the lower capacity ones.

% On the other hand, we saw in Table \ref{tab:benchmark_comparison} and Figure \ref{fig:control-vs-bm} that Smoothness also has a positive correlation with model performance and other benchmarks, showing that better models tend to be more smoother and create more diverse outputs compared to the weaker ones. Ultimately, we conclude that \smoothness{} is not a property of performance; however, it is a characteristic of a \model{} as a judge model which could be desirable depending on the use-case.

We further examine the effect of model capacity on the different metrics of \mmscore. As seen in Figure \ref{fig:all-model-versions}, larger-capacity models tend to perform better across \nmi, \relaxsym, and \control. However, there are exceptions—for example, \internvlTwoFourB{} demonstrates greater controllability in rotation (R) and perspective shift (PS) compared to \internvlTwoEightB. Additionally, smoothness (\smoothness{}) does not increase monotonically with model capacity. This suggests that stronger models may be more confident in their responses, leading to less diversity in their similarity scores compared to lower-capacity models.

On the other hand, Table \ref{tab:benchmark_comparison} and Figure \ref{fig:control-vs-bm} show that \smoothness{} correlates positively with model performance and other benchmarks, indicating that better models tend to produce smoother and more diverse outputs than weaker ones. Ultimately, we conclude that \smoothness{} is not strictly a property of model performance but rather a characteristic of a \model{} as a judge model that may be desirable (or not) depending on the use case.

\begin{figure*}[ht]
    \centering

    \includegraphics[width=0.95\linewidth,trim={.1cm .2cm .2cm .2cm},clip]{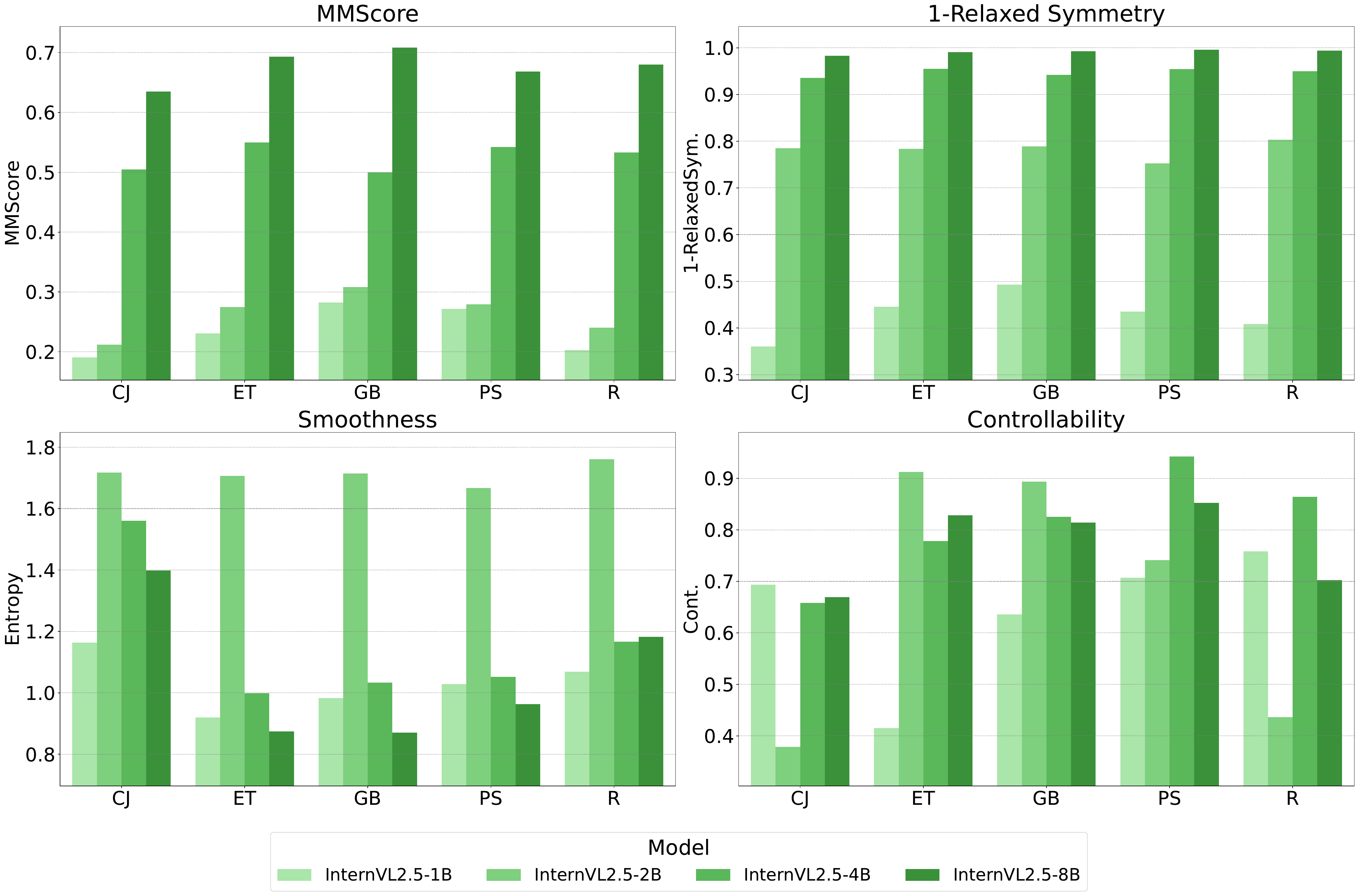}
    % \caption{Caption for Third PDF.}
    % \label{fig:third-plot}

    \caption{Aggregated \mmscore{} metrics across different versions of InternVL2.5 models.}
    \label{fig:all-model-versions}
\end{figure*}

\subsection{Correlations}
\label{sec:bm-correlations}

In this section, we further plot the correlations of the different metrics and show them in Figures \ref{fig:sym-vs-bm}, \ref{fig:control-vs-bm}, \ref{fig:control-vs-bm}. As seen, all these metrics have positive correlations as seen in the scatter plots.

\begin{figure*}[ht]
    \centering
    \includegraphics[width=0.95\linewidth,trim={.1cm .2cm .2cm .2cm},clip]{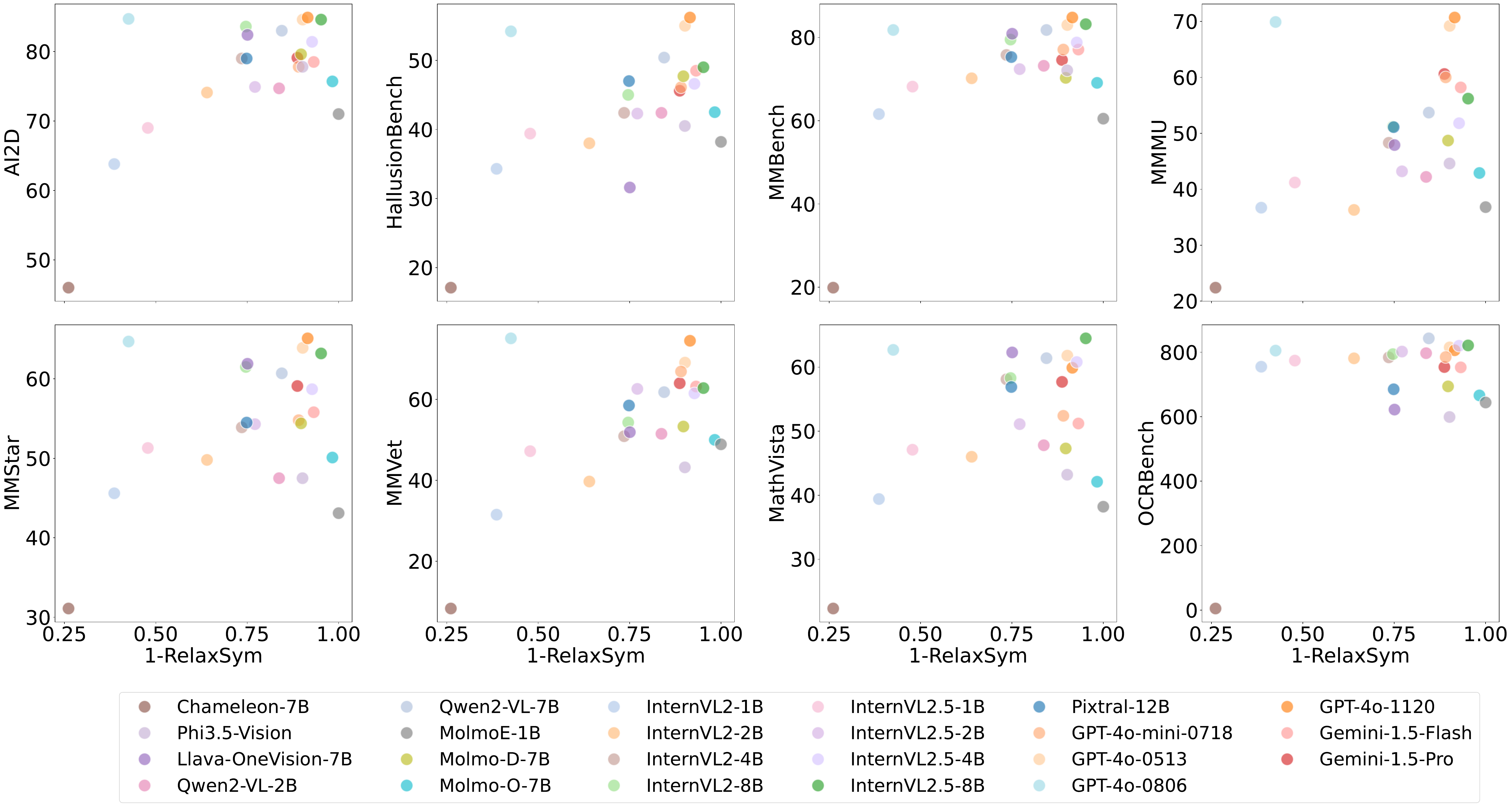}
    \caption{Other benchmarks versus \mmscore{} on \relaxsymone{}.}
    \label{fig:sym-vs-bm}
\end{figure*}
\begin{figure*}[ht]
    \centering
    \includegraphics[width=0.95\linewidth,trim={.1cm .2cm .2cm .2cm},clip]{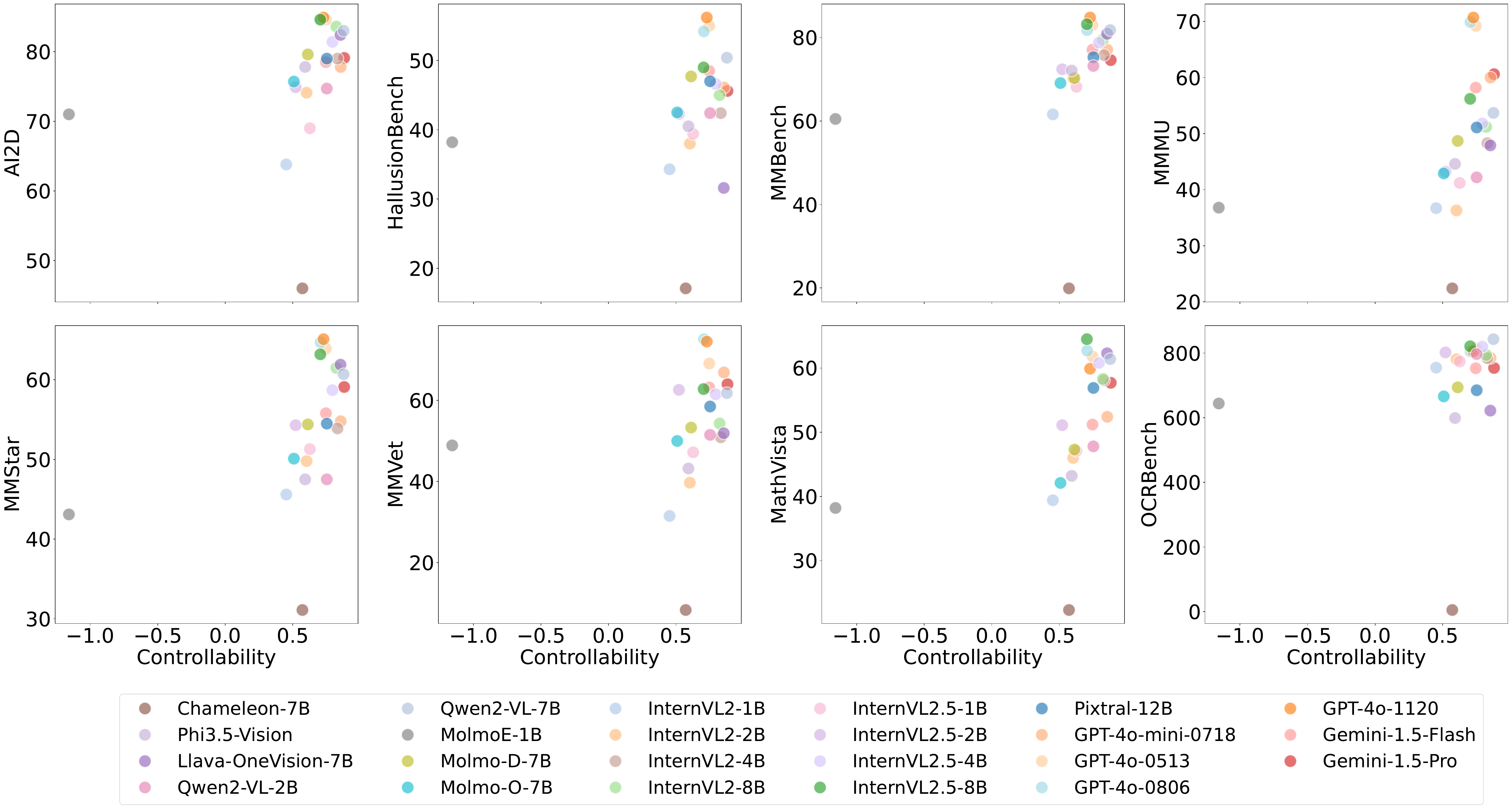}
    \caption{Other benchmarks versus \control{} on \mmscore{}.}
    \label{fig:control-vs-bm}
\end{figure*}
\begin{figure*}[ht]
    \centering
    \includegraphics[width=0.95\linewidth,trim={.1cm .2cm .2cm .2cm},clip]{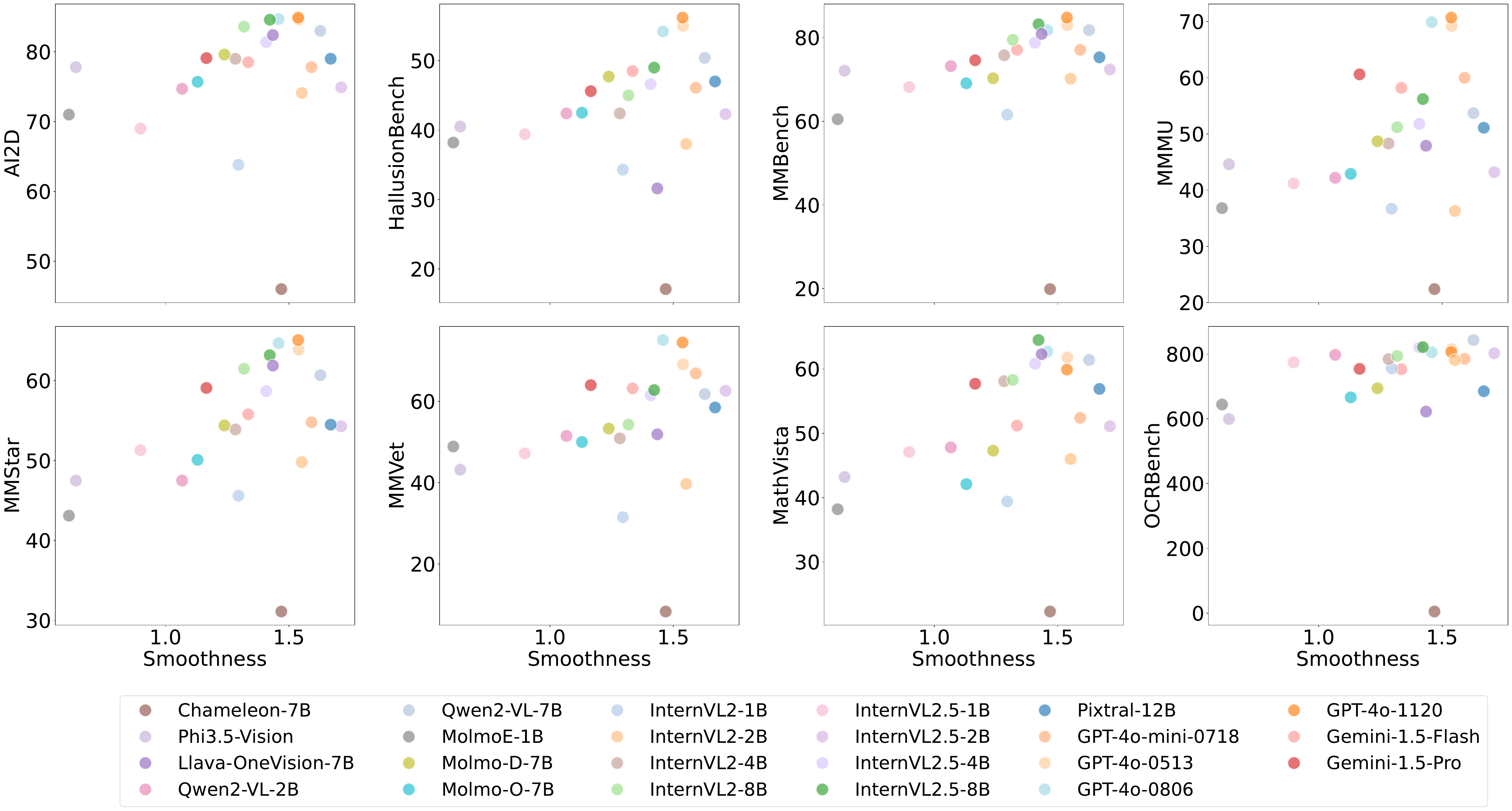}
    \caption{Other benchmarks versus Smoothness (\smoothness).}
    \label{fig:smoothness-vs-bm}
\end{figure*}

\subsection{Encoders vs \modelss{}}    
\label{sec:ecoders_vs_decoders}

For the image-image task, we explore how image encoders compare to \modelss{} on our metrics. To this end, three DINOv2 versions (\dinoBase, \dinoSmall, and \dinoLarge) and the LAION- and OpenAI- CLIP-trained ViTs (base and large) are chosen to encode images. Since feature controllability on image-encoders is limited to the image augmentation transformation (CJ, R, PS, GB, ET), we only compare image-encoders to \modelss{} on \mmscorecoco{} and \mmscorein.

To generate the similarity score of a given image-pair with an image-encoder, we compute the cosine similarity of the representation of each image and scale the scores between 1-10, and round them to the nearest integer. To generate the criteria-sensitive similarity score, we create the representations of the image-pair by simply using the representations output by the encoder for each image. On the other hand, when generating the criteria-invariant score, where the criteria is a specific transformation ($T$), we generate the representation of each image as the average of the representations of the encoder for $k$ versions of the image where random amounts of $T$ are applied to the image. In our experiments, we set $k=5$.

We report results in Figure \ref{fig:mi-encoder-vs-lmms}. We see encoders do better than open-source \modelss{} most of the time and are comparable to closed-source models (besides CJ). This shows although significantly smaller, encoders can be at least as good as \modelss{}, enabling similarity scoring at a much lower cost. Also, encoder-generated scores are trivially symmetric as well since the underlying cosine similarity is symmetric. However, they lack in controllability as they are limited to image-only comparisons and can only consider criteria that can be applied to the image using augmentations, i.e., spatial position transform cannot be applied to images for encoders.

\vspace{-2mm}

\begin{figure*}
    \includegraphics[width=0.92\linewidth,trim={.1cm .2cm .2cm .2cm},clip]{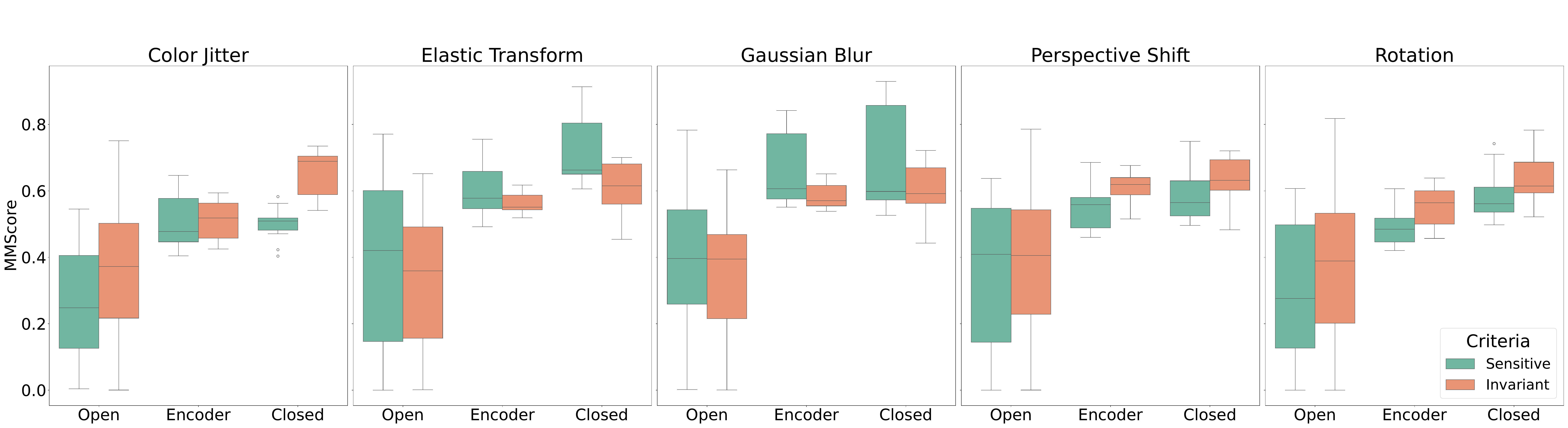}
    
    \caption{A simple vision encoder outperforms open-sourced \modelss{} and has on par performance with closed sourced models which are much more expensive, for image-image tasks (results combine \mmscorecoco{} and \mmscorein), and similar pattern is observed across different transformations.}
    \label{fig:mi-encoder-vs-lmms}
\end{figure*}

% \input{imgs/sym-plot}
% \newpage
\section{\mmscore{} Details}
\label{sec:mmscore-info}

\subsection{Dataset Creation}
The \mmscore{} framework takes in a source dataset and creates augmented versions of the data to obtain data pairs to probe the evaluation skills of a model. 
In our instances, we use \coco{} \citep{lin2014microsoft}, \imagenet{} \citep{deng2009imagenet} and \wu{} \citep{kamath2023s} datasets as the source for the original data points. We utilize \coco{} and \imagenet{} as image-only datasets and \wu{} as an image-text dataset. We select 500 random images from each of \coco{} and \imagenet{} and all the image-text pairs from both subsets provided by the \wu{} dataset to be used in our instantiation of \mmscore. Full details of our released datasets are given in Table \ref{tab:mmscore-info}.

\begin{table*}[ht]
\centering
\caption{Information of different splits in \mmscorecoco, \mmscorein, \mmscorewuimgimg, and \mmscorewuimgtext.}
\label{tab:mmscore-info}
\resizebox{\textwidth}{!}{%
\begin{tabular}{llccc}
\toprule
\textbf{Modality} & \textbf{Source} & \textbf{Number of Selected} & \textbf{Splits} & \textbf{New Data Points / Total Data-Pair Comparisons} \\
\midrule
% \multirow{11}{*}{MMSCore (Image-Image)} & \multirow{6}{*}{COCO} & \multirow{6}{*}{500} & CJ & 1000 / 3000 \\ 
%  & & & R & 1000 / 3000 \\ 
%  & & & ET & 1000 / 3000 \\ 
%  & & & PS & 1000 / 3000 \\ 
%  & & & GB & 1000 / 3000 \\ 
% \cmidrule{2-5}
%  & \multirow{6}{*}{IN100} & \multirow{6}{*}{500} & CJ & 1000 / 3000 \\ 
%  & & & R & 1000 / 3000 \\ 
%  & & & ET & 1000 / 3000 \\ 
%  & & & PS & 1000 / 3000 \\ 
%  & & & GB & 1000 / 3000 \\ 
% \cmidrule{2-5}
\multirow{22}{*}{\mmscoreimgimg} & \multirow{6}{*}{\coco} & \multirow{6}{*}{500} & CJ & 1000 / 3000 \\ 
 & & & R & 1000 / 3000 \\ 
 & & & ET & 1000 / 3000 \\ 
 & & & PS & 1000 / 3000 \\ 
 & & & GB & 1000 / 3000 \\ 
\cmidrule{2-5}
 & \multirow{6}{*}{\imagenet} & \multirow{6}{*}{500} & CJ & 1000 / 3000 \\ 
 & & & R & 1000 / 3000 \\ 
 & & & ET & 1000 / 3000 \\ 
 & & & PS & 1000 / 3000 \\ 
 & & & GB & 1000 / 3000 \\ 
\cmidrule{2-5}
 & \multirow{6}{*}{\wu{} (subset A)} & \multirow{6}{*}{418} & SP & 0 / 3344 \\ 
 & & & SP \& CJ & 1254 / 3344 \\ 
 & & & SP \& R & 1254 / 3344 \\ 
 & & & SP \& ET & 1254 / 3344 \\ 
 & & & SP \& PS & 1254 / 3344 \\ 
 & & & SP \& GB & 1254 / 3344 \\ 
\cmidrule{2-5}
 & \multirow{6}{*}{\wu{} (subset B)} & \multirow{6}{*}{408} & SP & 0 / 3264 \\ 
 & & & SP \& CJ & 1224 / 3264 \\ 
 & & & SP \& R & 1224 / 3264 \\ 
 & & & SP \& ET & 1224 / 3264 \\ 
 & & & SP \& PS & 1224 / 3264 \\ 
 & & & SP \& GB & 1224 / 3264 \\ 
\midrule
\multirow{2}{*}{\mmscoreimgtext} & \wu{} (subset A) & 418 & SP & 1254 / 3344 \\ 
 & \wu{} (Subset B) & 408 & SP & 1224 / 3264 \\ 
\midrule
\textbf{In total} & - & 1826 & all splits & \textbf{22390 / 69648} \\ 
\bottomrule
\end{tabular}
}
\end{table*}

% Considering that the \mmscore{} framework aims to measure how well \modelss{} can detect different features and score them between data points, 
To isolate the effect of different data characteristics on model performance,
\mmscore{} creates pairs of image-image and image-text data that are identical except for one or a few controlled features. The generated data consists of points from the original dataset paired with their transformed version.
% these image-image pairs by using all three source datasets and construct the image-text pairs using the two subsets of \wu{}. 
For \coco{} and \imagenet, we create a different control sample for each one of the transformations in $\{$color jitter, rotation, gaussian blur, perspective shift, elastic transformation$\}$, which defines the characteristic that differs between images. For the data from \wu{}, we construct the data pairs by either only using the `spatial position' transform, or `spatial position' transform in addition to one of the previous five characteristics to additionally assess coupling effects. 
However, since transforms are not well-defined for texts, only `spatial position' transform is applied for the image-text pairs. Note that the image-image pairs from \wu{} are the most challenging since they all have at least the `spatial position' transform, which is a well-known blind-spot of \modelss{} as shown by previous literature \citep{kamath2023s, wang2024picture}. As a result, we end up creating five image-image sub-datasets for each of \coco{} and \imagenet, six subsets for each of the two subsets of \wu{}, using each of the transformations, and one image-text sub-dataset for each of the subsets of \wu{}. The details of the transforms applied to each category are shown in Figure \ref{fig:mmscore-examples}.

Next, for each original image, we construct three types of pairs: an identical, a transformed, and an irrelevant pair. In all three versions of these pairs, the first data point is the original (non-transformed) image. For the `identical' pair, the second data point is another version of the image with $95\%$ of its original size for the image-image pair and the correct caption for the image-text pair. The second data point in the `transformed' pair is the original image (caption) with the transformation applied to it for the image-image (image-text) pair. Finally, the `irrelevant' pair's second data point is a transformed version of a random image (caption) from the rest of the dataset. 

Equipped with the constructed control samples, \mmscore{} prompts the \model{} to score the similarity of each data pair based on a set of criteria. 
The criteria consists of the conditions indicating whether the model under examination should be `sensitive' or `invariant' to the transformations applied for that specific sub-dataset. These two settings (sensitive or invariant) measure how well each model can recognize the differences between the data pair and follow the prompt's criteria. If a model can successfully capture a specific feature, it will have no problem being variant or invariant to it; however, if it cannot detect it or has a bias towards a feature, it will favor being sensitive or invariant to that feature over its opposite. 
Using a human study, described in Appendix \ref{sec:human-study}, the ground-truth score of the `identical' and `irrelevant' pair are set to 10 and 1, respectively, in both `sensitive' and `invariant' settings. However, for the `transformed' pair, based on the human study we set the score 10 in the `invariant' version, and `6' in the `sensitive' version of the prompt.
% On a scale of 1 to 10, we consider the ground-truth score of the `identical' and `irrelevant' pair 10 and 1, respectively, in both `sensitive' and `invariant' settings. However, for the `transformed' pair, we consider the score 10 in the `invariant' version, and `8' in the `sensitive' version of the prompt.
To make sure the performance gap between models is not merely a consequence of biased prompt wording, \mmscore{} comes with five template prompts with different lengths and wordings but with the same semantic meaning, that are randomly selected for each data pair, to make sure the prompting does not affect the model's performance. These prompt templates are reported in Appendix \ref{sec:prompt-templates}.

Ultimately, we end up with 4 different datasets created by \mmscore: \mmscorecoco{}, \mmscorein{}, \mmscorewuimgimg{}, and \mmscorewuimgtext{}. \mmscorecoco{} and \mmscorein{} compare and score image-pairs and have 5 splits (Color Jitter (CJ), Rotation (R), Gaussian Blur (GB), Perspective Shift (PS), and Elastic Transformation (ET). \mmscorewuimgimg{} consists of 2 subsets, each with 6 splits; one split with only the Spatial Position transform (SP), and the rest with SP combined with one of the previous five transformations (CJ, R, GB, PS, and ET). \mmscorewuimgtext{} consists of only the SP split for each of the two subsets in the \wu{} dataset. Details of each split in Appendix \ref{sec:mmscore-info}.

\subsection{Human Study for Ground Truth Scores}
\label{sec:human-study}
To validate the alignment between our ground-truth scores and human perception, we conducted a human study on \textbf{image-image} pairs from \mmscorein{}. We excluded image-text comparisons due to their trivial nature for human judgment. For example, given an image showing a book to the left of a cap, comparing it to the sentence “book left of cap” (identical), “book right of cap” (transformed), or “can behind candle” (irrelevant) would result in nearly unanimous responses, offering limited insight.

Our study involved 76 volunteer participants and covered 300 image pairs sampled across three transformation types—color jitter, perspective shift, and rotation—under both “sensitive” and “invariant” settings. Results from this evaluation led us to adjust the ground-truth score of transformed pairs in the “sensitive” condition to 6 (on a 1–10 scale), as this better captured the perceptual similarity reported by humans. Furthermore, the study confirmed that “identical” pairs consistently received the highest scores, while “irrelevant” pairs received the lowest, supporting the validity of our scoring protocol. A screenshot of the study can be seen in Figure \ref{fig:human-study-ss} and the results are reported in Table \ref{tab:human-study-results}.

\begin{figure}
    \centering
    \includegraphics[width=0.95\linewidth]{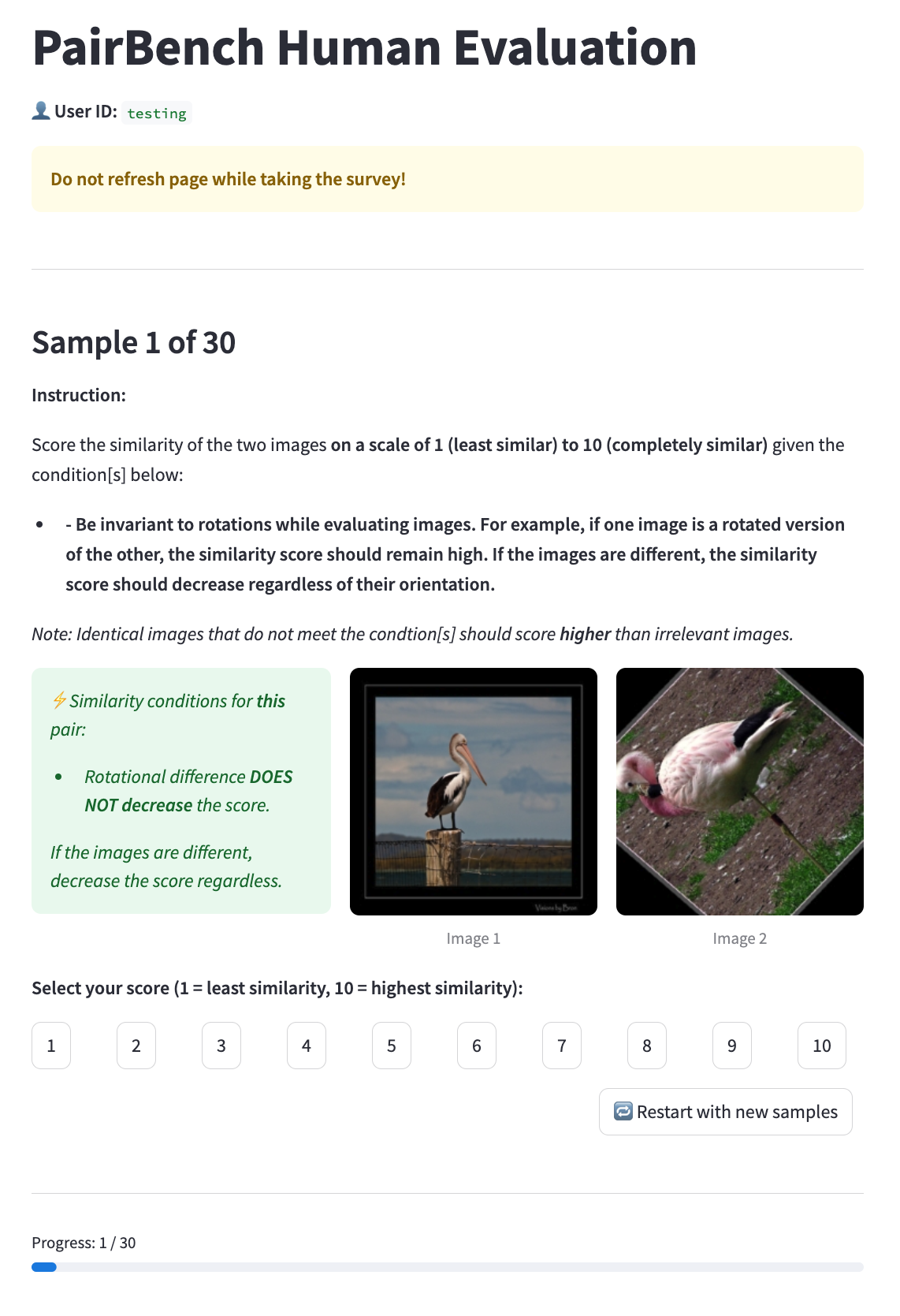}
    \caption{Example screenshot of the platform used consisting of the type of questions that participants were asked during the study.}
    \label{fig:human-study-ss}
\end{figure}

\begin{table}[ht]
\centering
\caption{Human similarity scores (mean ± std) across different transformation settings, which we used to set our ground truth scores.}
\label{tab:human-study-results}
\resizebox{\linewidth}{!}{%
\begin{tabular}{lcc|cc|cc}
\toprule
\textbf{Pair Type} 
& \multicolumn{2}{c|}{\textbf{Colorjitter}} 
& \multicolumn{2}{c|}{\textbf{Perspective}} 
& \multicolumn{2}{c}{\textbf{Rotate}} \\
& \textbf{Sens} & \textbf{Invar} 
& \textbf{Sens} & \textbf{Invar} 
& \textbf{Sens} & \textbf{Invar} \\
\midrule
Identical   & 9.8 ± 0.53  & 9.95 ± 0.23  & 9.82 ± 0.42  & 9.9 ± 0.32   & 10.0 ± 0.0   & 9.89 ± 0.51 \\
Transformed & 5.5 ± 2.3   & 9.68 ± 0.85  & 6.89 ± 1.97  & 9.06 ± 1.33  & 6.31 ± 2.01  & 9.36 ± 1.01 \\
Irrelevant  & 1.35 ± 0.74 & 1.42 ± 0.67  & 1.33 ± 0.81  & 1.61 ± 1.17  & 1.27 ± 0.64  & 1.23 ± 0.47 \\
\bottomrule
\end{tabular}
}
\end{table}

\subsection{Prompt Templates for Different \mmscore{} datasets}
\label{sec:prompt-templates}
We provide the 5 different templates that we choose at random for each data pair for the image-image and image-text prompts.

The following are the templates we utilize for \mmscorecoco and \mmscorein, and \mmscorewuimgimg, i.e., the image-image pairs.

\begin{tcolorbox}[enhanced,attach boxed title to top center={yshift=-3mm,yshifttext=-1mm},
  colback=blue!5!white,colframe=blue!20!gray,colbacktitle=blue!20!gray,
  title=Image-Image Prompt Template V1,fonttitle=\bfseries,
  boxed title style={size=small,colframe=blue!20!gray} ]

        \emph{User prompt}: You are tasked with evaluating the similarity between two images while paying attention to the following conditions: \texttt{\{conditions\}}. Your goal is to judge the similarity of the images overall, where satisfying the conditions increases the similarity score. If the images are identical but fail to meet any of the conditions, they should still receive a higher score than completely unrelated images. Provide a similarity score on a scale from 1 to 10, where 1 represents entirely dissimilar images and 10 represents identical images that satisfy all conditions. Ensure your response is strictly in the following format:
        
        \begin{verbatim}
        Score: <1-10>
        Reason: <reason for score>
        \end{verbatim}
                
        Do not include anything else in your response. What score would you assign to this pair of images? \texttt{"data1"} \texttt{"data2"}.

\end{tcolorbox}

\begin{tcolorbox}[enhanced,attach boxed title to top center={yshift=-3mm,yshifttext=-1mm},
  colback=blue!5!white,colframe=blue!20!gray,colbacktitle=blue!20!gray,
  title=Image-Image Prompt Template V2,fonttitle=\bfseries,
  boxed title style={size=small,colframe=blue!20!gray} ]

        \emph{User prompt}: As a similarity evaluator, your responsibility is to assess the similarity of the given images while considering these conditions: \texttt{\{conditions\}}. The similarity score should reflect both how well the images align with the conditions and their overall resemblance. Images that are identical but do not meet the conditions should receive a moderate score, while completely unrelated images should receive the lowest score. Provide your score on a scale of 1 to 10, with 10 being identical images that fully meet the conditions. Ensure your response is in the following format:
        
        \begin{verbatim}
        Score: <1-10>
        Reason: <reason for score>
        \end{verbatim}
                
        Provide nothing else. What is your score? \texttt{"data1"} \texttt{"data2"}

\end{tcolorbox}

\begin{tcolorbox}[enhanced,attach boxed title to top center={yshift=-3mm,yshifttext=-1mm},
  colback=blue!5!white,colframe=blue!20!gray,colbacktitle=blue!20!gray,
  title=Image-Image Prompt Template V3,fonttitle=\bfseries,
  boxed title style={size=small,colframe=blue!20!gray} ]

        \emph{User prompt}: Evaluate the similarity of the images based on the following conditions: \texttt{\{conditions\}}. The score should take into account how well the images align with these conditions, as well as their overall resemblance. Even if the images are identical but fail to meet the conditions, they should still receive a higher score than completely different images. Provide a score from 1 to 10, where 1 indicates no similarity and 10 indicates identical images that fully satisfy the conditions. Respond only in this format:
        
        \begin{verbatim}
        Score: <1-10>
        Reason: <reason for score>
        \end{verbatim}
                
        Nothing else should be included. What score would you give? \texttt{"data1"} \texttt{"data2"}

\end{tcolorbox}

\begin{tcolorbox}[enhanced,attach boxed title to top center={yshift=-3mm,yshifttext=-1mm},
  colback=blue!5!white,colframe=blue!20!gray,colbacktitle=blue!20!gray,
  title=Image-Image Prompt Template V4,fonttitle=\bfseries,
  boxed title style={size=small,colframe=blue!20!gray} ]

        \emph{User prompt}: Judge the similarity of these images based on: \texttt{\{conditions\}}. The similarity score should reflect both the overall resemblance of the images and how well they satisfy the conditions. Identical images that do not meet the conditions should still score higher than completely unrelated images. Provide a score on a scale of 1 to 10, with 1 being no similarity and 10 being identical images that satisfy all conditions. Respond strictly in this format:
        
        \begin{verbatim}
        Score: <1-10>
        Reason: <reason for score>
        \end{verbatim}
                
        Do not include additional text. What's your rating? \texttt{"data1"} \texttt{"data2"}

\end{tcolorbox}

\begin{tcolorbox}[enhanced,attach boxed title to top center={yshift=-3mm,yshifttext=-1mm},
  colback=blue!5!white,colframe=blue!20!gray,colbacktitle=blue!20!gray,
  title=Image-Image Prompt Template V5,fonttitle=\bfseries,
  boxed title style={size=small,colframe=blue!20!gray} ]

        \emph{User prompt}: Rate the similarity of these images on a scale of 1 to 10 while considering \texttt{\{conditions\}}. The score should account for both how well the images satisfy the conditions and their overall resemblance. If the images are identical but do not meet the conditions, they should still score higher than unrelated images. Your response must follow this exact format:
        
        \begin{verbatim}
        Score: <1-10>
        Reason: <reason for score>
        \end{verbatim}
                
        Provide nothing else. What's your score? \texttt{"data1"} \texttt{"data2"}

\end{tcolorbox}

And these are the image-text versions for \mmscorewuimgtext:

\begin{tcolorbox}[enhanced,attach boxed title to top center={yshift=-3mm,yshifttext=-1mm},
  colback=green!5!white,colframe=green!20!gray,colbacktitle=green!20!gray,
  title=Image-Text Prompt Template V1,fonttitle=\bfseries,
  boxed title style={size=small,colframe=green!20!gray} ]

        \emph{User prompt}: You are assigned the task of determining how accurately a text description represents an image of two objects. Pay special attention to the following conditions: \texttt{\{conditions\}}. Your goal is to assess the alignment between the image and the text, where adherence to the conditions increases the similarity score. If the text partially describes the image but fails to meet specific conditions, it should still score higher than a text entirely unrelated to the image. Assign a similarity score on a scale from 1 to 10, where 1 signifies no correlation, and 10 denotes perfect alignment satisfying all conditions. The data pair (text and image) will follow this prompt. Respond strictly in the following format:
        
        \begin{verbatim}
        Score: <1-10>
        Reason: <reason for score>
        \end{verbatim}
        
        What score would you assign to this text-image pair?\
        \texttt{"data1"}\
        \texttt{"data2"}.\
        
\end{tcolorbox}

\begin{tcolorbox}[enhanced,attach boxed title to top center={yshift=-3mm,yshifttext=-1mm},
  colback=green!5!white,colframe=green!20!gray,colbacktitle=green!20!gray,
  title=Image-Text Prompt Template V2,fonttitle=\bfseries,
  boxed title style={size=small,colframe=green!20!gray} ]

        \emph{User prompt}: Figure out how well this image matches the description provided. The image shows two objects, and the text is meant to describe how they're arranged. Look at these specific conditions: \texttt{\{conditions\}}. If the text captures some parts of the image but misses others, it should still get a better score than something totally off. Score this match on a scale of 1 to 10, where 1 means there's no match and 10 means the description nails it and matches every condition perfectly. The text and image will follow this prompt. Answer in this format only:
        
        \begin{verbatim}
        Score: <1-10>
        Reason: <reason for score>
        \end{verbatim}
        
        What's your score?\
        \texttt{"data1"}\
        \texttt{"data2"}.\
        
\end{tcolorbox}

\begin{tcolorbox}[enhanced,attach boxed title to top center={yshift=-3mm,yshifttext=-1mm},
  colback=green!5!white,colframe=green!20!gray,colbacktitle=green!20!gray,
  title=Image-Text Prompt Template V3,fonttitle=\bfseries,
  boxed title style={size=small,colframe=green!20!gray} ]

        \emph{User prompt}: Evaluate the degree to which a text description accurately represents an image featuring two objects, taking into account the following conditions: \texttt{\{conditions\}}. Assign a score based on how well the image-text pair matches, where:
        - A perfect description that satisfies all conditions scores 10.
        - Texts that partially align with the image but fail to meet conditions should still score higher than completely unrelated ones.
        The data pair will follow this prompt. Provide your score on a scale of 1 to 10 using the exact format below:
        
        \begin{verbatim}
        Score: <1-10>
        Reason: <reason for score>
        \end{verbatim}
        
        What score would you give?\
        \texttt{"data1"}\
        \texttt{"data2"}.\
        
\end{tcolorbox}

\begin{tcolorbox}[enhanced,attach boxed title to top center={yshift=-3mm,yshifttext=-1mm},
  colback=green!5!white,colframe=green!20!gray,colbacktitle=green!20!gray,
  title=Image-Text Prompt Template V4,fonttitle=\bfseries,
  boxed title style={size=small,colframe=green!20!gray} ]

        \emph{User prompt}: You are tasked with reviewing how well a text description aligns with an image of two objects. The score should reflect not only the accuracy of the alignment but also how well the description satisfies the following conditions: \texttt{\{conditions\}}. Even if the text description captures some parts of the image while failing the conditions, it should still receive a higher score than a completely irrelevant description. The text and image will be provided below. Assign a score on a 1 to 10 scale, where 1 is no similarity and 10 is perfect alignment that meets all conditions. Answer only in this format:
        
        \begin{verbatim}
        Score: <1-10>
        Reason: <reason for score>
        \end{verbatim}
        
        What score would you assign?\
        \texttt{"data1"}\
        \texttt{"data2"}.\
        
\end{tcolorbox}

\begin{tcolorbox}[enhanced,attach boxed title to top center={yshift=-3mm,yshifttext=-1mm},
  colback=green!5!white,colframe=green!20!gray,colbacktitle=green!20!gray,
  title=Image-Text Prompt Template V5,fonttitle=\bfseries,
  boxed title style={size=small,colframe=green!20!gray} ]

        \emph{User prompt}: Assess the degree to which a text description corresponds to an image of two objects, taking into account the following conditions: \texttt{\{conditions\}}. The scoring should reflect:
        - A perfect alignment with the image that satisfies all conditions merits a score of 10.
        - Descriptions that partially match the image but fail to meet certain conditions should still receive a higher score than entirely unrelated descriptions.
        - A score of 1 should be reserved for cases where no correlation exists between the text and the image.
        The text and image pair will be provided below. Provide your evaluation using the following format:
        
        \begin{verbatim}
        Score: <1-10>
        Reason: <reason for score>
        \end{verbatim}
        
        What score would you assign?\
        \texttt{"data1"}\
        \texttt{"data2"}.\
        
\end{tcolorbox}

% \input{sections/appendix/85_reproducibility}
% \input{sections/appendix/86_human_study}
%%%%%%%%%%%%%%%%%%%%%%%%%%%%%%%%%%%%%%%%%%%%%%%%%%%%%%%%%%%%

% \newpage
% \input{sections/99_checklist}

\end{document}